\documentclass[]{digital_agent_hub}

\usepackage{pifont}
\usepackage{float}
\usepackage{longtable}
\usepackage{listings}
\usepackage{tabularx}
\usepackage{amssymb}
\usepackage{makecell}
\usepackage{xurl}

\newcommand{\gameworld}{{\sffamily\bfseries GameWorld}}
\newcommand{\gameworldrt}{{\sffamily\bfseries GameWorld-RT}}
\newcommand{\gameid}[1]{\begingroup\urlstyle{tt}\nolinkurl{#1}\endgroup}

\newcommand{\gamecite}[2]{#1~\cite{#2}}

\definecolor{headerblue}{HTML}{eef2f8}
\definecolor{bggray}{HTML}{F2F2F2}
\definecolor{textgray}{HTML}{4A4A4A}
\definecolor{rankbgyellow}{HTML}{FFF8E1}
\definecolor{projectpagelink}{HTML}{DA2F8A}

\newcommand{\captfont}[1]{#1}

\begin{document}

\title{\gameworld{}: Towards Standardized and \\ Verifiable Evaluation of Multimodal Game Agents}

\author{Mingyu Ouyang$^{*}$$^{1}$, Siyuan Hu$^{*}$$^{1}$, Kevin Qinghong Lin$^{2}$, Hwee Tou Ng$^{1}$$^{\dag}$, Mike Zheng Shou$^{1}$$^{\dag}$}
\affiliation{$^{1}$National University of Singapore, $^{2}$University of Oxford}

\projectpage{\href{https://gameworld-bench.github.io}{\textcolor{projectpagelink}{https://gameworld-bench.github.io}}}
\firstpagefootnote{Technical Report. $^{*}$Equal contribution. $^{\dag}$Corresponding authors.}

\abstract{
Towards an embodied generalist for real-world interaction, Multimodal Large Language Model (MLLM) agents still suffer from challenging latency, sparse feedback, and irreversible mistakes.
Video games offer an ideal testbed with rich visual observations and closed-loop interaction, demanding fine-grained perception, long-horizon planning, and precise control.
However, systematically evaluating these capabilities is currently hindered by heterogeneous action interfaces and heuristic verification.
To this end, we introduce \gameworld{}, a benchmark designed for standardized and verifiable evaluation of MLLMs as generalist game agents in browser environments.
Two game agent interfaces are studied:
(i)~\emph{Computer-use agents} that directly emit keyboard and mouse controls, and (ii)~\emph{Generalist multimodal agents} that act in a semantic action space via deterministic \emph{Semantic Action Parsing}.
\gameworld{} contains \textbf{34 diverse games} and \textbf{170 tasks}, each paired with \emph{state-verifiable} metrics for \emph{outcome-based} evaluation.
The results across \textbf{18 model-interface pairs} suggest that even the best performing agent is far from achieving human capabilities on video games.
Extensive experiments of repeated full-benchmark reruns demonstrate the robustness of the benchmark, while further studies on real-time interaction, context-memory sensitivity, and action validity expose more challenges ahead for game agents.
Together, by offering a standardized, verifiable, and reproducible evaluation framework, \gameworld{} lays a robust foundation for advancing research on multimodal game agents and beyond.
}
\maketitle

\begin{figure}[H]
    \centering
    \vspace{-3em}
    \includegraphics[width=0.9\linewidth]{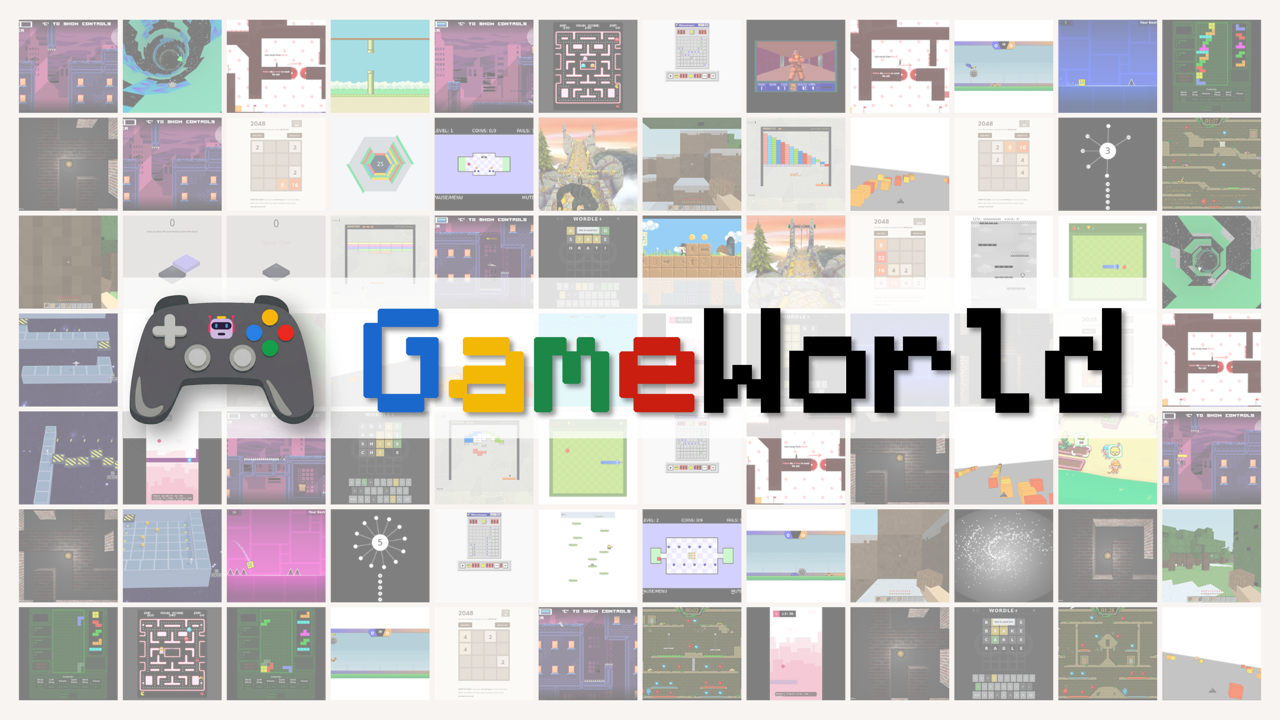}
    \vspace{-0.4em}
    \caption{\captfont{\gameworld{} covers \textbf{34} diverse games with \textbf{170} tasks for standardized evaluation of game agents.}}
    \vspace{-0.2em}
    \label{fig:teaser}
\end{figure}

\makeabstract

\setcounter{tocdepth}{2}
\tableofcontents
\clearpage

\section{Introduction}
\label{sec:intro}

``\textit{A game is a series of interesting choices.}'' --- \textit{Sid Meier}.
Video games tightly couple visual perception, strategic planning, precise timing, and sustained action over long horizons, making them a compelling testbed for evaluating intelligent agents.
Unlike static visual QA or single-turn tool use, games require an agent to repeatedly interpret a changing visual scene, commit to actions with real consequences, and recover from mistakes over many steps.
Within video games, browser games are especially attractive for benchmarking: they are lightweight, mechanically diverse, and easy to reset, providing a scalable alternative to heavyweight game engines or emulators.

Recent Multimodal Large Language Model (MLLM) benchmarks have begun exploring the evaluation of foundation models on games. LMGame-Bench~\cite{hu2025lmgamebench} probes perception, memory, and reasoning through a modular harness across six games. BALROG~\cite{paglieri2025balrog} emphasizes long-horizon play in classic games with both language and vision tracks. VideoGame-Bench~\cite{zhang2025videogamebench} scales to 23 titles with extended trajectories. Orak~\cite{park2025orak} introduces an MCP interface for 12 games. In parallel, execution-based evaluation in interactive environments, as demonstrated by OSWorld~\cite{xie2024osworld} for computer-use tasks, has shown that real interaction reveals performance gaps that static datasets hide. These efforts collectively improve the realism and scale of game-based evaluation, yet several systematic challenges remain unaddressed. Current evaluation is still hindered by heterogeneous action interfaces, latency coupling in real-time interaction, and the lack of outcome-based or verifiable evaluation. Many existing benchmarks still rely on heuristic, OCR, or VLM-as-judge methods, making results harder to verify, reproduce, and diagnose.

To bridge this gap, we introduce \gameworld{}, a standardized benchmark for multimodal game agents in browser environments. \gameworld{} comprises 34 browser games spanning five genres (Runner, Arcade, Platformer, Puzzle, and Simulation) with 170 diverse tasks. A browser-based sandbox pauses game execution during model inference, decoupling inference latency from gameplay so that scores reflect decision quality rather than response speed. 
Each task is paired with an outcome-based state-verifiable evaluator over serialized \texttt{gameAPI} state, producing deterministic progress and success signals without perceptual noise.
Under this shared runtime, we study two agent interfaces: \emph{Computer-Use Agents} (CUAs), which emit raw keyboard and mouse controls, and \emph{Generalist Multimodal Agents}, which act through deterministic \emph{Semantic Action Parsing}. Together, we evaluate 18 model--interface pairs of game agents.

Beyond the main leaderboard, we conduct a set of analyses to study robustness of our benchmark and interface-wise behavior. Repeated full-benchmark reruns show that \gameworld{} yields stable aggregate measurements with only limited run-to-run variation, supporting its use as a reproducible evaluation platform rather than a one-off leaderboard snapshot. We further establish \gameworldrt{}, an unpaused real-time benchmark variant in which environment dynamics continue during inference, making response latency part of the task itself. 
Combining main results with complementary analyses on context-memory sensitivity and action validity, we reveal four broader findings: \textbf{(i)} Current game agents can often \textbf{make meaningful partial progress but remain far from reliable task completion and human-level performance}. \textbf{(ii)} Capability-aligned curriculum profiles show that game-agent performance \textbf{largely inherits the strengths of the underlying foundation models}, with comparatively stronger results on reactive-control and symbolic-reasoning games but clear weaknesses on basic timing grounding, spatial navigation, and long-horizon coordination tasks. \textbf{(iii)} Real-time interaction is a distinct challenge since \textbf{reasoning speed, correctness, and action timing are more tightly coupled}. \textbf{(iv)} The two agent interfaces exhibit \textbf{similar capability bottleneck} and \textbf{distinct trade-offs} in \textbf{context-memory rounds} and \textbf{instruction-following reliability}.
Together, these analyses expose strengths and limitations of current game agents and help guide future improvement directions.

Our contributions are as follows:
\begin{itemize}
    \item \textbf{A standardized and comprehensive benchmark for multimodal game agents.}
    \gameworld{} provides \textbf{34 browser games} spanning \textbf{5 genres} and \textbf{170 tasks}. It supports \textbf{both} \emph{Computer-Use Agents} and \emph{Generalist Multimodal Agents} under a shared executable action space via deterministic \emph{Semantic Action Parsing}, together with a sandbox that decouples inference latency from gameplay, enabling standardized evaluation across different control interfaces.

    \item \textbf{A universal outcome-based state-verifiable evaluator.}
    Unlike prior game benchmarks that rely on noisy visual heuristics or VLM-as-judge pipelines, \gameworld{} evaluates entirely through outcome-based metrics computed from serialized \texttt{gameAPI} state. We compute deterministic task success and normalized progress directly from task-relevant game variables, ensuring noise-free and fully reproducible evaluation.

    \item \textbf{A suite of interface-aware benchmark analyses.}
    \gameworld{} contributes repeated-evaluation robustness studies to characterize the reproducibility of the benchmark itself. It further provides capability-aligned curriculum analyses, the real-time benchmark variant \gameworldrt{}, context-memory sensitivity analysis, and action-validity diagnostics to study latency coupling, capability bottlenecks, context-memory trade-offs, and instruction-following reliability across both game-agent interfaces.
\end{itemize}

\begin{figure}[!t]
    \centering
    \includegraphics[width=\linewidth]{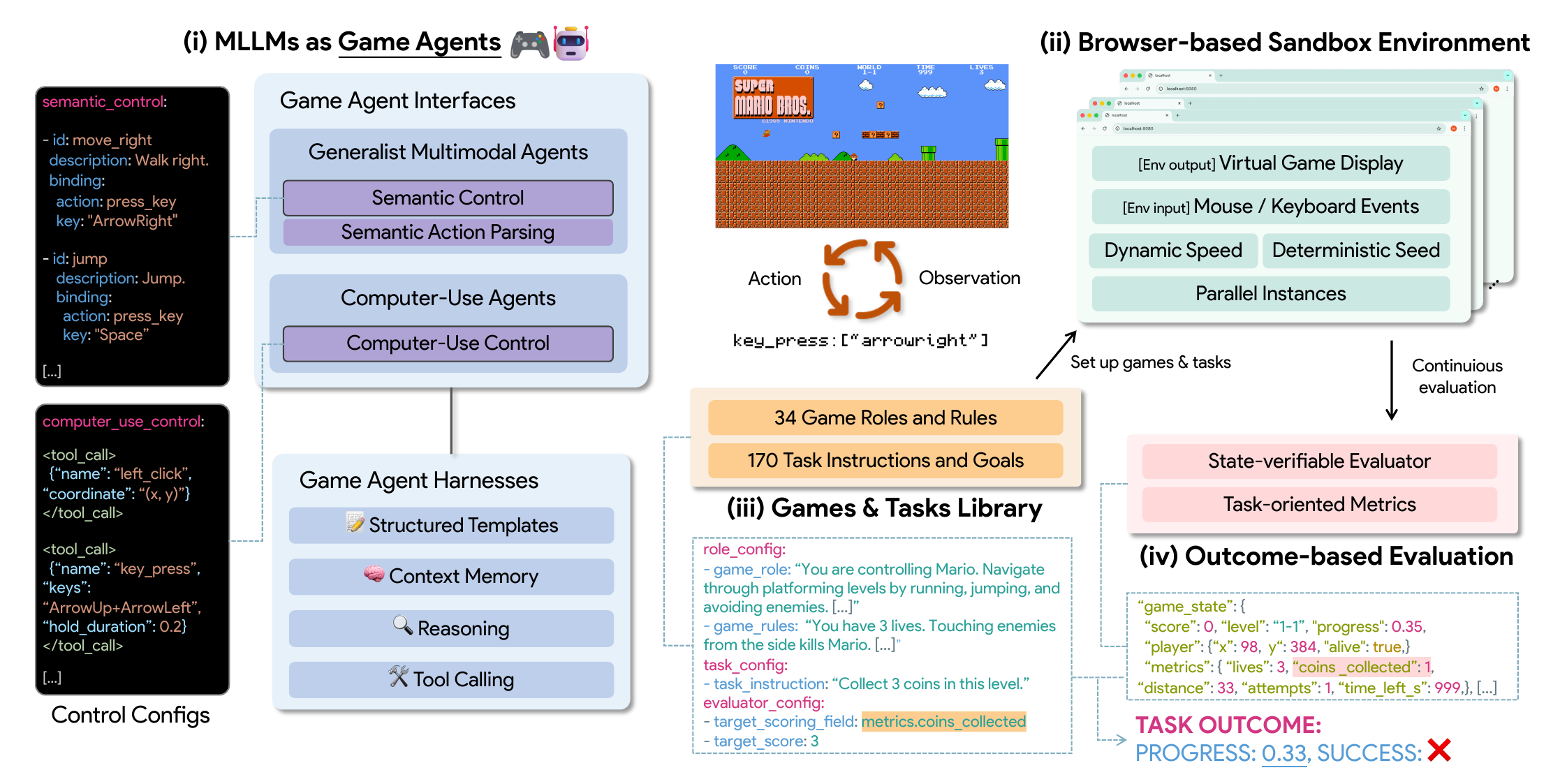}
    \caption{
    Overview of the \gameworld{} benchmark with four modules:
    (i) \textbf{MLLMs as game agents},
    (ii) \textbf{Browser-based sandbox environment},
    (iii) \textbf{Games \& tasks library},
    and (iv) \textbf{Outcome-based state-verifiable evaluation}. This closes a continuous and interactive observation-action-verification loop for systematically evaluating game agents.
    }
    \label{fig:overview}
\end{figure}

\section{Game Agent}
\label{sec:game_agent}

\newcommand{\cmark}{\ding{51}}
\newcommand{\xmark}{\ding{55}}

\definecolor{niceblue}{HTML}{E7F0F8}
\definecolor{niceorange}{HTML}{FBECDD}
\definecolor{nicered}{HTML}{FAEBED}

\newlength{\actionwidth}
\settowidth{\actionwidth}{{\scriptsize\ttfamily drag((x1,y1),(x2,y2))\quad}}
\newcommand{\action}[2]{%
    \makebox[\actionwidth][l]{\texttt{~-~#1}}%
    \textcolor{textgray}{\texttt{\qquad\qquad\#~#2}}%
}

\newlength{\gameactionwidth}
\settowidth{\gameactionwidth}{{\scriptsize\ttfamily move\_backward()\quad}}
\newcommand{\gameaction}[2]{%
    \makebox[\gameactionwidth][l]{\texttt{~-~#1}}%
    \textcolor{textgray}{\texttt{\qquad\quad\#~#2}}%
}

\begin{table}[t]
    \centering
    \scriptsize
    \setlength{\tabcolsep}{4pt}
    \begin{tabular}{p{0.44\linewidth}p{0.44\linewidth}}
    \toprule
    \rowcolor{niceblue}
    \multicolumn{2}{c}{\textbf{Game Agent Interfaces}} \\
    \midrule
    \rowcolor{niceblue}
    \multicolumn{1}{>{\centering\arraybackslash}p{0.44\linewidth}|}{\textbf{Computer-Use Agent}} &
    \multicolumn{1}{>{\centering\arraybackslash}p{0.44\linewidth}}{\textbf{Generalist Multimodal Agent}} \\
    \midrule
    \multicolumn{1}{p{0.44\linewidth}|}{
    \textbf{Action Space: Computer-use function calls} \newline
    \textcolor{textgray}{Native tools of mouse and keyboard events.} \newline
    \action{mouse\_move(x,y)}{move pointer} \newline
    \action{left\_click(x,y)}{primary click} \newline
    \action{right\_click(x,y)}{secondary click} \newline
    \action{double\_click(x,y)}{double click} \newline
    \action{click\_hold(x,y)}{triple click} \newline
    \action{drag((x1,y1),(x2,y2))}{drag pointer} \newline
    \action{scroll\_up(n)}{scroll up} \newline
    \action{scroll\_down(n)}{scroll down} \newline
    \action{type(text)}{text entry} \newline
    \action{press\_key(key)}{single key} \newline
    \action{press\_keys(key1,key2)}{key combo} \newline
    \action{wait(duration)}{idle}
    } &
    \multicolumn{1}{p{0.44\linewidth}}{
    \textbf{Action Space: Game-specific function calls} \newline
    \textcolor{textgray}{Semantic functions parsed into low-level controls.} \newline
    \gameaction{move\_forward()}{player: forward} \newline
    \gameaction{move\_backward()}{player: backward} \newline
    \gameaction{move\_left()}{player: strafe left} \newline
    \gameaction{move\_right()}{player: strafe right} \newline
    \gameaction{look\_up()}{view: camera up} \newline
    \gameaction{look\_down()}{view: camera down} \newline
    \gameaction{look\_left()}{view: camera turn left} \newline
    \gameaction{look\_right()}{view: camera turn right} \newline
    \gameaction{action\_jump()}{action: player jump} \newline
    \gameaction{action\_duck()}{action: player duck} \newline
    \gameaction{weapon\_fire()}{item: fire weapon} \newline
    \gameaction{no\_op()}{idle: no operation}
    } \\
    \midrule
    \rowcolor{niceblue}
    \multicolumn{2}{c}{\textbf{Unified Control Space (Atomic Events)}} \\
    \multicolumn{2}{c}{
        \parbox[t]{0.80\linewidth}{
            \hspace*{1cm}\begin{tabular}[t]{@{}l l@{}}
                \textcolor{textgray}{Mouse:} & \texttt{mouse\_move(x,y)} \quad \texttt{mouse\_down(button)} \quad \texttt{mouse\_up(button)} \quad \texttt{scroll(amount)} \\
                \textcolor{textgray}{Keyboard:} & \texttt{key\_down(key)} \quad \texttt{key\_up(key)} \\
                \textcolor{textgray}{Others:} & \texttt{wait(duration)} \quad \texttt{idle()} 
            \end{tabular}
        }
    } \\
    \bottomrule
    \end{tabular}
    \caption{
    \captfont{
    Two game agent interfaces and action-space taxonomy. Both interfaces are normalized to a unified control space of atomic human-computer interaction events.
    }
    }
    \label{tab:agent_interfaces}
\end{table}

A central challenge in benchmarking MLLMs as game agents is that models generate the actions in multiple forms to interact with the games. Even for the same operation, tool-call functions can be vary: a screen click becomes \texttt{left\_click(x,y)} in one API and \texttt{computer(\allowbreak{}action="click",\allowbreak{}coordinate=[x,y])} in another. Models also differ in abstraction because of the agent implementation, with some emitting raw keyboard and mouse controls while others reason in terms of high-level game actions. To standardize evaluation across these heterogeneous interfaces, we define two agent interfaces, Computer-Use and Generalist (Figure~\ref{fig:overview}, Module~i), and normalize all outputs into a shared executable action space defined over atomic human-computer interaction events.

\subsection{Agent Interfaces}

At each step, the agent observes a screenshot of the current game state, produces an action through the model, and the environment executes it. A verifiable evaluator then checks the resulting state against the task objective. The agent's raw output is normalized into a shared set of executable atomic events: \texttt{mouse\_move}, \texttt{mouse\_down}, \texttt{mouse\_up}, \texttt{key\_down}, \texttt{key\_up}, \texttt{scroll}, and \texttt{wait}. These events define the executor-level unified control space. Each game role exposes only the subset needed for that environment while preserving a common runtime contract across models.

As shown in Table~\ref{tab:agent_interfaces}, we distinguish two game-agent interfaces: \textbf{(i)} \textbf{Computer-Use Agents} that directly emit low-level keyboard and mouse controls (Section~\ref{sec:computer_use_agents}), and \textbf{(ii)} \textbf{Generalist Multimodal Agents} that act in a semantic space and are executed through deterministic \emph{Semantic Action Parsing} (Section~\ref{sec:generalist_agents}). 
To comprehensively assess the current landscape of game agents, we evaluate both state-of-the-art proprietary models and open-source models. Models with native computer-use capabilities are evaluated under both CUA and Generalist interfaces. This shared protocol also enables interface-aware analyses of benchmark robustness, real-time interaction, context-memory sensitivity, and action validity under one common runtime and verifier.

\subsection{Computer-Use Agents: Low-Level Controls}
\label{sec:computer_use_agents}
Computer-Use Agents (CUAs) directly emit low-level keyboard and mouse interactions such as \texttt{mouse\_move(x,y)}, \texttt{left\_click(x,y)}, and \texttt{press\_key(key)}, bearing full responsibility for both strategic decision-making and precise action grounding. These commands are executed under the same unified runtime contract and grounded into the shared executor-level event space. Since CUAs must output exact coordinates and key sequences from visual observations, this interface most closely mirrors how a human player interacts with the game, and is therefore highly sensitive to inference latency in real-time settings.

We enforce a \emph{one-action-per-step} constraint for evaluation consistency over CUAs: each model response must contain exactly one executable action that satisfies the role-specific keyboard or mouse control specification. Note that key combinations are allowed. Actions that fall outside the game's permitted control interface (e.g., OS-level APIs) are rejected, ensuring that CUA scores reflect in-game capabilities under a fixed action budget.

\subsection{Generalist Agents: Semantic Action Parsing}
\label{sec:generalist_agents}

Generalist Multimodal Agents excel at semantic planning but typically lack the ability to produce precise pixel coordinates or fine-grained key-timing sequences required for direct game control. To place them under the same benchmark runtime and verifier as CUAs, we introduce \emph{Semantic Action Parsing}: for each game and role, a deterministic parser maps every semantic action to a fixed low-level interaction command under the same unified runtime contract. Because this mapping is deterministic, it removes parser-side stochasticity and supports more interpretable interface-conditioned comparisons under the same executor-level physical event space. 
We further enforce \emph{Action Atomicity} at the model-step level: each model response must specify one interaction command per step. What is disallowed is any multi-command macro that bundles several semantically distinct decisions into one step.

\subsection{Agent Harnesses}

Foundation models alone are insufficient for sustained gameplay: the agent needs structured prompts, short and long-term memory, and model-specific tool interfaces to act coherently over long horizons~\cite{hu2025lmgamebench,yao2022react,schick2023toolformer}. Therefore, we wrap each model in a shared agent harness that standardizes these components across all models.
Appendix~\ref{sec:app_agent_details} further details the harness components used in our implementation.

\subsubsection{Structured Prompt}
To reduce prompt-induced variance across models and games, we define a fixed prompt template with four components: \texttt{\#Game Rules}, \texttt{\#Role and Controls}, \texttt{\#Task Instruction}, and \texttt{\#Output Format}. The template structure stays constant across all experiments; only the game-specific rules, role description, and task objective are swapped per configuration, keeping cross-model comparisons controlled.
The exact shared templates, per-game prompt blocks, and model-specific output-format blocks are listed in Appendices~\ref{sec:app_prompt_templates}, \ref{sec:app_prompt_library}, and \ref{sec:app_output_formats}.

\subsubsection{Context Memory}
The agent maintains a rolling memory module that stores the most recent rounds of interaction. Each round records the sequence \texttt{user\_prompt $\rightarrow$ screenshot $\rightarrow$ reasoning $\rightarrow$ action}, and recent rounds are prepended as an \texttt{Action History} block before the current observation. This gives the agent short-horizon trajectory context, allowing it to avoid repeating failed actions and to maintain consistency across consecutive steps. Our experiments on the effect of context memory on the performance of game agents are in Section~\ref{sec:memory_ablation}.
\subsubsection{Reasoning}
Reasoning is becoming increasingly important for agent capabilities, especially on long-horizon tasks where the agent must maintain subgoals rather than react frame by frame. This is also particularly relevant for visual reasoning: more deliberate inspection of visual inputs helps MLLMs parse visual information more reliably. Tool-assisted operations such as image zooming or cropping can further improve environment understanding with close observations. In video games, such text or visual reasoning is often necessary to support accurate perception and decision making. However, the longer reasoning time also introduces additional latency, which can be detrimental to the performance of game agents (See Section~\ref{sec:real_time}).

\subsubsection{Customized Function Calling}
We register the game's semantic actions and computer-use primitives as callable tools for each model, using each model provider's native function-calling (also known as tool-calling) interface (e.g., OpenAI function calling, Claude tool use, Gemini function declarations). This preserves each model's native agentic capability within its own API contract for the best performance, while keeping the harness-level protocol uniform across all models.
Appendices~\ref{sec:app_action_validation} and \ref{sec:app_semantic_action_parsing} provide the exact legality checks and deterministic action-resolution rules used by the runtime.

\definecolor{tabletwored}{HTML}{F02F1D}
\definecolor{tabletwogreen}{HTML}{3CA52C}
\newcommand{\tabletwocmark}{\textcolor{tabletwogreen}{\ding{51}}}
\newcommand{\tabletwoxmark}{\textcolor{tabletwored}{\ding{55}}}

\begin{table}[!t]
\centering
\caption{
\captfont{
Comparison with representative game or computer-use agent benchmarks.
\#Tasks denotes the number of \emph{specific instructions, goals, or questions}.
}
}
\scriptsize
\setlength{\tabcolsep}{2pt}
\renewcommand{\arraystretch}{1.2}
\resizebox{\textwidth}{!}{
\begin{tabular}{lccccccccl}
\toprule
\rowcolor{headerblue}
\textbf{Benchmark} &
\textbf{\makecell{\#\\Games}} &
\textbf{\makecell{\#\\Tasks}} &
\textbf{\makecell{\#\\Models}} &
\textbf{\makecell{Vision-\\Centric}} &
\textbf{\makecell{Config.\\Init. State}} &
\textbf{\makecell{Task-\\Oriented}} &
\textbf{\makecell{Parallel\\Inst.}} &
\textbf{\makecell{Verif.\\Eval.}} &
\textbf{Notes} \\
\midrule
\rowcolor{bggray}
\multicolumn{10}{c}{\textcolor{textgray}{\textbf{\textit{Static Benchmarks}}}} \\
\midrule
GameQA~\cite{tong2025gamerl} & 30 & 158 & 8 & \tabletwoxmark & NA & \tabletwoxmark & NA & \tabletwoxmark & Code2Logic QA. \\
VideoGameQA~\cite{taesiri2025videogameqa} & 800+ & 9 & 16 & \tabletwocmark & NA & \tabletwoxmark & NA & \tabletwoxmark & Non-interactive QA dataset. \\
\midrule
\rowcolor{bggray}
\multicolumn{10}{c}{\textcolor{textgray}{\textbf{\textit{Interactive Benchmarks}}}} \\
\midrule
MCU~\cite{zheng2025mcu} & 1 & 150 & 4 & \tabletwocmark & \tabletwocmark & \tabletwocmark & \tabletwoxmark & \tabletwoxmark & Minecraft only. \\
LMGame-Bench~\cite{hu2025lmgamebench} & 6 & 6 & 13 & \tabletwoxmark & \tabletwoxmark & \tabletwocmark & \tabletwoxmark & \tabletwocmark & Text-centric benchmark. \\
VideoGame-Bench~\cite{zhang2025videogamebench} & 23 & 23 & 5 & \tabletwocmark & \tabletwoxmark & \tabletwocmark & \tabletwoxmark & \tabletwoxmark & Heuristics evaluation. \\
FlashAdventure~\cite{ahn2025flashadventure} & 34 & 34 & 7 & \tabletwocmark & \tabletwoxmark & \tabletwocmark & \tabletwoxmark & \tabletwocmark & Flash-based stories; CUA-as-a-Judge. \\
V-MAGE~\cite{zheng2025v} & 5 & 30 & 7 & \tabletwocmark & \tabletwocmark & \tabletwocmark & \tabletwoxmark & \tabletwoxmark & 5 games only. \\
BALROG~\cite{paglieri2025balrog} & 6 & 48 & 12 & \tabletwoxmark & \tabletwoxmark & \tabletwocmark & \tabletwocmark & \tabletwocmark & Visual input degrades performance. \\
NitroGen~\cite{magne2025nitrogen} & 10 & 30 & 1 & \tabletwocmark & \tabletwoxmark & \tabletwoxmark & \tabletwoxmark & \tabletwocmark & No language-conditioned tasks. \\
Orak~\cite{park2025orak} & 12 & 12 & 15 & \tabletwoxmark & \tabletwoxmark & \tabletwocmark & \tabletwoxmark & \tabletwocmark & States pre-processed into text. \\
GameVerse~\cite{zhang2026gameverse} & 15 & 15 & 7 & \tabletwocmark & \tabletwoxmark & \tabletwocmark & \tabletwoxmark & \tabletwoxmark & Semantic + GUI control. \\
\midrule
\textbf{\gameworld{}} & \textbf{34} & \textbf{170} & \textbf{18} & \tabletwocmark & \tabletwocmark & \tabletwocmark & \tabletwocmark & \tabletwocmark & \textbf{Scalable tasks, state-verifiable evaluation.} \\
\bottomrule
\end{tabular}
}
\label{tab:bench_comparison}
\end{table}

\section{\gameworld{} Benchmark}
\label{sec:benchmark}

\begin{table}[!t]
    \centering
    \scriptsize
    \setlength{\tabcolsep}{5pt}
    \caption{
        \captfont{Game genre of the \gameworld{} benchmark. Each row includes one representative screenshot, the dominant interaction mechanics, and the game IDs used in our evaluation.}
    }
    \begin{tabular}{p{0.14\linewidth}p{0.24\linewidth}p{0.23\linewidth}p{0.26\linewidth}}
    \toprule
    \rowcolor{headerblue}
    \multicolumn{1}{p{0.14\linewidth}|}{\textbf{Game Genre}} &
    \multicolumn{1}{p{0.24\linewidth}|}{\textbf{Example Screenshot}} &
    \multicolumn{1}{p{0.23\linewidth}|}{\textbf{Key Mechanics}} &
    \multicolumn{1}{p{0.24\linewidth}}{\textbf{Games}} \\
    \midrule
    \multicolumn{1}{p{0.14\linewidth}|}{\vspace*{8mm}\textbf{\makecell[c]{Arcade \\ (7)}}} &
    \multicolumn{1}{p{0.24\linewidth}|}{\vspace{-4pt}\includegraphics[width=0.98\linewidth]{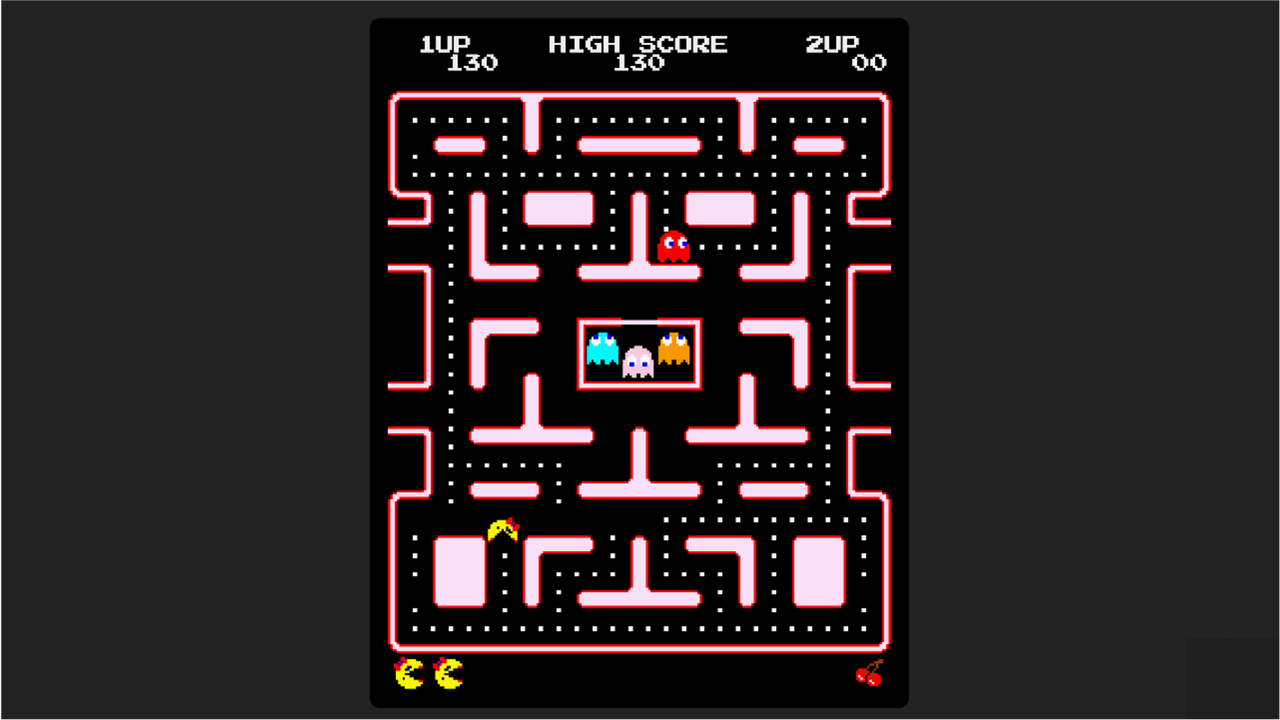} Source: \texttt{pac-man}} &
    \multicolumn{1}{p{0.23\linewidth}|}{\makecell[tl]{Fast-paced, closed-loop \\ control with \textbf{dynamic} \\ multi-entity tracking, \\ reactive evasion, and reward \\ collection.}} &
    \multicolumn{1}{p{0.24\linewidth}}{\makecell[tl]{
        \texttt{5-breakout} \\
        \texttt{8-core-ball} \\
        \texttt{15-google-snake} \\
        \texttt{23-pacman} \\
        \texttt{25-rocket-league-2d} \\
        \texttt{33-worlds-hardest-game} \\
        \texttt{34-worlds-hardest-game-2}
        } \vspace{-1mm}} \\
    \midrule
    \multicolumn{1}{p{0.14\linewidth}|}{\vspace*{9mm}\textbf{\makecell[c]{Platformer \\ (8)}}} &
    \multicolumn{1}{p{0.24\linewidth}|}{\vspace{-4pt}\includegraphics[width=0.98\linewidth]{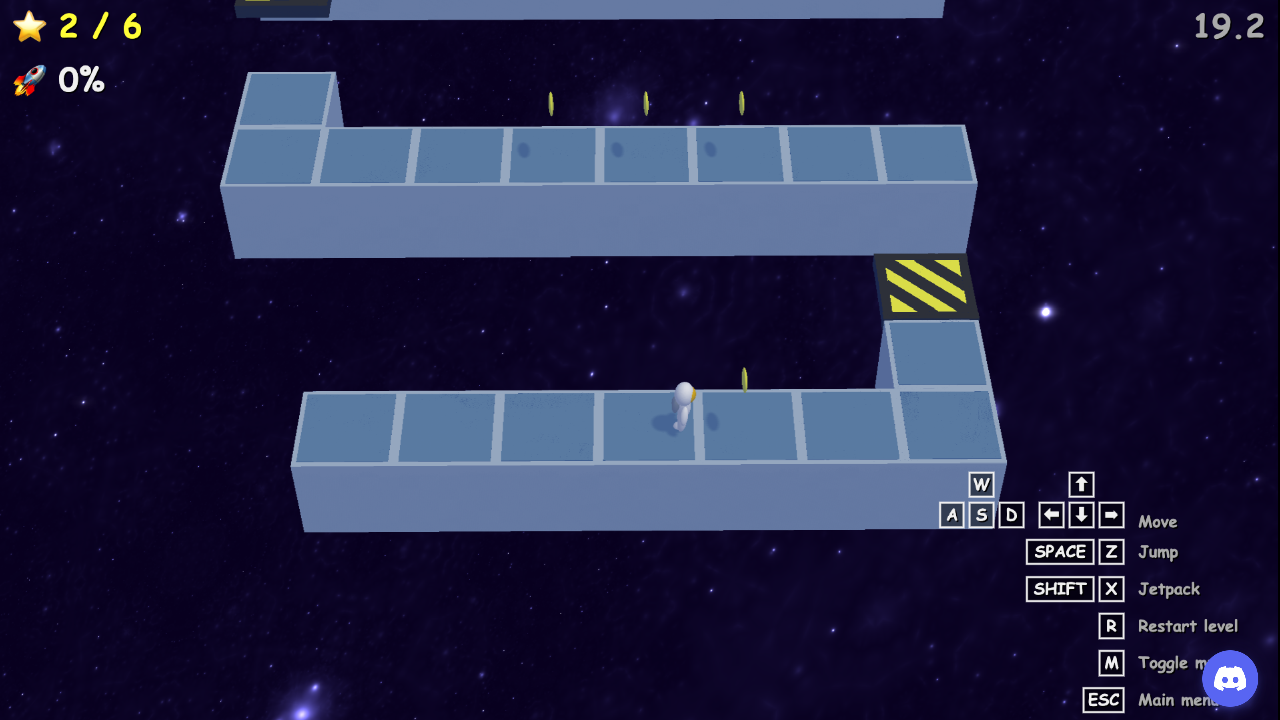} Source: \texttt{captaincallisto}} &
    \multicolumn{1}{p{0.23\linewidth}|}{\makecell[tl]{Spatiotemporal navigation  \\ demanding  \textbf{precise} \\ physics-based movement, \\ localized planning, and \\ hazard evasion across  \\ structured terrains.}} &
    \multicolumn{1}{p{0.24\linewidth}}{\makecell[tl]{
        \texttt{2-another-gentlemans-adventure} \\
        \texttt{6-captaincallisto} \\
        \texttt{10-doodle-jump} \\
        \texttt{14-geodash} \\
        \texttt{17-mario-game} \\
        \texttt{22-ovo} \\
        \texttt{24-restless-wing-syndrome} \\
        \texttt{30-vex-3}
    } \vspace{-1mm}} \\
    \midrule
    \multicolumn{1}{p{0.14\linewidth}|}{\vspace*{9mm}\textbf{\makecell[c]{Puzzle \\ (7)}}} &
    \multicolumn{1}{p{0.24\linewidth}|}{\vspace{-4pt}\includegraphics[width=0.98\linewidth]{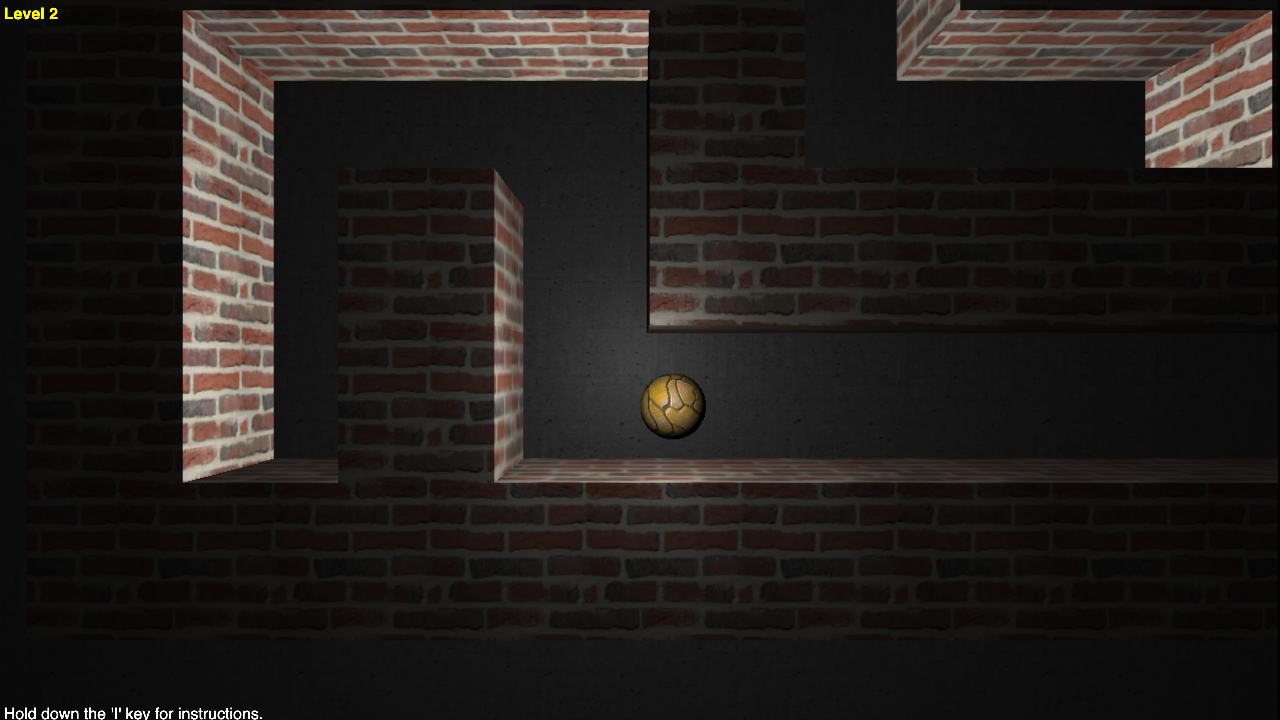} Source: \texttt{astray}} &
    \multicolumn{1}{p{0.23\linewidth}|}{\makecell[tl]{Discrete state-space \\ exploration focusing  \\ on \textbf{long-horizon}  strategic \\ planning and logical \\ decision-making.}} &
    \multicolumn{1}{p{0.24\linewidth}}{\makecell[tl]{
        \texttt{1-2048} \\
        \texttt{3-astray} \\
        \texttt{16-hextris} \\
        \texttt{19-minesweeper} \\
        \texttt{27-stack} \\
        \texttt{29-tetris} \\
        \texttt{32-wordle}
        } \vspace{-1mm}} \\
    \midrule
    \multicolumn{1}{p{0.14\linewidth}|}{\vspace*{9mm}\textbf{\makecell[c]{Runner \\ (8)}}} & 
    \multicolumn{1}{p{0.24\linewidth}|}{\vspace{-4pt}\includegraphics[width=0.98\linewidth]{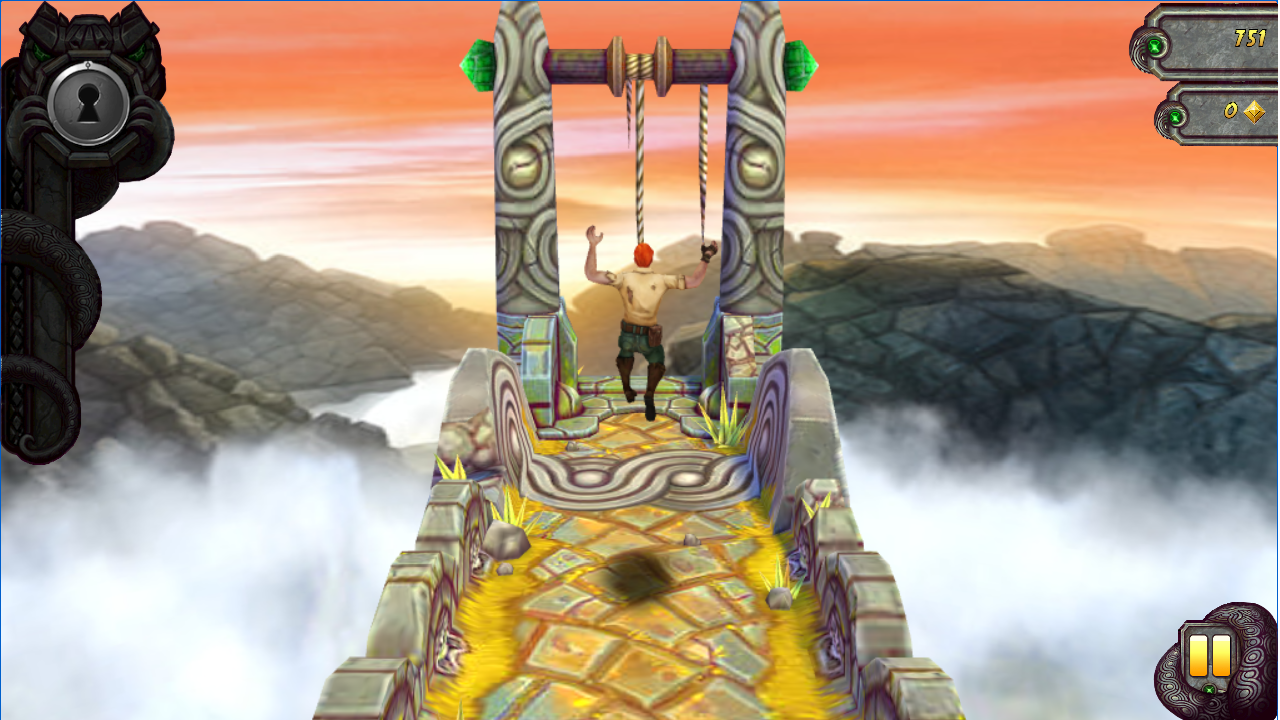} Source: \texttt{temple-run-2}} &
    \multicolumn{1}{p{0.23\linewidth}|}{\makecell[tl]{Continuous state progression \\ requiring \textbf{high-frequency} \\ reactive control and  precise \\ timing for obstacle avoidance.}} &
    \multicolumn{1}{p{0.24\linewidth}}{\makecell[tl]{
        \texttt{4-boxel-rebound} \\
        \texttt{7-chrome-dino} \\
        \texttt{9-cubefield} \\
        \texttt{11-edge-surf} \\
        \texttt{13-flappy-bird} \\
        \texttt{21-ns-shaft} \\
        \texttt{26-run-3} \\
        \texttt{28-temple-run-2}
        } \vspace{0mm}} \\
    \midrule
    \multicolumn{1}{p{0.14\linewidth}|}{\vspace*{8mm}\textbf{\makecell[c]{Simulation \\ (4)}}} &
    \multicolumn{1}{p{0.24\linewidth}|}{\vspace{-4pt}\includegraphics[width=0.98\linewidth]{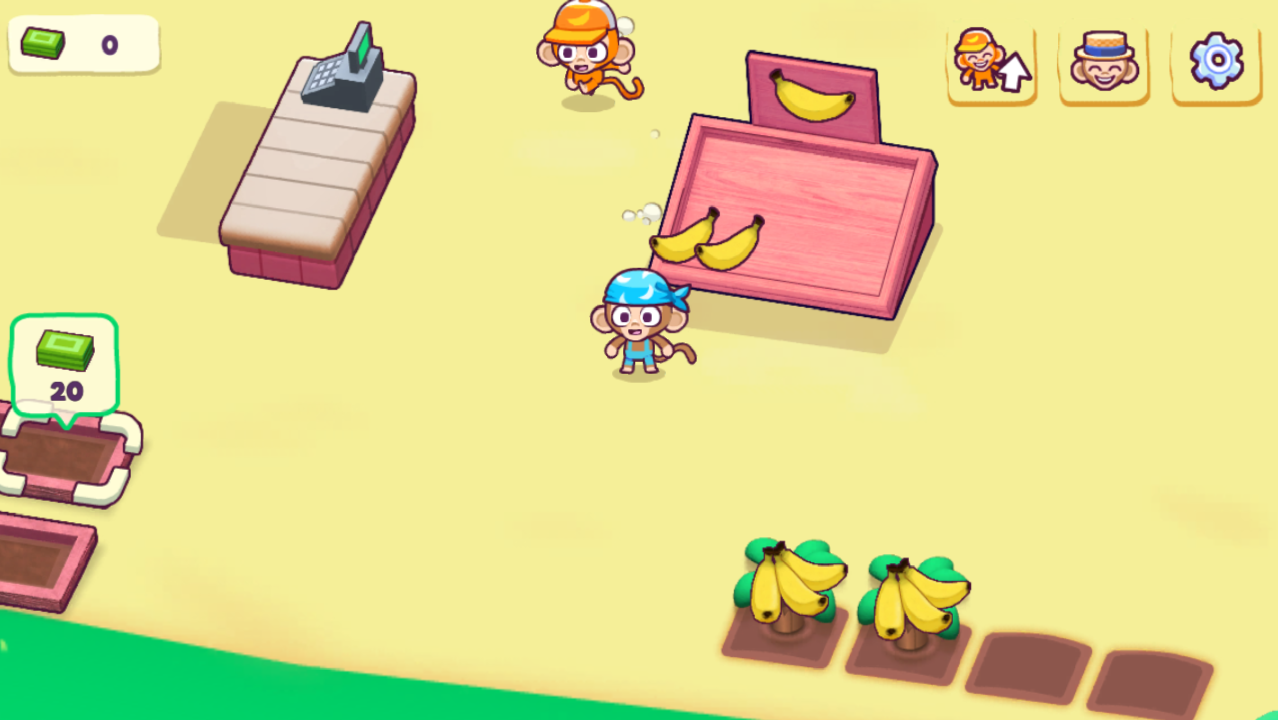} Source: \texttt{monkey-mart}} &
    \multicolumn{1}{p{0.23\linewidth}|}{\makecell[tl]{\textbf{Open-ended}, multi-objective  \\  environments evaluating \\ resource management, \\ multi-character cooperation,  \\ or strategic exploration.}} &
    \multicolumn{1}{p{0.24\linewidth}}{\makecell[tl]{
        \texttt{12-fireboy-and-watergirl} \\
        \texttt{18-minecraft-clone-glm} \\
        \texttt{20-monkey-mart} \\
        \texttt{31-wolf3d}
        } \vspace{-1mm}} \\
    \bottomrule
    \end{tabular}
    \label{tab:game_genres}
\vspace{-2mm}
\end{table}

\subsection{Benchmark Design}

Evaluating agents in games introduces challenges that existing game agent benchmarks have not fully addressed. Most cover few games within narrow genres, limiting the diversity and scale needed for comprehensive evaluation. In real-time games, agent inference latency directly affects outcomes: a two-second pause can mean the character has already fallen off the platform. Moreover, most existing benchmarks rely on heuristic, OCR or VLM-as-judge methods for evaluation, introducing noise that makes results difficult to verify.

\gameworld{} addresses each of these with: \textbf{(i)} a curated benchmark spanning five genres, with standardized task definitions including: task instruction, configurable initialization state, target metric, and evaluation configurations; \textbf{(ii)} a sandbox environment that manages game execution and decouples runtime latency from agent evaluation (Section~\ref{sec:sandbox}); and \textbf{(iii)} a state-verifiable evaluator that provides outcome-based metrics from serialized \texttt{gameAPI} state (Section~\ref{sec:eval}). 
Table~\ref{tab:bench_comparison} compares \gameworld{} with representative prior computer-use or video game benchmarks. 
Additional implementation details on preset composition, suite expansion, and runtime coordination are provided in Appendices~\ref{sec:app_runtime} and \ref{sec:app_loop}.

\subsection{Games and Tasks}

As shown in Table~\ref{tab:game_genres}, \gameworld{} comprises \textbf{34} browser-based games and \textbf{170} task instructions spanning five genres: \textbf{Runner}, \textbf{Arcade}, \textbf{Platformer}, \textbf{Puzzle}, and \textbf{Simulation}. The genres are selected to cover distinct capabilities that game agents must exhibit. Runner and Arcade games demand high-frequency reactive control and multi-entity tracking under continuous time pressure. Platformers require precise, physics-aware spatial navigation. Puzzles test logical reasoning and long-horizon planning in discrete state spaces. Simulations present open-ended, multi-objective environments involving resource management or 3D spatial reasoning.

Each task pairs a natural-language instruction with a quantitative target and a verifiable evaluator. Instructions are goal-oriented but open-ended in execution: the agent receives no intermediate guidance and must autonomously decide actions from visual observations within a fixed step budget. We define two complementary metrics: \textbf{Success Rate} ($\mathcal{SR} \in \{0,1\}$), the fraction of runs meeting the target, and \textbf{Progress} ($\mathcal{PG} \in [0,1]$), a normalized measure of how far the agent advanced toward the objective, providing partial credit for incomplete runs.

\subsection{Game Information}

Table~\ref{tab:game_info} lists the 34 games used in the \gameworld{} benchmark.
Beyond the genre-level taxonomy in Table~\ref{tab:game_genres}, this inventory provides a game-by-game view with IDs, source citations, short mechanic summaries, and representative screenshots. The collection are designed spanning a wide range of interaction structures, from sparse board-state reasoning in games such as \texttt{2048} and \texttt{Minesweeper}, to continuous real-time control in \texttt{Temple Run 2} and \texttt{Pac-Man}, as well as open-ended simulation in \texttt{Monkey Mart} and \texttt{Minecraft Clone}. This diversity is reflected not only in task mechanics but also in visual presentation, including 2D and 3D viewpoints, diverse HUDs, minimal puzzle layouts, and character-centric platforming scenes, which together motivate a unified benchmark interface across all games.

\newcommand{\gameshot}[1]{%
    {\setlength{\fboxsep}{0pt}\setlength{\fboxrule}{0.2pt}%
    \fbox{\includegraphics[width=1.0\linewidth,height=1.58cm,keepaspectratio]{#1}}}%
}

\newcommand{\gameinfotableheader}{
    \toprule
    \textbf{ID} & \textbf{Game} & \textbf{Description} & \textbf{Gameplay Screenshot} \\
    \midrule
}

\renewcommand{\arraystretch}{1.08}
\setlength{\tabcolsep}{5pt}

\begin{table}[!t]
    \centering
    \scriptsize
    \caption{Game inventory for the 34-game \gameworld{} benchmark.}
    \label{tab:game_info}
    \begin{tabular}{@{}p{0.18\linewidth}p{0.15\linewidth}p{0.38\linewidth}>{\centering\arraybackslash}m{0.18\linewidth}@{}}
    \gameinfotableheader
    1-2048 & \gamecite{2048}{game01-2048} & Sliding-tile puzzle where the player merges matching tiles to build larger values under limited board space. & \gameshot{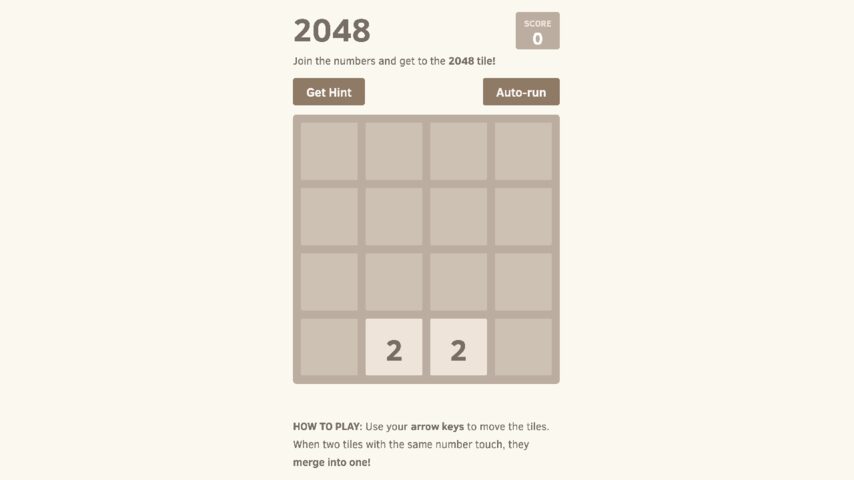} \\
    2-another-gentlemans-adventure & \gamecite{Another \newline Gentleman's \newline Adventure}{game02-another-gentlemans-adventure} & Platform adventure centered on movement, jumping, coin collection, and enemy avoidance. & \gameshot{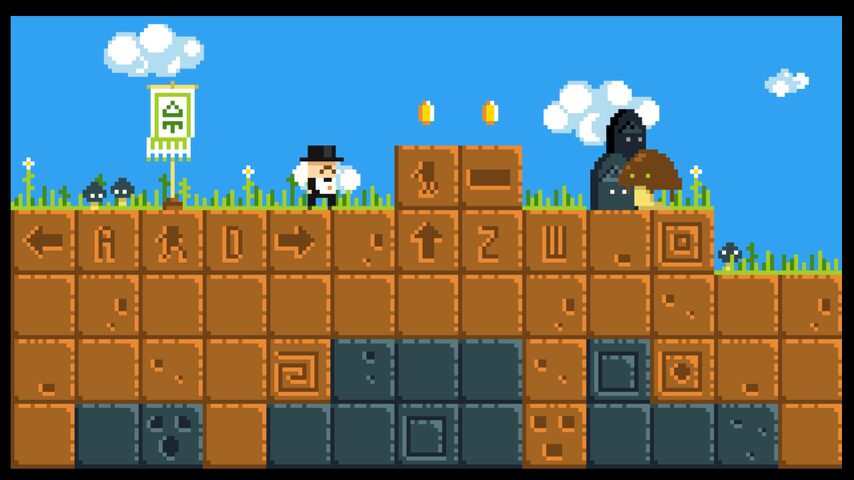} \\
    3-astray & \gamecite{Astray}{game03-astray} & Maze-navigation puzzle in which the player must steer through a labyrinth to find the exit. & \gameshot{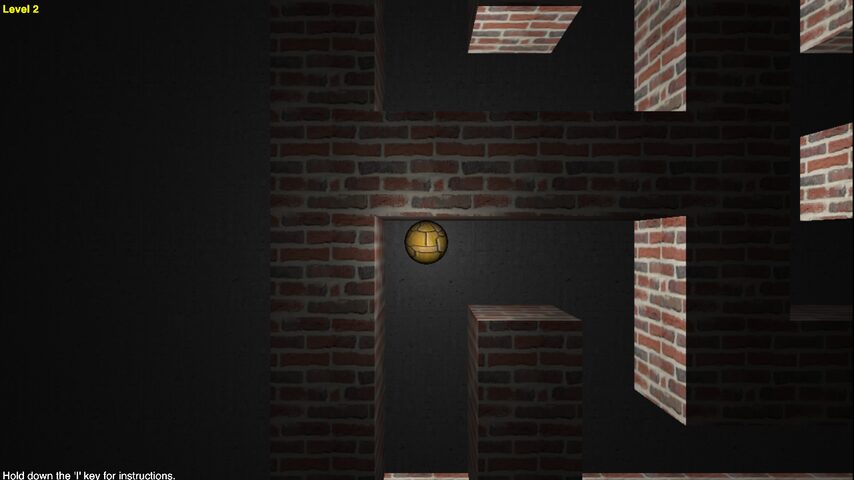} \\
    4-boxel-rebound & \gamecite{Boxel Rebound}{game04-boxel-rebound} & Precision auto-runner where the player times jumps to survive hazards and reach the end of each level. & \gameshot{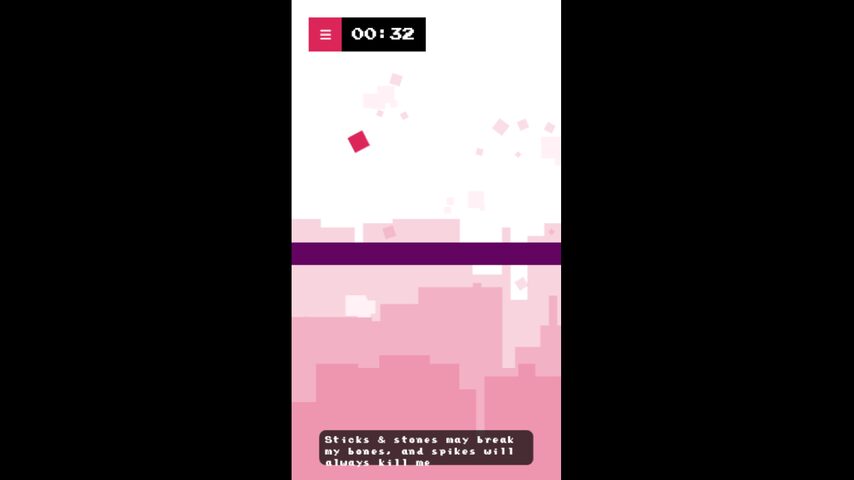} \\
    5-breakout & \gamecite{Breakout}{game05-breakout} & Classic brick-breaking arcade game where the player controls a paddle to keep the ball in play and clear bricks. & \gameshot{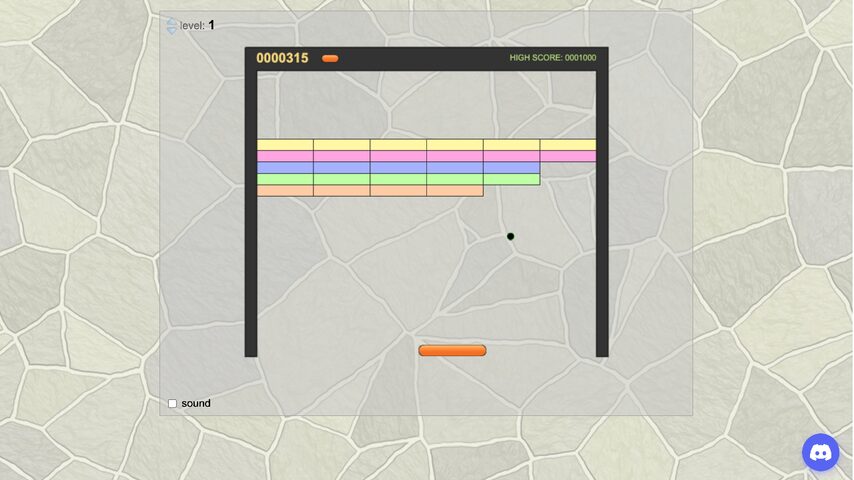} \\
    6-captaincallisto & \gamecite{Captain Callisto}{game06-captaincallisto} & Platform adventure with traversal, jumping, and jetpack-assisted movement toward the exit. & \gameshot{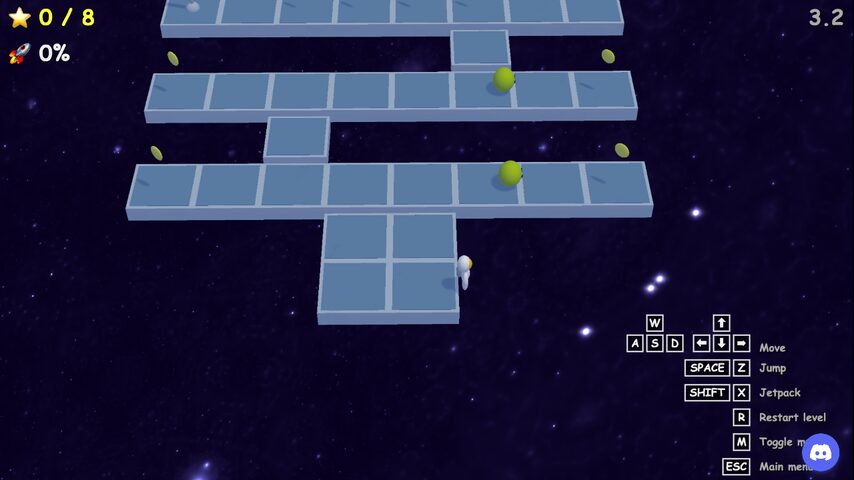} \\
    7-chrome-dino & \gamecite{Chrome Dino}{game07-chrome-dino} & Endless runner in which the dinosaur must jump over obstacles and stay alive as speed increases. & \gameshot{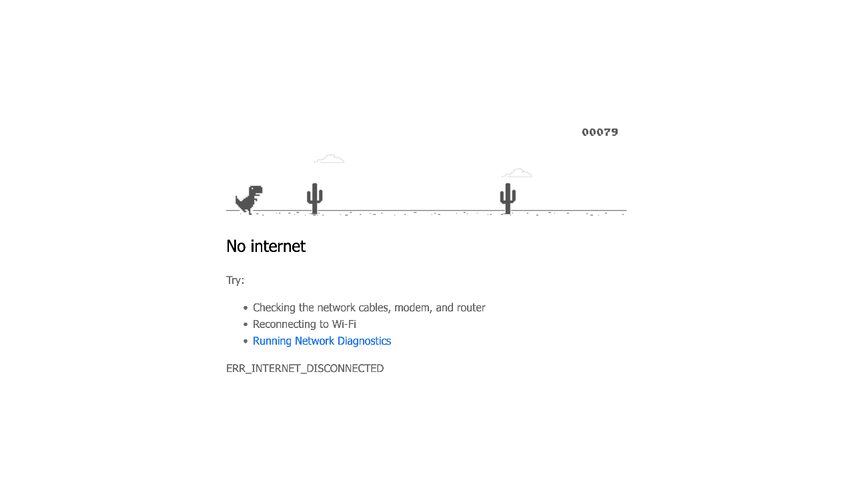} \\
    8-core-ball & \gamecite{Core Ball}{game08-core-ball} & Timing-based arcade game where numbered balls must be fired into a rotating core without collisions. & \gameshot{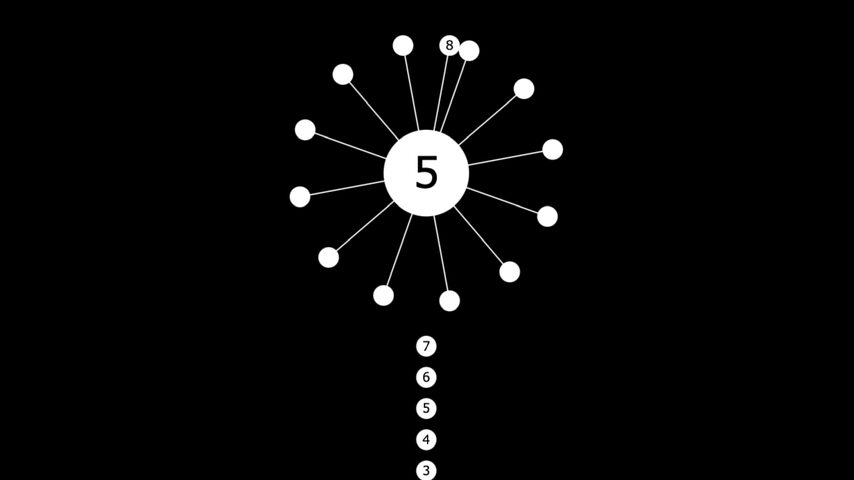} \\
    9-cubefield & \gamecite{Cubefield}{game09-cubefield} & Endless 3D runner where the player steers through dense cube fields and survives as long as possible. & \gameshot{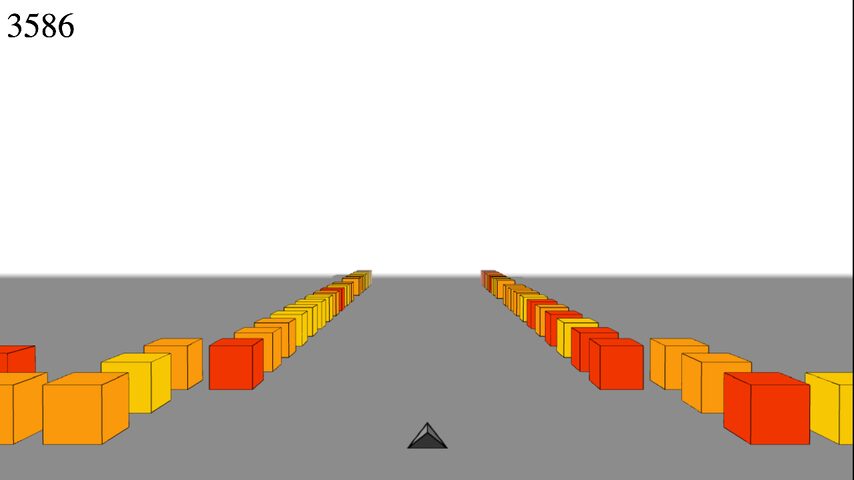} \\
    10-doodle-jump & \gamecite{Doodle Jump}{game10-doodle-jump} & Vertical platformer where the player chains landings to keep climbing through increasingly complex layouts. & \gameshot{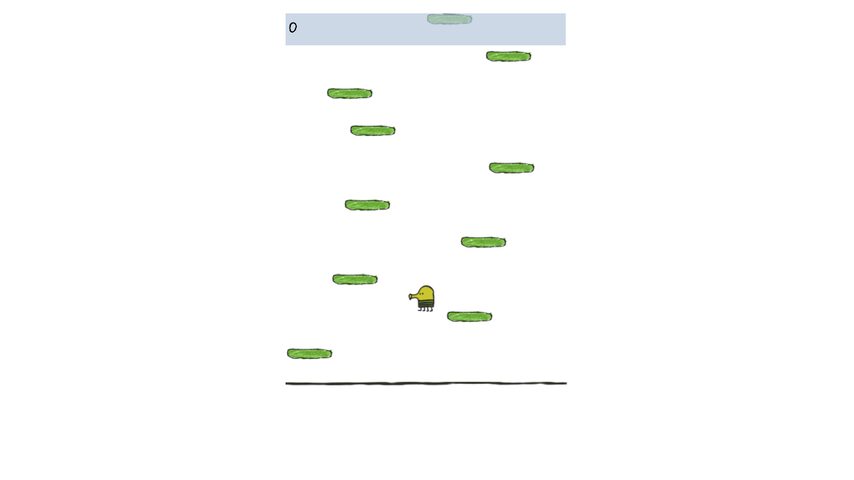} \\
    11-edge-surf & \gamecite{Edge Surf}{game11-edge-surf} & Surfing endless runner focused on obstacle avoidance, item collection, and survival over long distances. & \gameshot{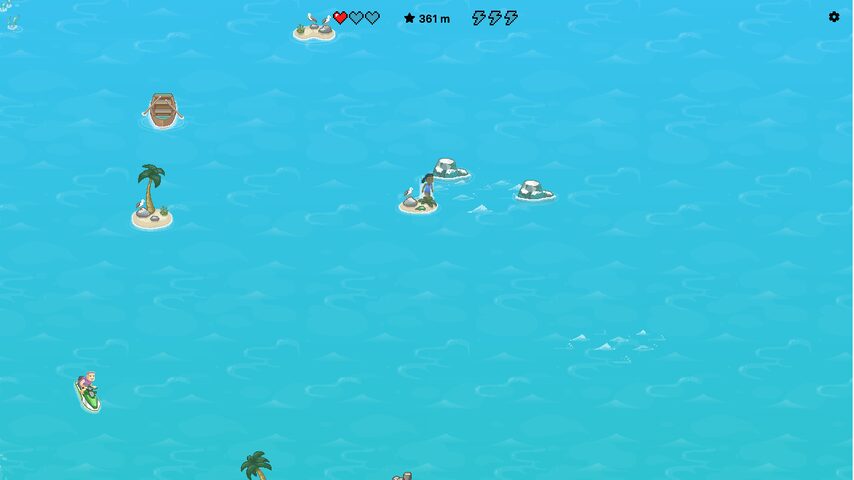} \\
    12-fireboy-and-watergirl & \gamecite{Fireboy and \newline Watergirl}{game12-fireboy-and-watergirl} & Cooperative puzzle-platformer where two characters with asymmetric constraints must coordinate to finish a level. & \gameshot{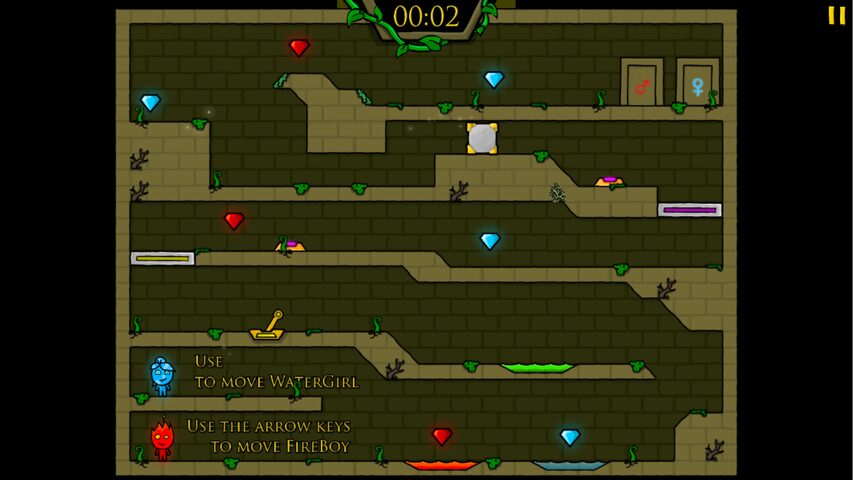} \\
    \bottomrule
    \end{tabular}
\end{table}

\begin{table}[!t]
    \centering
    \scriptsize
    \caption*{\textbf{Table~\ref{tab:game_info} (continued):} Game inventory for the 34-game \gameworld{} benchmark.}
    \begin{tabular}{@{}p{0.18\linewidth}p{0.15\linewidth}p{0.38\linewidth}>{\centering\arraybackslash}m{0.18\linewidth}@{}}
    \gameinfotableheader
    13-flappy-bird & \gamecite{Flappy Bird}{game13-flappy-bird} & One-button flying game that tests precise timing while weaving through pipes. & \gameshot{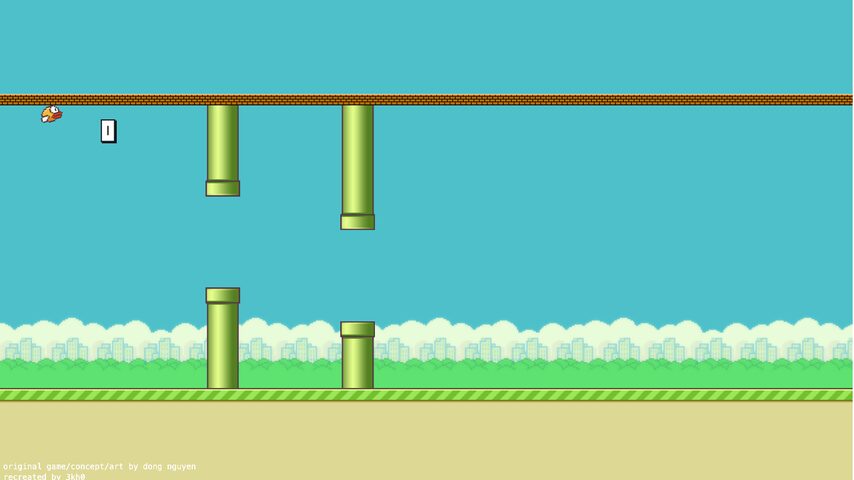} \\
    14-geodash & \gamecite{GeoDash}{game14-geodash} & Geometry-Dash-style auto-runner where success depends on tightly timed jumps over spikes and gaps. & \gameshot{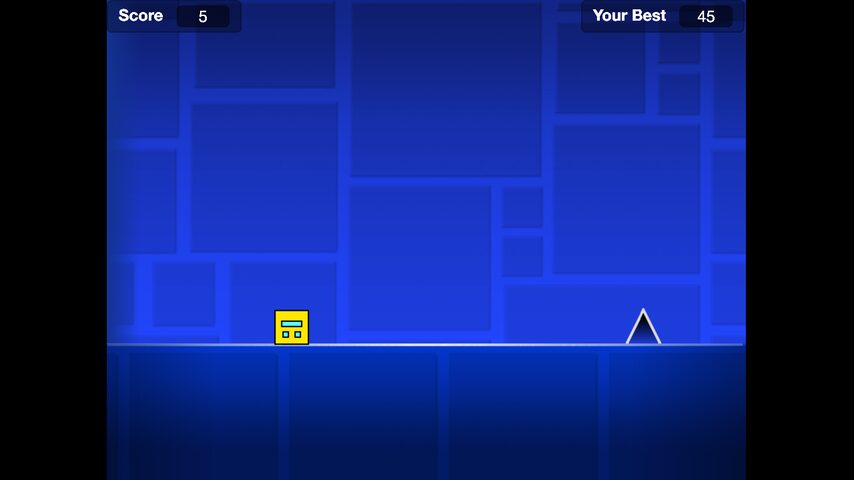} \\
    15-google-snake & \gamecite{Google Snake}{game15-google-snake} & Classic Snake variant where the agent grows by eating food while avoiding walls and self-collisions. & \gameshot{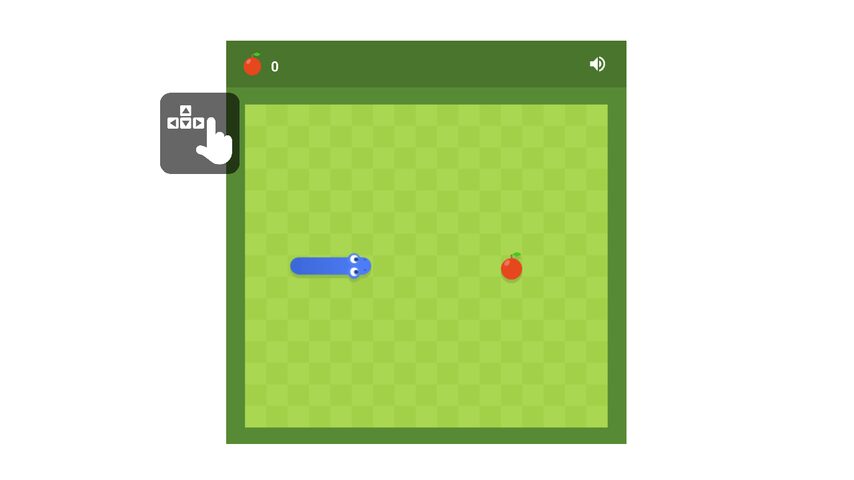} \\
    16-hextris & \gamecite{Hextris}{game16-hextris} & Hexagon-based matching puzzle where the agent rotates and places colored blocks to prevent overflow. & \gameshot{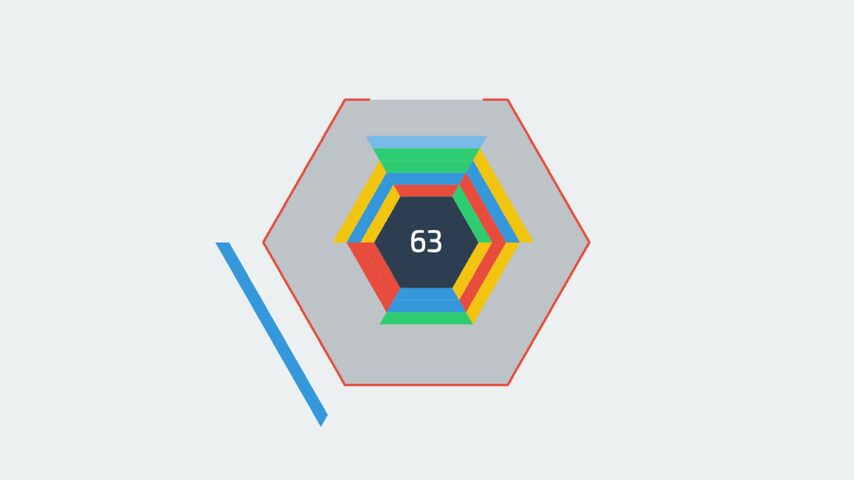} \\
    17-mario-game & \gamecite{Mario Game}{game17-mario-game} & Super-Mario-style platformer with enemy avoidance, jumping, and long-horizon navigation to the flagpole. & \gameshot{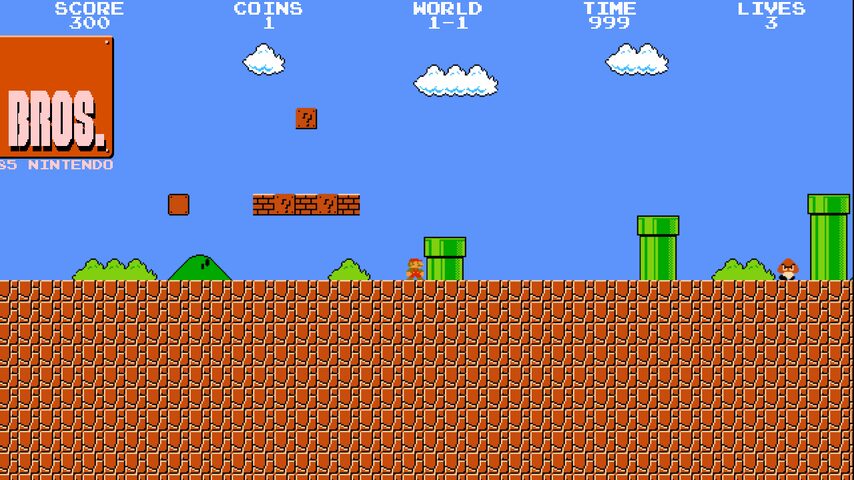} \\
    18-minecraft-clone-glm & \gamecite{Minecraft Clone}{game18-minecraft-clone-glm} & First-person sandbox game focused on movement, camera control, resource gathering, and direct world interaction. & \gameshot{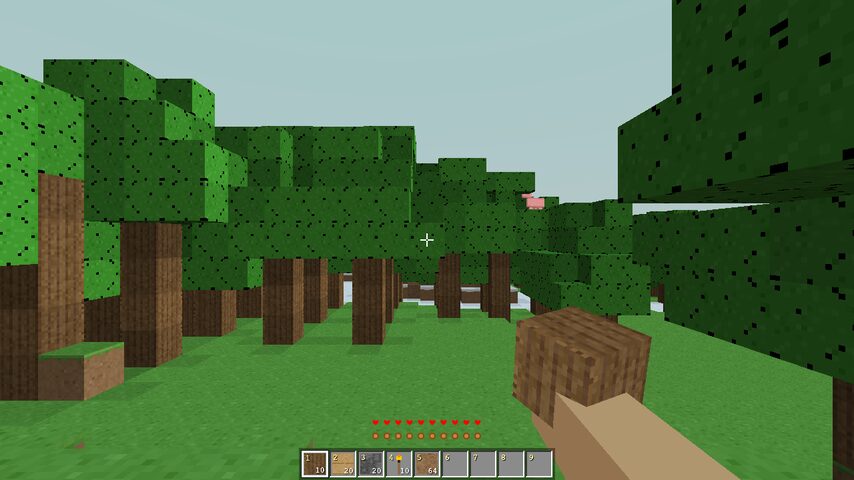} \\
    19-minesweeper & \gamecite{Minesweeper}{game19-minesweeper} & Logic puzzle that requires deducing mine locations from local numeric clues without triggering a mine. & \gameshot{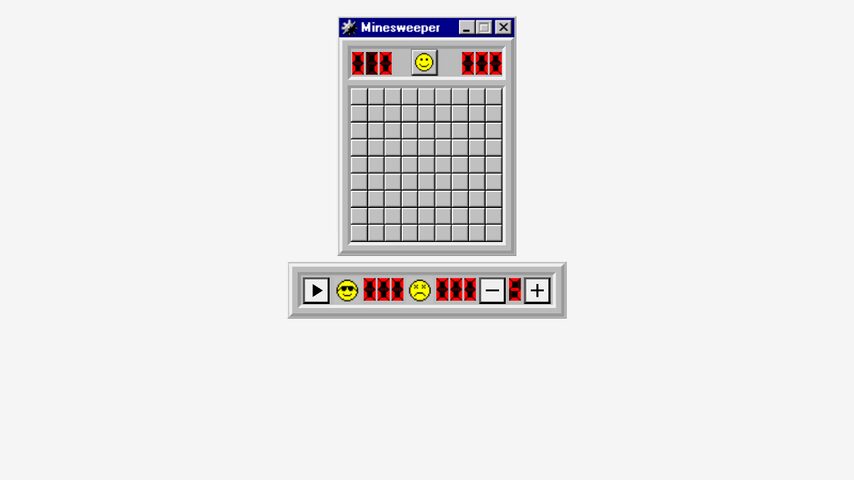} \\
    20-monkey-mart & \gamecite{Monkey Mart}{game20-monkey-mart} & Store-management simulation where the player harvests goods, stocks shelves, and serves customers efficiently. & \gameshot{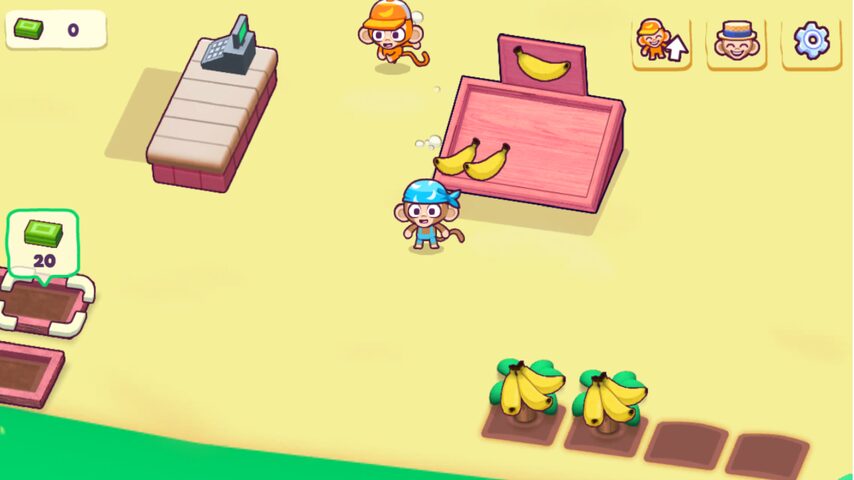} \\
    21-ns-shaft & \gamecite{NS-Shaft}{game21-ns-shaft} & Falling-platform runner in which the player descends through shifting platforms while avoiding hazards. & \gameshot{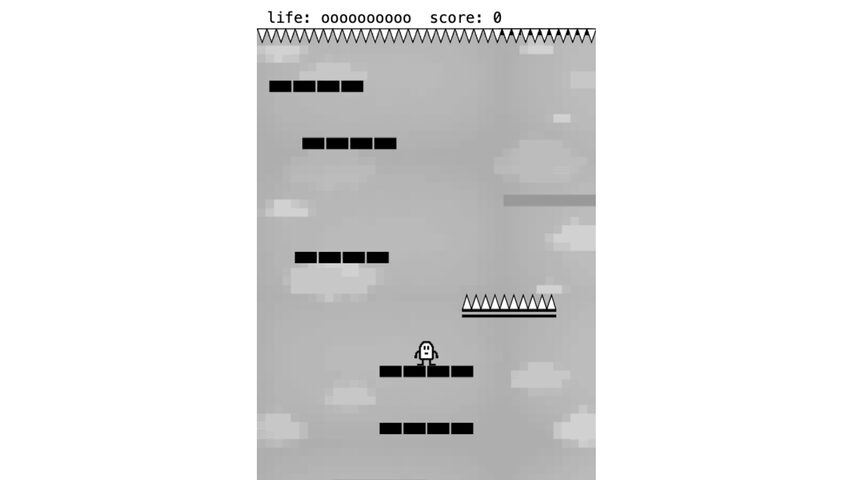} \\
    22-ovo & \gamecite{OvO}{game22-ovo} & Fast platformer with traps, wall interactions, and jump timing for level-by-level navigation. & \gameshot{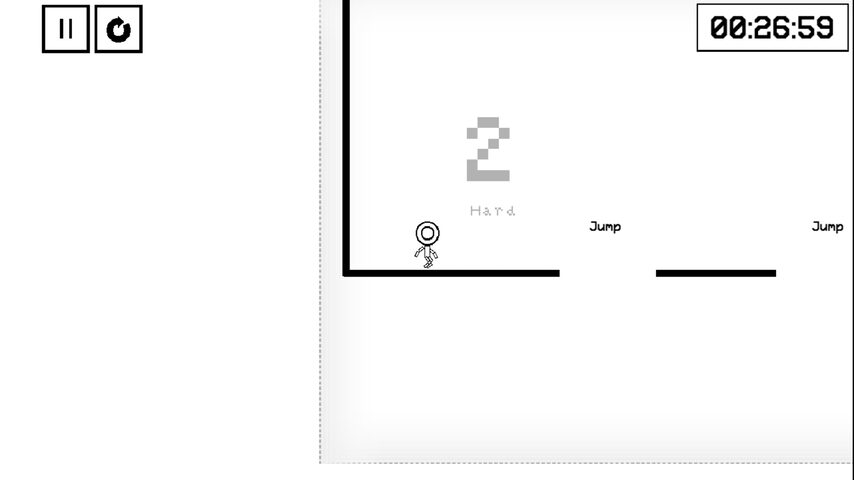} \\
    23-pacman & \gamecite{Pac-Man}{game23-pacman} & Maze-chase arcade game focused on pellet collection, ghost avoidance, and opportunistic ghost hunting. & \gameshot{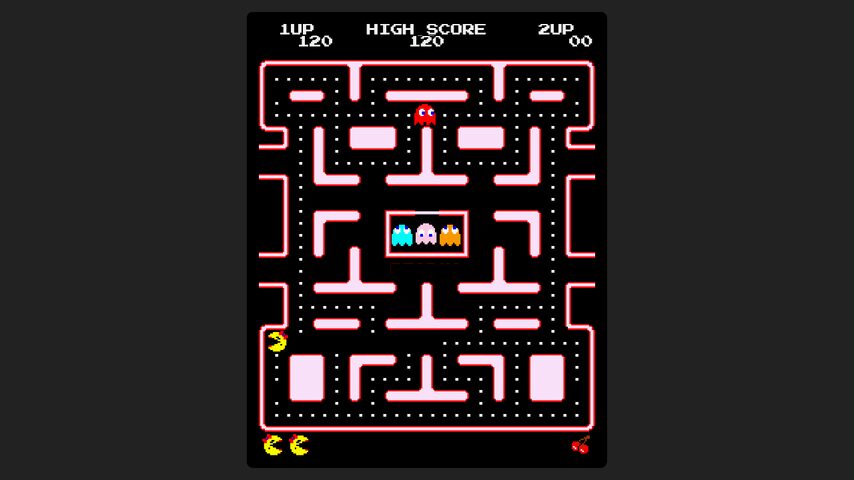} \\
    24-restless-wing-syndrome & \gamecite{Restless Wing \newline Syndrome}{game24-restless-wing-syndrome} & Platformer with periodic automatic flapping, requiring the player to work with a constrained movement rhythm. & \gameshot{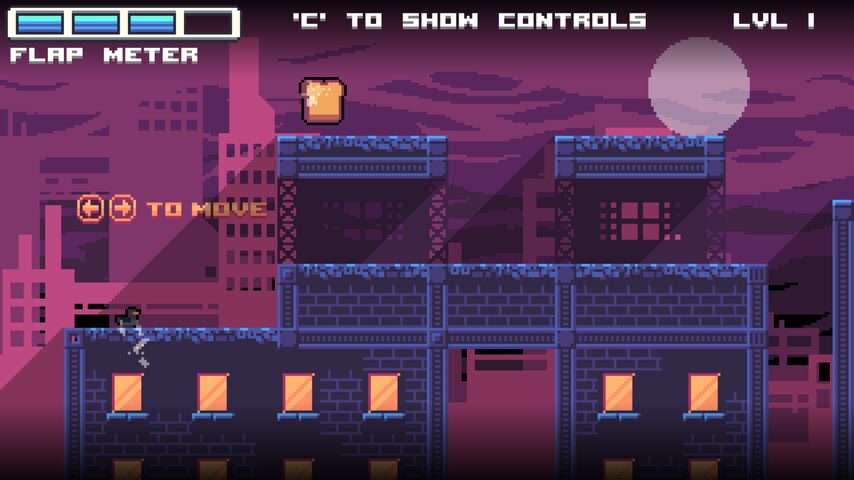} \\
    \bottomrule
    \end{tabular}
\end{table}

\begin{table}[!t]
    \centering
    \scriptsize
    \caption*{\textbf{Table~\ref{tab:game_info} (continued):} Game inventory for the 34-game \gameworld{} benchmark.}
    \begin{tabular}{@{}p{0.18\linewidth}p{0.15\linewidth}p{0.38\linewidth}>{\centering\arraybackslash}m{0.18\linewidth}@{}}
    \gameinfotableheader
    25-rocket-league-2d & \gamecite{Rocket League 2D}{game25-rocket-league-2d} & Side-view car-soccer game requiring positioning, jumping, and ball control to score goals. & \gameshot{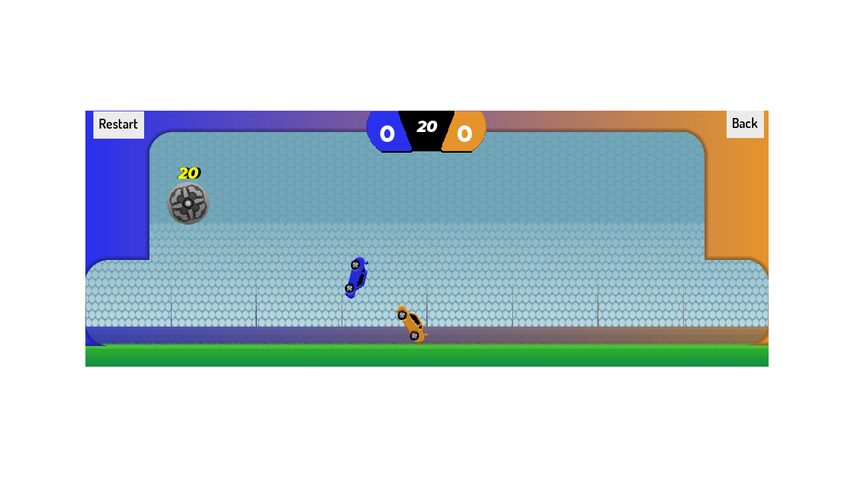} \\
    26-run-3 & \gamecite{Run 3}{game26-run-3} & Tunnel runner that combines lateral movement and jumps to cross gaps in a rotating corridor. & \gameshot{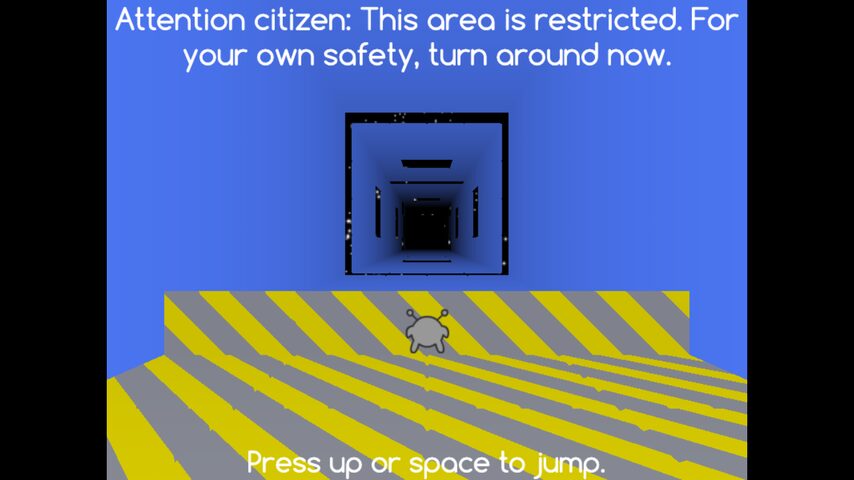} \\
    27-stack & \gamecite{Stack}{game27-stack} & Timing puzzle in which moving blocks must be dropped with precise alignment to keep the tower stable. & \gameshot{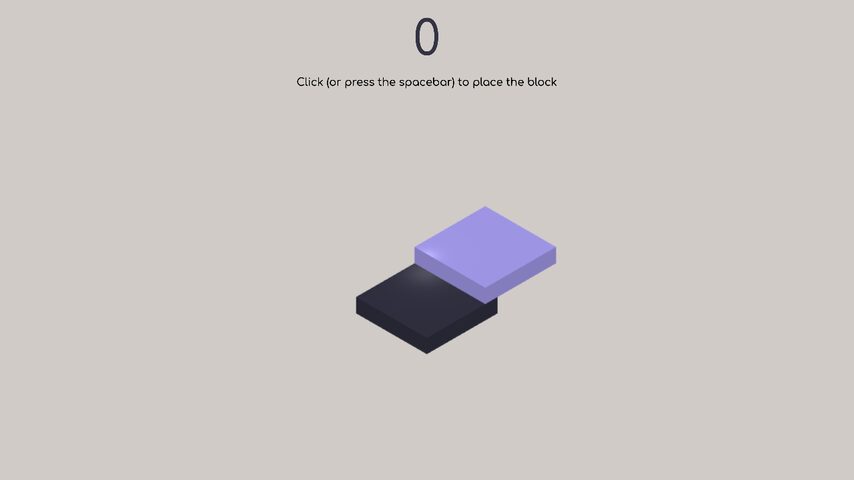} \\
    28-temple-run-2 & \gamecite{Temple Run 2}{game28-temple-run-2} & Endless runner requiring turn, jump, and slide decisions under high-speed reactive pressure. & \gameshot{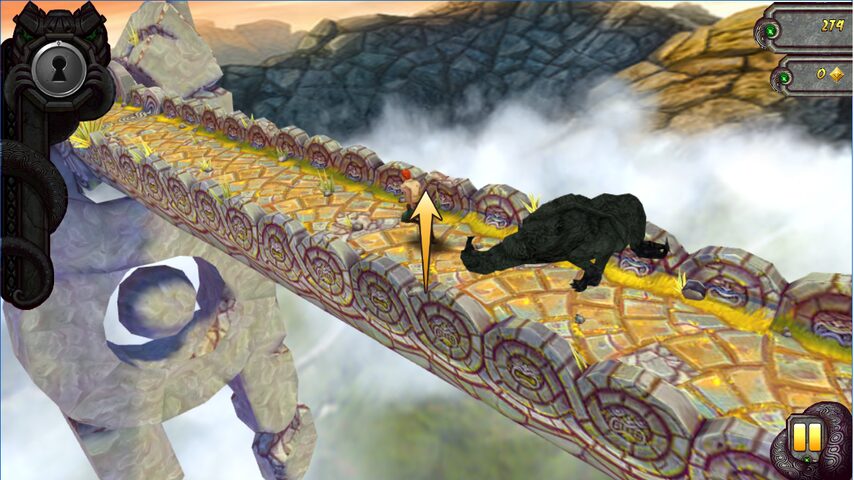} \\
    29-tetris & \gamecite{Tetris}{game29-tetris} & Falling-block puzzle focused on line clearing, spatial planning, and managing long-term board structure. & \gameshot{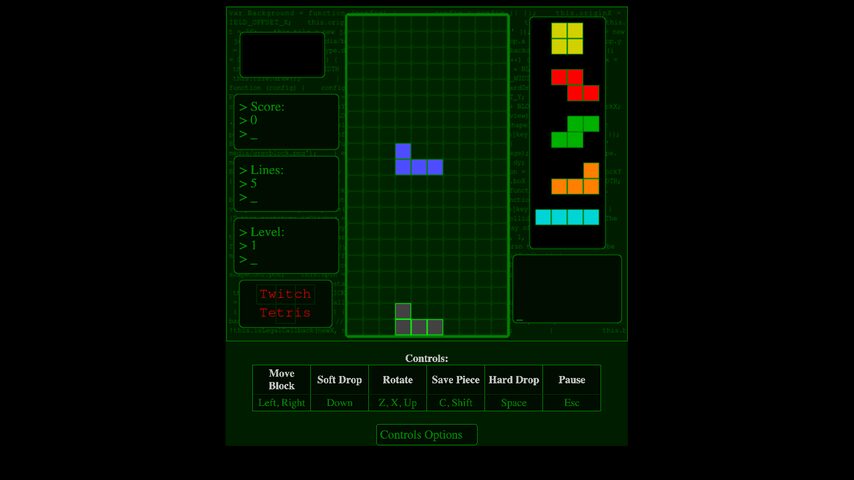} \\
    30-vex-3 & \gamecite{Vex 3}{game30-vex-3} & Precision platformer built around checkpoints, trap avoidance, and accurate movement through hazard-heavy levels. & \gameshot{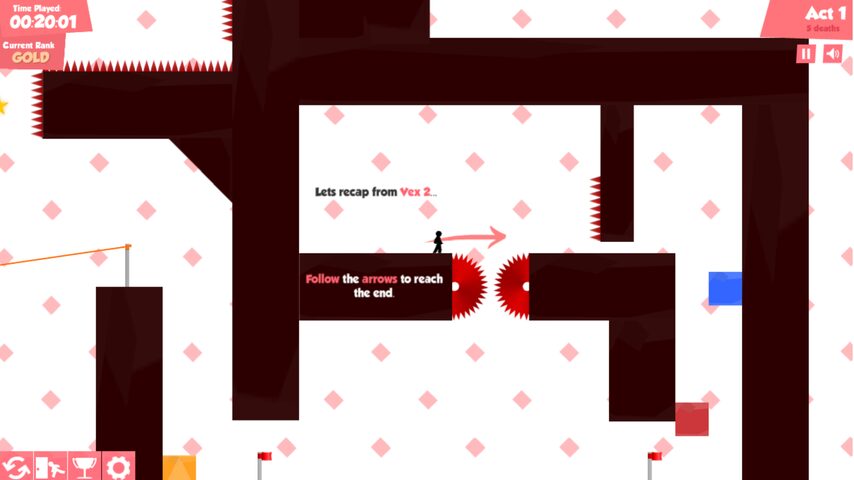} \\
    31-wolf3d & \gamecite{Wolfenstein 3D}{game31-wolf3d} & First-person shooter benchmark emphasizing navigation, target detection, and combat survival in a 3D maze. & \gameshot{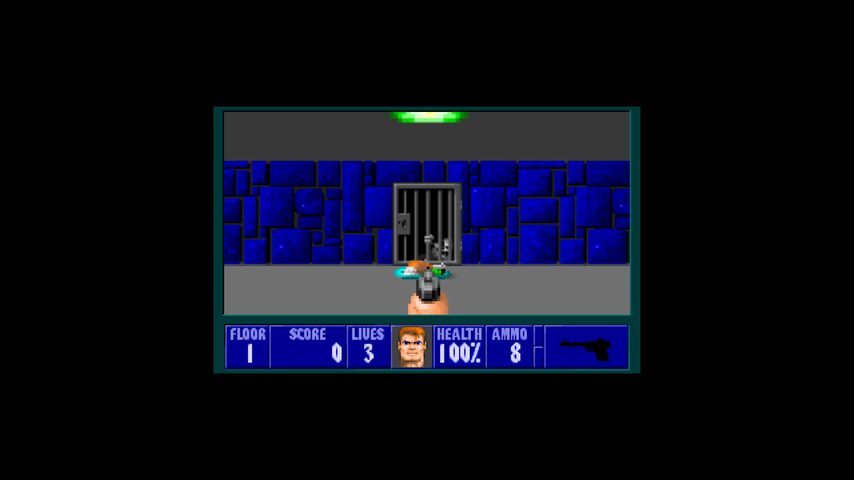} \\
    32-wordle & \gamecite{Wordle}{game32-wordle} & Word-guessing puzzle where the player uses color feedback to infer a hidden five-letter word. & \gameshot{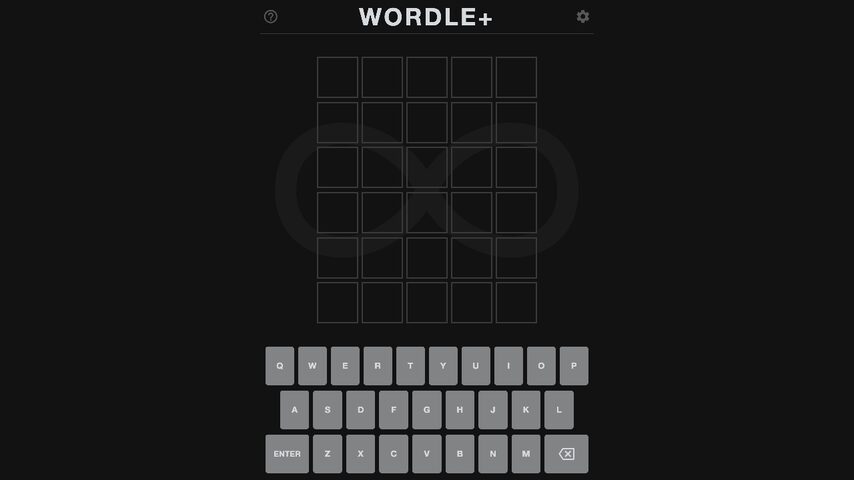} \\
    33-worlds-hardest-game & \gamecite{World's Hardest Game}{game33-worlds-hardest-game} & Precision dodge maze where the player collects coins and reaches the exit while avoiding moving enemies. & \gameshot{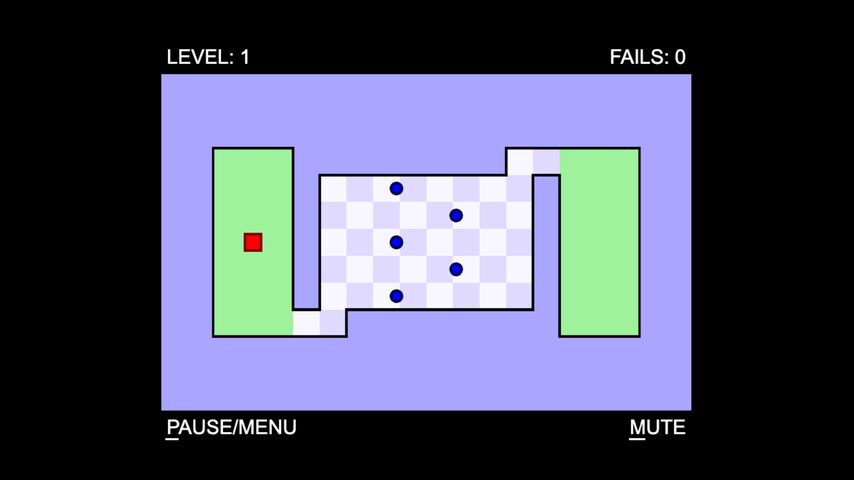} \\
    34-worlds-hardest-game-2 & \gamecite{World's Hardest Game 2}{game34-worlds-hardest-game-2} & A harder follow-up dodge maze with denser enemy patterns and stricter movement precision. & \gameshot{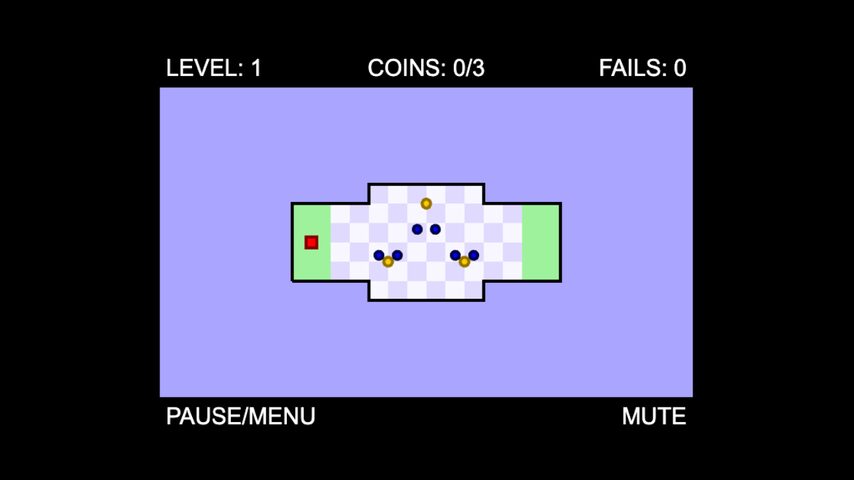} \\
    \bottomrule
    \end{tabular}
\end{table}

\normalsize

\subsection{Browser-Based Sandbox Environment}
\label{sec:sandbox}

The central design goal of the sandbox is to decouple agent decision quality from inference speed. In real-time games, a slower model faces a harder game state by the time it acts, conflating thinking time with gameplay ability. To eliminate this confound, the sandbox can pause game execution during model inference, so every agent faces identical game dynamics regardless of response latency. Scores then reflect what the agent decides, not how fast it responds. The sandbox also supports real-time evaluation for studying how latency affects gameplay in practice.

The sandbox ensures each game runs in an isolated browser instance following a strict observation-action loop: capture a screenshot, query the model, execute one action. Before the first agent decision and after each reset, the environment waits until the game reports an actionable state (by default, \texttt{ready} or \texttt{playing}) and absorbs transient loading or menu phases through this readiness gate. Besides this, the sandbox also supports configurable game speed and deterministic seed settings for evaluation reproducibility.
Appendix~\ref{sec:app_browser} details the browser manager, readiness gate, and Game API contract behind this sandbox.

\subsection{Outcome-Based State-Verifiable Evaluation}
\label{sec:eval}

One key characteristic of \gameworld{} is that evaluation is based on interaction \emph{outcomes} rather than on the model response itself, with the underlying game state remains \emph{verifiable} throughout agent execution. Most existing game benchmarks evaluate agents through OCR, pixel-level heuristics, or VLM-as-judge pipelines, all of which may introduce noise into the evaluation. \gameworld{} instead adopts outcome-based state-verifiable evaluation: for each game, we inject a structured JavaScript bridge that exposes serialized \texttt{gameAPI} state directly to the evaluator, including lifecycle status, terminal metadata, and task-relevant gameplay variables such as score, level, coordinates, lives, coins, or checkpoints. This yields deterministic, fully verifiable signals with no perceptual noise. In total, we instrument 233 task-relevant state fields across 34 games (averaging available 6.85 fields per game), with each field is manually designed to capture a gameplay quantity relevant to task evaluation.
Appendices~\ref{sec:app_loop} and \ref{sec:app_game_api} further show the details of the observation-action-evaluation loop and provide a concrete example of the serialized \texttt{gameAPI} verifiable-state schema.

At every step, the evaluator reads the current \texttt{gameAPI} state, resolves a task score from either a configured scalar field or an aggregate over multiple fields, and computes the two metrics defined in Section~\ref{sec:benchmark}: $\mathcal{SR}$ (whether the task succeeds) and $\mathcal{PG}$ (normalized task progress from the configured start score to the target score). This task-level $\mathcal{PG}$ is distinct from any native in-game \texttt{game\_state.progress}, which we keep only as diagnostic game progress. Stopping and status are then determined by target reach, terminal signals, task-specific end-field rules, and the fixed step budget. When an agent hits a terminal failure (e.g., losing all lives), the environment resets and the agent continues under the same step budget rather than being immediately terminated, while preserving the run-level best progress reached so far. This prevents a single early mistake from zeroing out an otherwise competent run.

\newcommand{\taxonomymodellogo}[2]{\raisebox{-0.05em}{\includegraphics[height=0.9em]{figures/logos/#1}} #2}

\begin{table}[t]
\centering
\scriptsize
\setlength{\tabcolsep}{2.5pt}
\resizebox{\linewidth}{!}{%
\begin{tabular}{@{}l c c l@{}}
\toprule
\rowcolor{headerblue}
\textbf{Model} & \makecell[c]{\textbf{Computer-Use} \\ \textbf{Agent}} & \makecell[c]{\textbf{Generalist} \\ \textbf{Agent}} & \textbf{Model Description} \\
\midrule
\rowcolor{bggray}
\multicolumn{4}{l}{\textcolor{textgray}{\textbf{\textit{Proprietary}}}} \\
\taxonomymodellogo{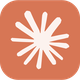}{Claude-Sonnet-4.6}~\cite{anthropic_claude}        & \cmark & \cmark & Anthropic multimodal model supporting computer-use. \\
\taxonomymodellogo{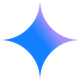}{Gemini-2.5-Computer-Use}~\cite{gemini_2_5_computer_use}  & \cmark &  & Computer-use model built on Gemini 2.5 Pro. \\
\taxonomymodellogo{gemini}{Gemini-3-Flash-Preview}~\cite{gemini_family}   &  & \cmark & Google fast multimodal foundation model. \\
\taxonomymodellogo{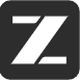}{GLM-4.6V}~\cite{hong2025glm}                 &  & \cmark & Z.ai VLM with tool use. \\
\taxonomymodellogo{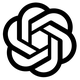}{GPT-5.2}~\cite{openai_gpt5_system_card}                  &  & \cmark & OpenAI multimodal function model with reasoning. \\
\taxonomymodellogo{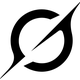}{Grok-4.1-Fast-Reasoning}~\cite{xai_grok}  &  & \cmark & xAI foundation model with fast reasoning. \\
\taxonomymodellogo{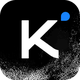}{Kimi-K2.5}~\cite{kimi}                &  & \cmark & Moonshot multimodal foundation model. \\
\taxonomymodellogo{gpt}{OpenAI-Computer-Use}~\cite{openai-cua}      & \cmark &  & OpenAI native computer-use agent. \\
\taxonomymodellogo{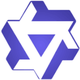}{Qwen3-VL-Plus}~\cite{Qwen3-VL}            & \cmark & \cmark & Alibaba hosted visual foundation model. \\
\taxonomymodellogo{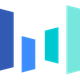}{Seed-1.8}~\cite{guo2025seed1}                 & \cmark & \cmark & ByteDance Seed's multimodal model. \\
\midrule
\rowcolor{bggray}
\multicolumn{4}{l}{\textcolor{textgray}{\textbf{\textit{Open-Source}}}} \\
\taxonomymodellogo{qwen}{Qwen3-VL-235B-A22B}~\cite{Qwen3-VL}       & \cmark & \cmark & Open flagship Qwen3-VL Mixture-of-Experts model. \\
\taxonomymodellogo{qwen}{Qwen3-VL-30B-A3B}~\cite{Qwen3-VL}             & \cmark & \cmark & Open compact Qwen3-VL Mixture-of-Experts model. \\
\taxonomymodellogo{seed}{UI-TARS-1.5-7B}~\cite{seed2025uitars15}           & \cmark &  & Open-weight native GUI agent by ByteDance Seed. \\
\bottomrule
\end{tabular}
}
\caption{
\captfont{
Model profiles in \gameworld{}. Each model is evaluated as a Computer-Use Agent, a Generalist multimodal agent, or both.
}
}
\label{tab:model_profiles}
\end{table}

\section{Experiments}
\label{sec:results}

\subsection{Experiment Setup}

We evaluate 13 base models in the \gameworld{} benchmark across both \textbf{Computer-Use Agent (CUA)} and \textbf{Generalist Agent} interfaces. In total, this yields 18 model--agent-interface pairs (8 CUAs + 10 Generalist Agents), summarized in the model taxonomy table in Table~\ref{tab:model_profiles}. The evaluated models include:
\begin{itemize}
    \item \textbf{Proprietary models:} Claude-Sonnet-4.6~\cite{anthropic_claude}, Gemini-2.5-Computer-Use~\cite{gemini_2_5_computer_use}, Gemini-3-Flash-Preview \cite{gemini_family}, GLM-4.6V~\cite{hong2025glm}, GPT-5.2~\cite{openai_gpt5_system_card}, Grok-4.1-Fast-Reasoning~\cite{xai_grok}, Kimi-K2.5~\cite{kimi}, OpenAI-Computer-Use~\cite{openai-cua}, Qwen3-VL-Plus~\cite{Qwen3-VL}, and Seed-1.8~\cite{guo2025seed1}. 
    \item \textbf{Open-source models:} Qwen3-VL-235B-A22B~\cite{Qwen3-VL}, Qwen3-VL-30B-A3B~\cite{Qwen3-VL}, and UI-TARS-1.5-7B~\cite{seed2025uitars15}.
\end{itemize}

For all models, we use the same paused evaluation protocol under a shared runtime and verifier: the game is paused during inference so that scores reflect decision quality rather than response speed. Each model outputs one interaction command per step with a fixed per-action execution duration (usually 200--500 ms, depending on the game), and a maximum budget of 100 actions per task. During the interactions, all metrics are continuously computed from verifiable game state by the evaluator defined in Section~\ref{sec:eval}. The exact model-side output-format prompts used in these evaluations are listed in Appendix~\ref{sec:app_output_formats}.

\subsection{Main Results}

\newcommand{\goldmedal}{\rlap{\,\raisebox{-0.1em}{\includegraphics[height=0.9em]{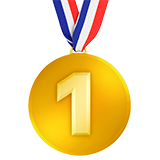}}}}
\newcommand{\silvermedal}{\rlap{\,\raisebox{-0.1em}{\includegraphics[height=0.9em]{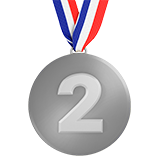}}}}
\newcommand{\bronzemedal}{\rlap{\,\raisebox{-0.1em}{\includegraphics[height=0.9em]{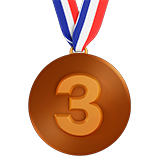}}}}
\newcommand{\modellogo}[2]{\raisebox{-0.05em}{\includegraphics[height=0.9em]{figures/logos/#1}} #2}
\begin{table}[!t]
\scriptsize
\centering
\caption{
\captfont{
\textbf{Main results on \gameworld{}} across 34 games and 170 tasks. We report genre-level and overall $\mathcal{SR}$ (Success Rate, \%) and $\mathcal{PG}$ (Progress, \%) for 18 models (10 generalist multimodal agents and 8 computer-use agents). The final rank is determined by overall $\mathcal{PG}$.
}
}
\setlength{\tabcolsep}{0.25mm}

\newcolumntype{S}{>{\centering\arraybackslash}p{0.61cm}}
\newcolumntype{M}{>{\centering\arraybackslash}p{0.75cm}}
\newcolumntype{R}{>{\centering\arraybackslash\cellcolor{rankbgyellow}}p{0.8cm}}

{
\resizebox{\linewidth}{!}{
\begin{tabular}{l *{2}{S} *{2}{M} *{6}{S} | *{2}{S} *{1}{R}}
\toprule
\multirow{2}{*}{\textbf{Model}} &
\multicolumn{2}{c}{\textbf{Arcade}} &
\multicolumn{2}{c}{\textbf{Platformer}} &
\multicolumn{2}{c}{\textbf{Puzzle}}  &
\multicolumn{2}{c}{\textbf{Runner}} &
\multicolumn{2}{c|}{\textbf{Simulation}} &
\multicolumn{3}{c}{\textbf{Overall}} \\
\cmidrule(lr){2-3} \cmidrule(lr){4-5} \cmidrule(lr){6-7} \cmidrule(lr){8-9} \cmidrule(lr){10-11} \cmidrule(lr){12-14}

 & $\mathcal{SR}$ & $\mathcal{PG}$ & $\mathcal{SR}$ & $\mathcal{PG}$ & $\mathcal{SR}$ & $\mathcal{PG}$ & $\mathcal{SR}$ & $\mathcal{PG}$ & $\mathcal{SR}$ & $\mathcal{PG}$ & $\mathcal{SR}$ & $\mathcal{PG}$ & Rank \\

\midrule
\multicolumn{14}{l}{\cellcolor{bggray}\textcolor{textgray}{\textbf{\textit{Human}}}}\\
Novice Player  & 45.7 & 55.5 & 60.0 & 65.6 & 51.4 & 63.1 & 60.0 & 72.0 & 60.0 & 62.0 & 55.3 & 64.1 & \cellcolor{rankbgyellow}-- \\
Expert Player  & 65.7 & 73.9 & 85.0 & 88.0 & 68.6 & 77.1 & 82.5 & 87.8 & 85.0 & 86.0 & 77.1 & 82.6 & \cellcolor{rankbgyellow}-- \\

\midrule
\multicolumn{14}{l}{\cellcolor{bggray}\textcolor{textgray}{\textbf{\textit{Computer-Use Agents}}}}   \\
\modellogo{claude}{Claude-Sonnet-4.6} & 8.6 & 27.2 & 22.5 & 36.5 & 20.0 & 43.8 & 30.0 & 55.6 & 10.0 & 16.8 & 19.4 & 38.3 & \cellcolor{rankbgyellow}\textbf{2}\silvermedal \\
\modellogo{gemini}{Gemini-2.5-Computer-Use} & 5.7 & 28.0 & 20.0 & 35.8 & 11.4 & 32.2 & 30.0 & 55.4 & 10.0 & 19.3 & 16.5 & 36.1 & \cellcolor{rankbgyellow}\textbf{3}\bronzemedal \\
\modellogo{gpt}{OpenAI-Computer-Use} & 5.7 & 24.7 & 17.5 & 31.3 & 20.0 & 45.8 & 27.5 & 53.0 & 5.0 & 12.0 & 16.5 & 35.8 & \cellcolor{rankbgyellow}4 \\
\modellogo{qwen}{Qwen3-VL-Plus} & 5.7 & 23.5 & 20.0 & 34.6 & 14.3 & 35.6 & 27.5 & 51.0 & 5.0 & 10.7 & 15.9 & 33.6 & \cellcolor{rankbgyellow}5 \\
\modellogo{seed}{Seed-1.8} & 8.6 & 31.1 & 25.0 & 40.3 & 25.7 & 52.0 & 27.5 & 50.6 & 5.0 & 11.0 & 20.0 & 39.8 & \cellcolor{rankbgyellow}\textbf{1}\goldmedal \\
\modellogo{qwen}{Qwen3-VL-235B-A22B} & 5.7 & 23.2 & 22.5 & 35.2 & 8.6 & 29.7 & 25.0 & 51.0 & 0.0 & 1.7 & 14.1 & 31.4 & \cellcolor{rankbgyellow}6 \\
\modellogo{qwen}{Qwen3-VL-30B-A3B} & 8.6 & 26.8 & 20.0 & 31.9 & 2.9 & 27.6 & 25.0 & 50.3 & 0.0 & 2.2 & 12.9 & 30.8 & \cellcolor{rankbgyellow}8 \\
\modellogo{seed}{UI-TARS-1.5-7B} & 5.7 & 31.4 & 15.0 & 24.4 & 5.7 & 29.9 & 27.5 & 52.4 & 0.0 & 3.8 & 12.4 & 31.1 & \cellcolor{rankbgyellow}7 \\

\midrule
\multicolumn{14}{l}{\cellcolor{bggray}\textcolor{textgray}{\textbf{\textit{Generalist Multimodal Agents}}}}\\
\modellogo{claude}{Claude-Sonnet-4.6} & 5.7 & 28.3 & 22.5 & 37.0 & 25.7 & 51.5 & 30.0 & 51.9 & 15.0 & 16.6 & 20.6 & 39.3 & \cellcolor{rankbgyellow}\textbf{3}\bronzemedal \\
\modellogo{gemini}{Gemini-3-Flash-Preview} & 5.7 & 26.3 & 25.0 & 41.2 & 25.7 & 54.8 & 32.5 & 55.4 & 10.0 & 21.1 & 21.2 & 41.9 & \cellcolor{rankbgyellow}\textbf{1}\goldmedal \\
\modellogo{glm}{GLM-4.6V} & 8.6 & 22.8 & 20.0 & 33.9 & 5.7 & 29.1 & 27.5 & 49.1 & 0.0 & 5.3 & 14.1 & 30.8 & \cellcolor{rankbgyellow}8 \\
\modellogo{gpt}{GPT-5.2} & 8.6 & 29.3 & 22.5 & 36.7 & 28.6 & 56.2 & 27.5 & 52.6 & 10.0 & 16.9 & 20.6 & 40.6 & \cellcolor{rankbgyellow}\textbf{2}\silvermedal \\
\modellogo{grok}{Grok-4.1-Fast-Reasoning} & 8.6 & 23.7 & 22.5 & 37.3 & 14.3 & 46.6 & 25.0 & 49.0 & 5.0 & 10.4 & 16.5 & 36.0 & \cellcolor{rankbgyellow}6 \\
\modellogo{Kimi}{Kimi-K2.5} & 8.6 & 26.4 & 20.0 & 35.3 & 25.7 & 51.4 & 27.5 & 49.7 & 5.0 & 11.7 & 18.8 & 37.4 & \cellcolor{rankbgyellow}5 \\
\modellogo{qwen}{Qwen3-VL-Plus} & 8.6 & 25.6 & 22.5 & 37.8 & 14.3 & 39.1 & 27.5 & 51.1 & 0.0 & 10.0 & 16.5 & 35.4 & \cellcolor{rankbgyellow}7 \\
\modellogo{seed}{Seed-1.8} & 11.4 & 33.5 & 22.5 & 34.6 & 22.9 & 48.7 & 27.5 & 51.2 & 10.0 & 18.8 & 20.0 & 39.0 & \cellcolor{rankbgyellow}4 \\
\modellogo{qwen}{Qwen3-VL-235B-A22B} & 5.7 & 23.2 & 17.5 & 29.5 & 8.6 & 33.3 & 27.5 & 50.4 & 0.0 & 3.6 & 13.5 & 30.8 & \cellcolor{rankbgyellow}9 \\
\modellogo{qwen}{Qwen3-VL-30B-A3B} & 2.9 & 20.3 & 20.0 & 36.5 & 2.9 & 26.1 & 27.5 & 51.1 & 0.0 & 3.5 & 12.4 & 30.6 & \cellcolor{rankbgyellow}10 \\

\bottomrule
\end{tabular}
}
}

\label{tab:result_by_genre}
\end{table}

\paragraph{Metric Definitions.}
Let $\mathcal{R}$ be the set of evaluated runs and let $N=|\mathcal{R}|$. For run $i$, let $q_{i,t}$ denote the task score read from verifiable game state at step $t$. In the runtime, $q_{i,t}$ is defined either by a configured scalar score field or by the sum of configured aggregate score fields. Let $b_i$ be the starting score of each task, $\tau_i$ the configured task target score, and $q_i^{\max}=\max_t q_{i,t}$ the best score observed in the run. By construction, benchmark tasks satisfy $\tau_i > b_i$. The run-level progress is then:
\begin{equation}
    \mathrm{progress}_i =
    \operatorname{clip}_{[0,1]}\!\left(\frac{q_i^{\max} - b_i}{\tau_i - b_i}\right).
    \label{eq:progress}
\end{equation}
When reset-on-fail is enabled, episode-local score tracking is cleared after each reset, but the run-level best progress is preserved. Therefore, $\mathrm{progress}_i$ measures the furthest normalized progress reached within the fixed step budget, rather than only the final episode before termination. Finally, for each model, we report the averaged $\mathcal{SR}$ and $\mathcal{PG}$ over all runs:
\begin{equation}
    \mathcal{SR} = \frac{1}{N}\sum_{i=1}^{N}\mathbf{1}[\mathrm{status}_i=\texttt{success}],\quad
    \mathcal{PG} = \frac{1}{N}\sum_{i=1}^{N}\mathrm{progress}_i
    \label{eq:success-rate}
\end{equation}
For better readability, both metrics are reported in percentage form in Table~\ref{tab:result_by_genre}.

\begin{figure}[!t]
    \centering
    \includegraphics[width=\linewidth]{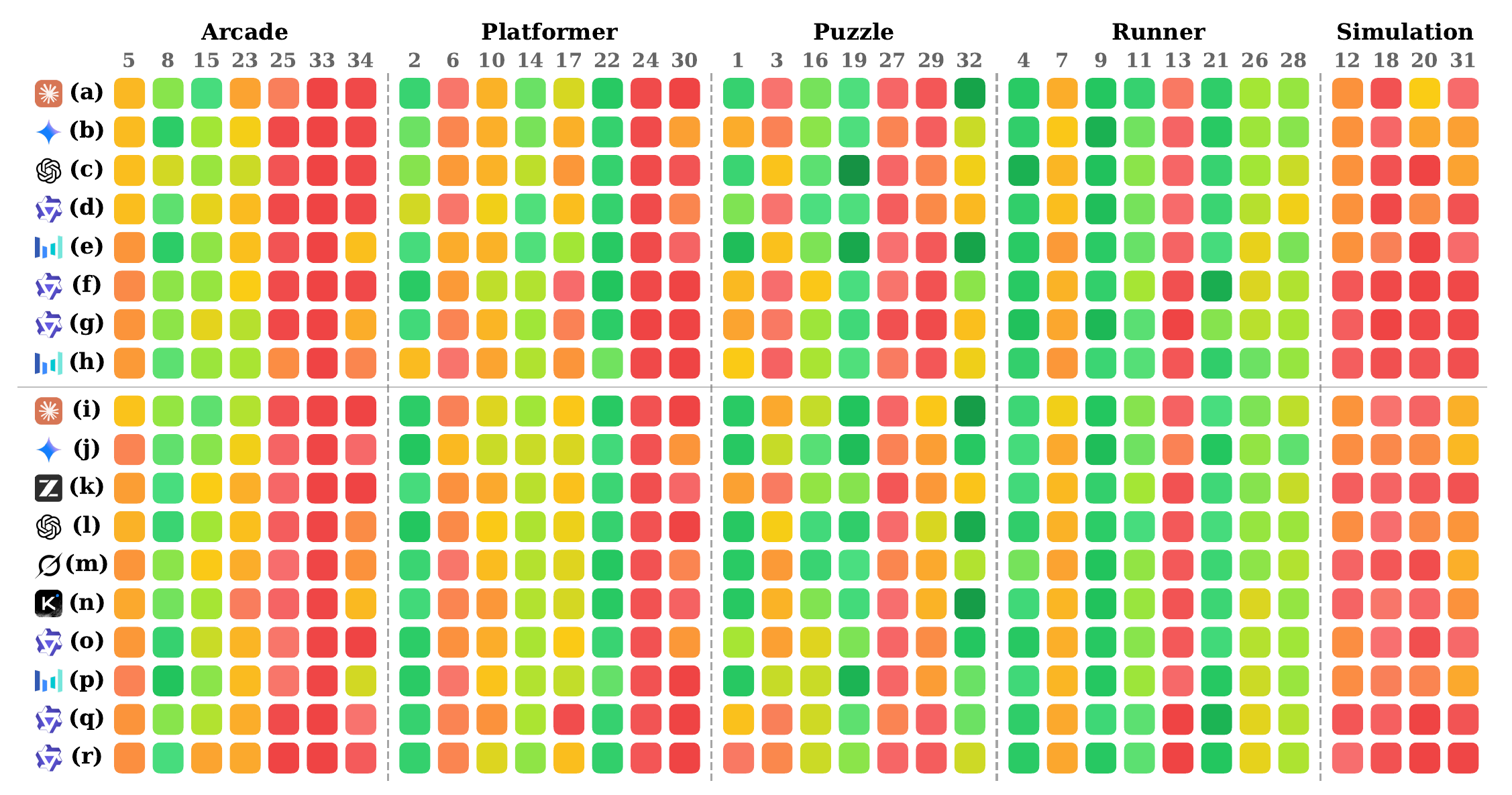}
    \caption{
        \captfont{Per-game progress heatmap across the \gameworld{} benchmark. Rows correspond to 18 evaluated game agents with model: \textbf{(a)} Claude-Sonnet-4.6, \textbf{(b)} Gemini-2.5-Computer-Use, \textbf{(c)} OpenAI-Computer-Use, \textbf{(d)} Qwen3-VL-Plus, \textbf{(e)} Seed-1.8, \textbf{(f)} Qwen3-VL-235B-A22B, \textbf{(g)} Qwen3-VL-30B-A3B, \textbf{(h)} UI-TARS-1.5-7B, \textbf{(i)} Claude-Sonnet-4.6, \textbf{(j)} Gemini-3-Flash-Preview, \textbf{(k)} GLM-4.6V, \textbf{(l)} GPT-5.2, \textbf{(m)} Grok-4.1-Fast-Reasoning, \textbf{(n)} Kimi-K2.5, \textbf{(o)} Qwen3-VL-Plus, \textbf{(p)} Seed-1.8, \textbf{(q)} Qwen3-VL-235B-A22B, and \textbf{(r)} Qwen3-VL-30B-A3B. \textbf{(a)-(h)} are Computer-Use Agents and \textbf{(i)-(r)} are Generalist Multimodal Agents. Colors represent average task progress for each game from {\setlength{\fboxrule}{0pt}\setlength{\fboxsep}{0.7pt}\colorbox{green!20}{\emph{high (green)}}} to {\setlength{\fboxrule}{0pt}\setlength{\fboxsep}{0.7pt}\colorbox{yellow!25}{\emph{medium (yellow)}}} to {\setlength{\fboxrule}{0pt}\setlength{\fboxsep}{0.7pt}\colorbox{red!15}{\emph{low (red)}}}.}
    }
    \label{fig:result_by_game}
\end{figure}

\paragraph{Agent Performance.}

Table~\ref{tab:result_by_genre} summarizes performance for 18 model--interface pairs, ranked by overall $\mathcal{PG}$. 
Overall performance remains far from satisfactory. Among Generalist agents, Gemini-3-Flash-Preview achieves the best overall $\mathcal{PG}=41.9$, followed by GPT-5.2 at 40.6; Claude-Sonnet-4.6 and Seed-1.8 reach 39.3 and 39.0, respectively. Among Computer-Use Agents, Seed-1.8 performs best at 39.8, with Claude-Sonnet-4.6 close behind. However, overall $\mathcal{SR}$ remains relatively low (12.4--21.2\%), indicating that models are often capable of making partial progress without meeting the full task target. To this end, our outcome-based state-verifiable evaluation provides a more fine-grained task-progress signal, making it possible to distinguish partial advancement from full completion and to diagnose capability gaps beyond binary task success rate.

Figure~\ref{fig:result_by_game} further visualizes per-game progress beyond genre averages. At the genre level, Runner games yield the highest progress for many models. Simulation tasks remain broadly challenging, with low success and progress for many models, highlighting the difficulty of open-ended objectives and longer-horizon state tracking.

\paragraph{Human Players.}
We also conduct a human study with two computer-science post-graduate students. One had no prior exposure to the benchmark games or tasks, and we report this participant as the \textit{Novice Player}. The other had studied all the game rules and practiced the controls beforehand, and we report this participant as the \textit{Expert Player}. For better consistency, we use the same action budget as in the agent evaluation: each task is limited to 100 primitive actions (mouse clicks or key presses). 

The performances of human players in Table~\ref{tab:result_by_genre} show that, under the same action budget, the best current agents remain far below the Novice Player (55.3 $\mathcal{SR}$ / 64.1 $\mathcal{PG}$), highlighting many challenges remain in building game agents for robust control, long-horizon planning, and reliable task completion.

\subsection{Benchmark Robustness Under Repeated Evaluation}

To test whether \gameworld{} behaves as a reproducible measurement platform rather than a one-off leaderboard snapshot, we perform repeated full-benchmark evaluation on two open-source backbones, Qwen3-VL-30B-A3B and Qwen3-VL-235B-A22B, each in both CUA and Generalist interfaces, yielding four model--interface pairs. Due to the cost of repeated full-benchmark runs, we restrict this validation study to these two open-source models. For each setting, we report mean $\pm$ standard deviation over ten full-benchmark reruns.

\begin{table}[H]
\scriptsize
\centering
\caption{
\captfont{Repeat-averaged overall $\mathcal{SR}$ and $\mathcal{PG}$ for the four Qwen model--interface pairs used in the repeated-evaluation study. We report mean $\pm$ standard deviation computed from the ten full-benchmark repeat averages.}
}
\setlength{\tabcolsep}{5pt}
\begin{tabular}{l c c c c}
\toprule
Model & Agent Interface & Repeats & Overall $\mathcal{SR}$ & Overall $\mathcal{PG}$ \\
\midrule
Qwen3-VL-30B-A3B & Computer-Use Agent & 10 & 12.7$\pm$1.2 & 30.9$\pm$1.1 \\
Qwen3-VL-30B-A3B & Generalist Agent & 10 & 12.5$\pm$1.3 & 30.7$\pm$1.1 \\
Qwen3-VL-235B-A22B & Computer-Use Agent & 10 & 13.8$\pm$0.7 & 30.4$\pm$0.7 \\
Qwen3-VL-235B-A22B & Generalist Agent & 10 & 13.6$\pm$1.4 & 30.1$\pm$0.5 \\
\bottomrule
\end{tabular}
\label{tab:qwen_repeat_stability}
\end{table}

Table~\ref{tab:qwen_repeat_stability} summarizes the resulting overall statistics. The central observation is stability: across all four settings, the standard deviation of overall $\mathcal{PG}$ remains in a low single-digit band, and the corresponding $\mathcal{SR}$ variation is likewise limited. This indicates that the benchmark can reproduce the same broad performance level and capability trends across reruns, which is necessary if the platform is to serve as a meaningful test bed for game agents. At the same time, the repeated runs also set an interpretation boundary: very small differences between nearby systems should not be overstated without rerun-based evidence.

\begin{figure}[!h]
    \centering
    \includegraphics[width=\linewidth]{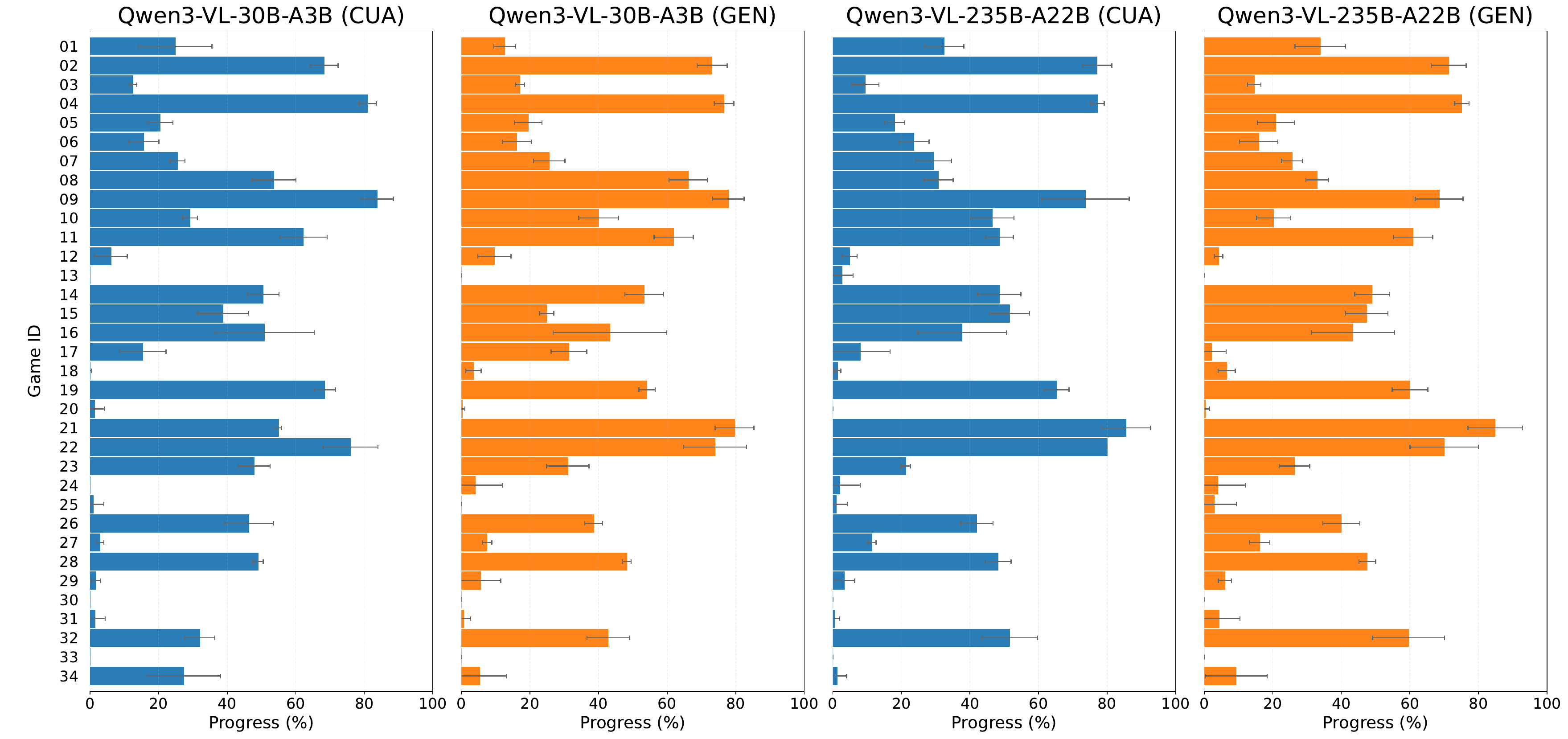}
    \caption{
        \captfont{Per-game average progress across the 34 benchmark games for the four Qwen model--interface pairs used in the repeated-evaluation study. Each panel corresponds to one model--interface pair, and each horizontal bar shows the mean progress over the same ten full-benchmark reruns. Error bars denote one run-level standard deviation.}
    }
    \label{fig:qwen_repeat_per_game_progress}
\end{figure}

Figure~\ref{fig:qwen_repeat_per_game_progress} provides the corresponding per-game view. Most games show tight run-to-run bands, while visibly larger variance is concentrated in a limited subset of control-sensitive or high-difficulty games such as \textit{Hextris}, \textit{Cubefield}, \textit{Wordle}, and \textit{World's Hardest Game 2}. This is the expected pattern for a robust benchmark: aggregate conclusions remain reproducible, while difficult games still expose meaningful differences in planning, control, and memory.

\subsection{Capability-Aligned Curriculum Analysis}

Genre-level averages alone cannot tell whether a failure is mainly caused by weak capabilities such as control grounding, reactive behavior, spatial navigation, or long-horizon reasoning. Therefore, for better interpretation, we conduct \emph{diagnosis-driven analysis} to understand why models fail beyond genre-level result aggregation. We group the games into a five-level curriculum in which each level is anchored by its dominant capability bottleneck. The curriculum makes these patterns more interpretable across both Generalist and Computer-Use agents, and provides a diagnosable structure for improving future game agents.

\begin{figure}[!th]
    \centering
    \begin{minipage}[t]{0.49\textwidth}
        \centering
        \includegraphics[width=\linewidth]{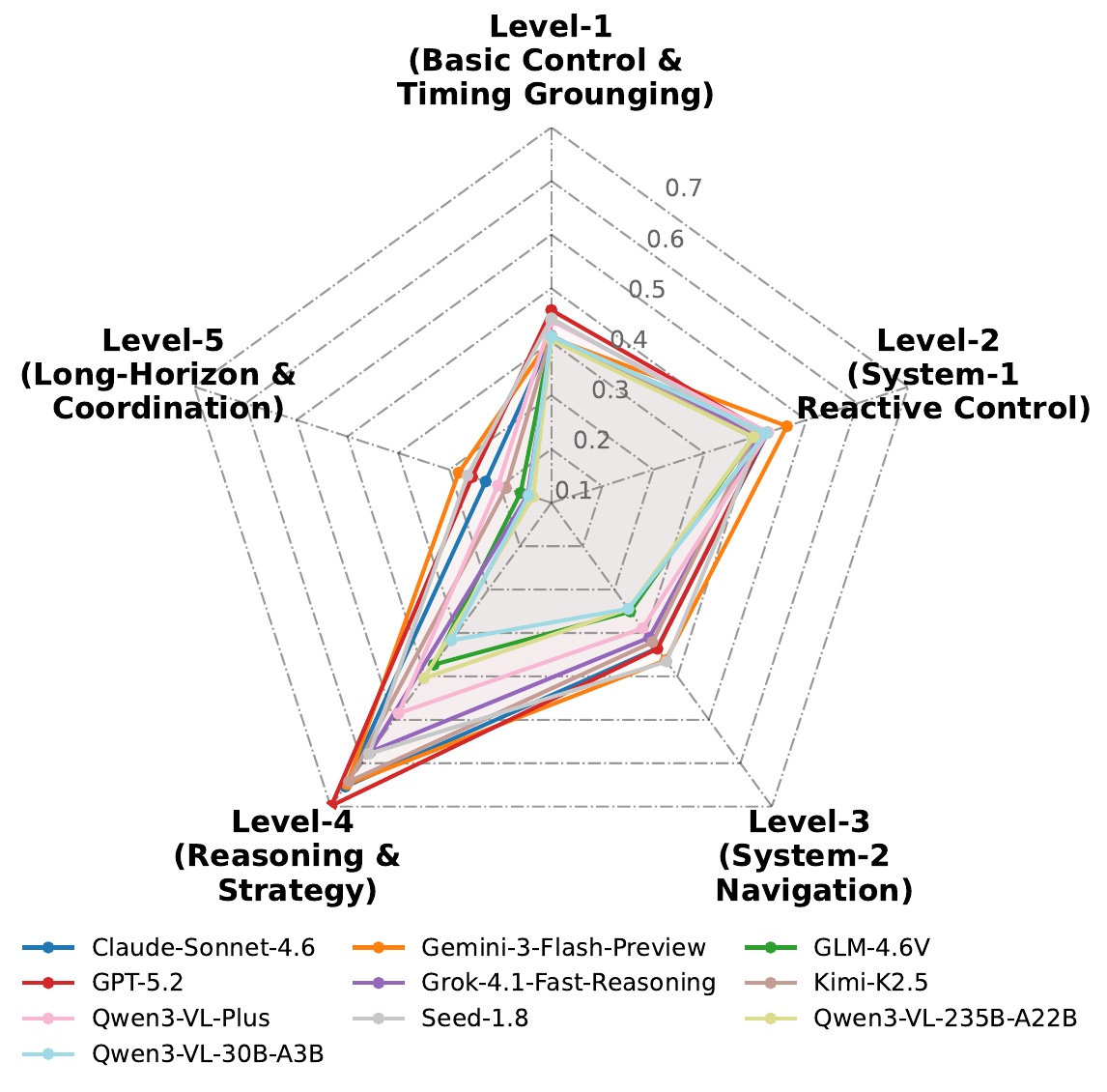}
    \end{minipage}%
    \begin{minipage}[t]{0.49\textwidth}
        \centering
        \includegraphics[width=\linewidth]{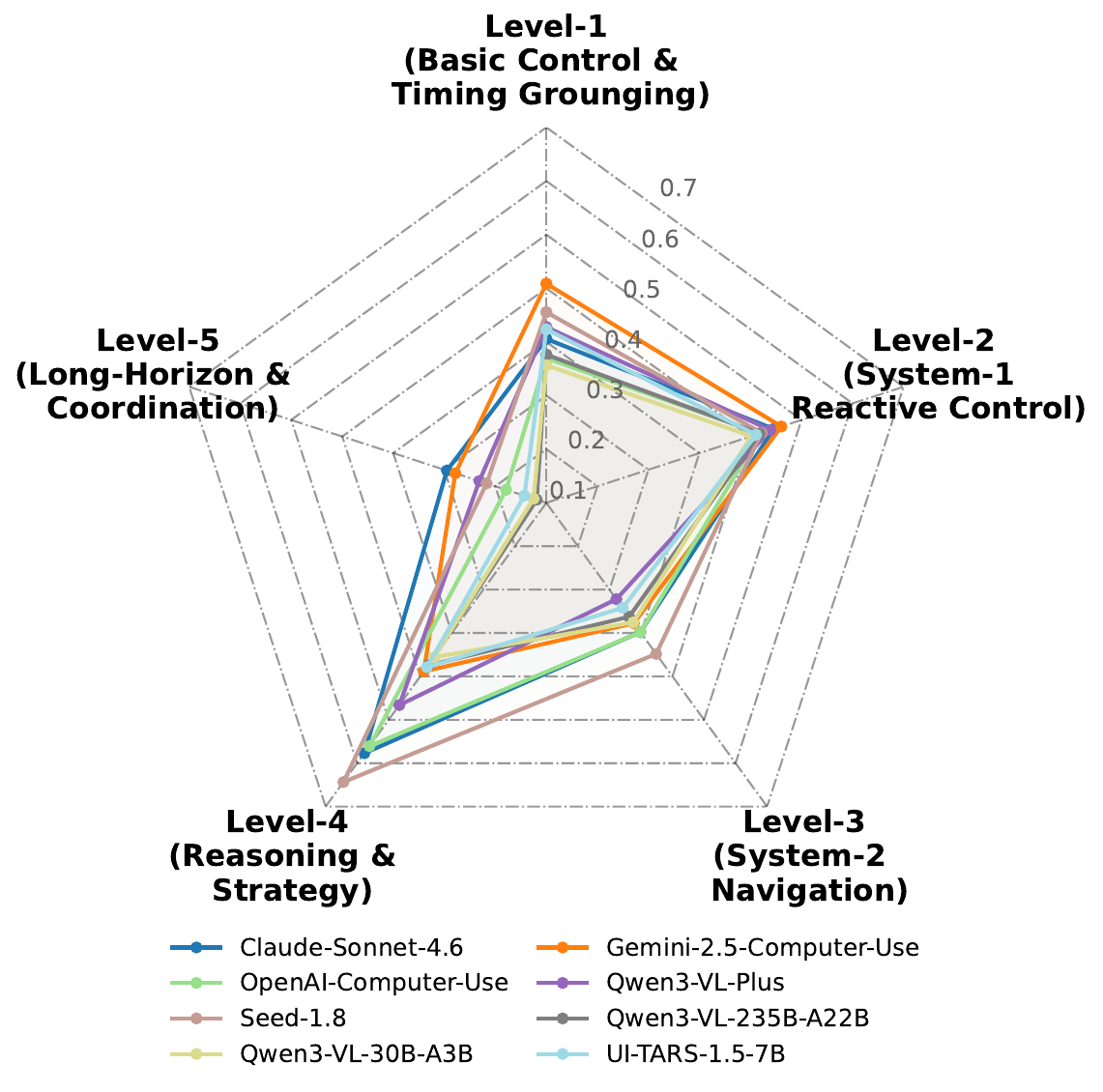}
    \end{minipage}
    \caption{
        \captfont{Capability-aligned five-level curriculum profiles across agent interfaces and models. 
        \textbf{Left:} Generalist agents. \textbf{Right:} Computer-Use agents. Each radar axis corresponds to one curriculum level and values are average task progress of all the games in the level.}
    }
    \label{fig:by_level_radar}
\end{figure}

\begin{itemize}
    \item \textbf{Level-1 (Basic Control and Timing Grounding):} This level isolates whether an agent can reliably map visual observations to valid atomic interactions such as clicking, issuing a single key press, or waiting, and can trigger them at the appropriate moment under low strategic load. Since planning demands are intentionally light, failures here mainly indicate weak action grounding, poor visual perception, or weak basic timing judgment; games include \gameid{5-breakout}, \gameid{8-core-ball}, and \gameid{27-stack}.
    \item \textbf{Level-2 (System-1 Reactive Control):} This level emphasizes high-frequency reflexes in continuously evolving scenes where immediate reaction dominates over deliberate planning. Performance in level-2 reflects the agent's sensitivity to latency, timing precision, and short-horizon motor stability; games include \gameid{4-boxel-rebound}, \gameid{7-chrome-dino}, \gameid{9-cubefield}, \gameid{10-doodle-jump}, \gameid{11-edge-surf}, \gameid{13-flappy-bird}, \gameid{14-geodash}, \gameid{21-ns-shaft}, \gameid{24-restless-wing-syndrome}, \gameid{26-run-3}, \gameid{28-temple-run-2}, and \gameid{30-vex-3}.
    \item \textbf{Level-3 (System-2 Spatial Navigation):} This level mostly tests whether agents can model a 2D or 3D geometric world and use it for deliberate pathfinding in structured layouts. Underperformance here usually reflects weak spatial reasoning, waypoint sequencing, or unstable coordination between high-level intent and precise action; games include \gameid{2-another-gentlemans-adventure}, \gameid{3-astray}, \gameid{6-captaincallisto}, \gameid{15-google-snake}, \gameid{17-mario-game}, \gameid{22-ovo}, \gameid{23-pacman}, \gameid{25-rocket-league-2d}, \gameid{31-wolf3d}, \gameid{33-worlds-hardest-game}, and \gameid{34-worlds-hardest-game-2}.
    \item \textbf{Level-4 (Symbolic Reasoning \& Strategy):} This level groups rule-intensive, discrete environments in which the bottleneck is strategy planning over a structured state space. Differences between this level and the control-oriented levels reveal the agent's limitations in symbolic planning, rule tracking, and long-horizon decision consistency; games include \gameid{1-2048}, \gameid{16-hextris}, \gameid{19-minesweeper}, \gameid{29-tetris}, and \gameid{32-wordle}.
    \item \textbf{Level-5 (Open-World Coordination \& Management):} This level captures the most open-ended settings in the current suite, where agents must coordinate navigation, interaction, and subgoal management in high-dimensional environments. Evaluation at this level usually reflects compounded failures in memory, policy stability, and error recovery; games include \gameid{12-fireboy-and-watergirl}, \gameid{18-minecraft-clone-glm}, and \gameid{20-monkey-mart}.
\end{itemize}

Figure~\ref{fig:by_level_radar} visualizes per-level progress under this curriculum for Generalist and Computer-Use agents. It can be observed that both interfaces exhibit a similar performance which peaks at Level 4 and 2, but drops sharply at Level 1 and 5. It suggests that game agents performs well in strategic decision making and reacting, as MLLMs always do in generic tasks, while long-horizon tasks and timing grounding still remaining bottleneck for game playing.

\subsection{Challenges and Analyses:\\ Real-Time Interaction, Context-Memory Sensitivity, Action Validity, and Failure Modes}
To expore the current limitation and future imporvement directions of game agents, we further examine real-time interaction, context-memory sensitivity, action validity, and interpretable failure modes. Beyond raw leaderboard metrics, these analysis dimensions highlight several future challenges for multimodal game agents.

\begin{table}[!b]
\centering
\begin{minipage}[t]{0.48\textwidth}
\centering
\captionof{table}{
\captfont{
\textbf{GameWorld-RT} results for Qwen3-VL-30B-A3B and Qwen3-VL-235B-A22B in Generalist and CUA interfaces. In \textbf{GameWorld-RT} benchmark, the environment continues running during model inference. `RT sec/step' is seconds per executed step.
}
}
\label{tab:latency}
\scriptsize
\setlength{\tabcolsep}{0.1pt}
\begin{tabular}{@{}>{\raggedright\arraybackslash}p{0.50\linewidth}>{\centering\arraybackslash}p{0.20\linewidth}>{\centering\arraybackslash}p{0.14\linewidth}>{\centering\arraybackslash}p{0.14\linewidth}@{}}
\toprule
\rowcolor{headerblue}
\textbf{Model} &
\textbf{\shortstack[c]{Real-Time\\sec/step}} &
\textbf{\shortstack[c]{$\mathcal{SR}$}} &
\textbf{\shortstack[c]{$\mathcal{PG}$}} \\
\midrule
\multicolumn{4}{l}{\cellcolor{bggray}\textcolor{textgray}{\textbf{\textit{Computer-Use Agents}}}}\\
Qwen3-VL-235B-A22B & 6.2 & 17.1 & 33.2 \\
Qwen3-VL-30B-A3B & 2.4 & 15.6 & 33.0 \\
\midrule
\multicolumn{4}{l}{\cellcolor{bggray}\textcolor{textgray}{\textbf{\textit{Generalist Multimodal Agents}}}}\\
Qwen3-VL-235B-A22B & 6.4 & 16.8 & 34.0 \\
Qwen3-VL-30B-A3B & 3.4 & 15.6 & 32.9 \\
\bottomrule
\end{tabular}

\end{minipage}%
\hspace{0.025\textwidth}
\begin{minipage}[t]{0.48\textwidth}
\centering
\captionof{table}{
\captfont{
Memory-round sensitivity across the full benchmark for Qwen3-VL-235B-A22B in Generalist and CUA interfaces. We vary memory rounds and report average input tokens, wall-clock sec/step, and overall $\mathcal{PG}$.
}
}
\label{tab:memory}
\scriptsize
\setlength{\tabcolsep}{0.1pt}
\begin{tabular}{@{}>{\centering\arraybackslash}p{0.18\linewidth}>{\raggedright\arraybackslash}p{0.47\linewidth}>{\centering\arraybackslash}p{0.13\linewidth}>{\centering\arraybackslash}p{0.11\linewidth}>{\centering\arraybackslash}p{0.10\linewidth}@{}}
\toprule
\rowcolor{headerblue}
\textbf{\shortstack[c]{Memory\\Rounds}} &
\textbf{Model} &
\textbf{\shortstack[c]{Input\\Tokens}} &
\textbf{\shortstack[c]{sec/\\step}} &
\textbf{$\mathcal{PG}$} \\
\midrule
\multirow[c]{2}{*}{0} & Qwen3-VL-235B-A22B & 1278 & 5.5 & 30.0 \\
 & Qwen3-VL-235B-A22B-CUA & 1891 & 7.2 & 30.3 \\
\midrule
\multirow[c]{2}{*}{1} & Qwen3-VL-235B-A22B & 2171 & 6.8 & 30.1 \\
 & Qwen3-VL-235B-A22B-CUA & 3771 & 10.1 & 29.0 \\
\midrule
\multirow[c]{2}{*}{2} & Qwen3-VL-235B-A22B & 3052 & 8.6 & 30.6 \\
 & Qwen3-VL-235B-A22B-CUA & 5627 & 12.8 & 28.7 \\
\bottomrule
\end{tabular}

\end{minipage}
\end{table}

\subsubsection{GameWorld-RT: Real-Time Benchmark}
\label{sec:real_time}
Beyond the default paused-inference evaluation, we also establish \gameworldrt{} as a separate benchmark variant for more realistic interactive evaluation. Real-time interaction is a critical dimension of digital-agent performance because many practical settings require agents to perceive, reason, and act under continuously evolving environmental dynamics. It also introduces a distinct and more deployment-faithful challenge: the agent must not only choose the right action, but do so quickly enough for that action to remain relevant when it is executed. In \gameworldrt{}, the environment does not pause while the model is reasoning, so response latency becomes part of the task itself. Table~\ref{tab:latency} reports Qwen3-VL-30B-A3B and Qwen3-VL-235B-A22B results on \gameworldrt{} in both Generalist and CUA interfaces.

\gameworldrt{} remains challenging across all four settings. The smaller 30B backbone is substantially faster, while the 235B backbone achieves slightly higher progress; however, success rates remain only in very low throughout, indicating that faster reaction alone does not solve the benchmark when the environment keeps running during inference. Real-time play therefore exposes a distinct difficulty in which reasoning speed and action timing are more tightly coupled.

We treat \gameworldrt{} as complementary to the default paused benchmark. The paused setting isolates decision quality by removing response-time confounds, whereas \gameworldrt{} captures a more real-world deployment-oriented setting in which reasoning, reaction time, and action timing are coupled. Note that results on \gameworldrt{} should not be compared directly with those on the default paused benchmark, because in the real-time setting the game continues to evolve during model inference, so the effective gameplay duration includes reasoning time and is therefore longer under the same action budget.

\subsubsection{Context-Memory Sensitivity}
\label{sec:memory_ablation}
We also analyze the memory-round ablation across the same 34-game benchmark. Table~\ref{tab:memory} shows that increasing memory substantially raises both prompt length and wall-clock latency: the Generalist interface grows from about 1.3k to 3.1k input tokens and from 5.5 to 8.6 seconds per step, while the CUA interface grows from about 1.9k to 5.6k tokens and from 7.2 to 12.8 seconds per step. More importantly, the performance effect is inconsistent for the two interfaces: for the Generalist Agents, $\mathcal{PG}$ rises modestly as memory rounds increase, while for the CUAs, performance steadily declines.

This split is plausible given the differences in action spaces. Generalist agents operate over semantic trajectories, so a longer history can preserve useful task context because the semantic content of each action is retained. CUA agents, by contrast, carry longer low-level action traces without semantic information, making their history harder to interpret jointly with the screenshot. This is more likely to accumulate distracting interaction details as memory grows. Since the time cost increases substantially in both interfaces, memory should be viewed as a selective benefit for game agents rather than a uniformly helpful module.

\subsubsection{Action Validity and Instruction Following}

\begin{table}[!t]
\centering
\begin{minipage}[t]{0.58\textwidth}
\centering
\captionof{table}{
\captfont{
Invalid-action categories and example model outputs.
}
}
\label{tab:invalid_action_examples}
\scriptsize
\setlength{\tabcolsep}{1pt}
\vspace{-2mm}
\renewcommand{\arraystretch}{1.0}
\begin{tabular}{p{0.2\linewidth}>{\raggedright\arraybackslash}p{0.78\linewidth}}
\toprule
\rowcolor{headerblue}
\textbf{Category} & \textbf{Example} \\ 
\midrule
\textbf{No-Tool-Call\newline(NTC)} &
{\setlength{\fboxsep}{2pt}\colorbox{bggray}{\textcolor{textgray}{Model responses in natural language without tools:}}}\newline
\texttt{\# Model Output:}\par
\hspace*{0.6em}\texttt{The obstacle is coming from the left, so I should move left first.}\par
\texttt{\# Invalid Reason:}\par
\hspace*{0.6em}The model does not generate an executable tool call but instead returns free-form natural language.\smallskip\newline
{\setlength{\fboxsep}{2pt}\colorbox{bggray}{\textcolor{textgray}{Malformed format due to truncation of very long reasoning:}}}\newline
\texttt{\# Model Output:}\par
\hspace*{0.6em}\texttt{"\textless think\textgreater{} I should first inspect the scene carefully before acting... \textless /think\textgreater{} \textless tool\_call\textgreater{} \{"name": " }\par
\texttt{\# Invalid Reason:}\par
\hspace*{0.6em}The tool-call block is never properly closed. \\
\midrule
\textbf{Out-of-Space\newline(OOS)} &
{\setlength{\fboxsep}{2pt}\colorbox{bggray}{\textcolor{textgray}{High-level actions requiring multiple steps are not allowed:}}}\newline
\texttt{\# Model Output:}\par
\hspace*{0.6em}\texttt{craft\_a\_workbench()}\par
\texttt{\# Invalid Reason:}\par
\hspace*{0.6em}The model returns a plausible tool call, but this semantic action is not registered.\smallskip\newline
{\setlength{\fboxsep}{1.5pt}\colorbox{bggray}{\textcolor{textgray}{In a keyboard-only game where the control space is stated}}}\newline
{\setlength{\fboxsep}{1.5pt}\colorbox{bggray}{\textcolor{textgray}{ explicitly in the game rules:}}}\newline
\texttt{\# Model Output:}\par
\hspace*{0.6em}\texttt{left\_click(x=512, y=384)}\par
\texttt{\# Invalid Reason:}\par
\hspace*{0.6em}The CUA model calls a computer-use tool, but mouse clicking is outside the allowed control space. \\
\bottomrule
\end{tabular}

\end{minipage}%
\hspace{0.02\textwidth}
\begin{minipage}[t]{0.39\textwidth}
\centering
\captionof{table}{
\captfont{
Invalid Action Rate (IAR) across all evaluated agents, broken down into No-Tool-Call (NTC) and Out-of-Space (OOS), i.e., $\mathrm{IAR} = \mathrm{No~Call} + \mathrm{OOS}$.
}
}
\label{tab:iar}
\scriptsize
\setlength{\tabcolsep}{2pt}
\renewcommand{\arraystretch}{0.95}
\begin{tabular}{p{0.5\linewidth}>{\centering\arraybackslash}p{0.14\linewidth}>{\centering\arraybackslash}p{0.14\linewidth}>{\centering\arraybackslash}p{0.14\linewidth}}
\toprule
\rowcolor{headerblue}
\textbf{Model} &
\textbf{\shortstack[c]{IAR\\(\%)}} &
\textbf{\shortstack[c]{NTC\\(\%)}} &
\textbf{\shortstack[c]{OOS\\(\%)}} \\
\midrule
\rowcolor{bggray}
\multicolumn{4}{l}{\textcolor{textgray}{\textit{Computer-Use Agents}}} \\
Claude-Sonnet-4.6 & 0.0 & 0.0 & 0.0 \\
Gemini-2.5-Computer-Use & 0.0 & 0.0 & 0.0 \\
OpenAI-Computer-Use & 0.0 & 0.0 & 0.0 \\
Qwen3-VL-Plus & <0.1 & <0.1 & 0.0 \\
Seed-1.8 & 0.0 & 0.0 & 0.0 \\
Qwen3-VL-235B-A22B & <0.1 & <0.1 & 0.0 \\
Qwen3-VL-30B-A3B & <0.1 & <0.1 & 0.0 \\
UI-TARS-1.5-7B & 0.4 & <0.1 & 0.4 \\
\midrule
\rowcolor{bggray}
\multicolumn{4}{l}{\textcolor{textgray}{\textit{Generalist Multimodal Agents}}} \\
Claude-Sonnet-4.6 & 0.0 & 0.0 & 0.0 \\
Gemini-3-Flash-Preview & 0.0 & 0.0 & 0.0 \\
GLM-4.6V & 8.3 & 7.6 & 0.7 \\
GPT-5.2 & 0.0 & 0.0 & 0.0 \\
Grok-4.1-Fast-Reasoning & <0.1 & <0.1 & 0.0 \\
Kimi-K2.5 & 0.0 & 0.0 & 0.0 \\
Qwen3-VL-Plus & <0.1 & <0.1 & 0.0 \\
Seed-1.8 & 0.0 & 0.0 & 0.0 \\
Qwen3-VL-235B-A22B & <0.1 & <0.1 & 0.0 \\
Qwen3-VL-30B-A3B  & 2.7 & 2.7 & <0.1 \\
\midrule
\rowcolor{bggray}
\textcolor{textgray}{\textbf{Overall Mean}} & 0.8 & 0.8 & 0.0 \\
\bottomrule
\end{tabular}

\end{minipage}
\end{table}

Agents cannot act in a free-form manner in interactive environments; they must obey role-specific control constraints and action-space rules at every step.
Beyond the main benchmark metrics, invalid-action statistics remain useful as a lightweight reliability signal.
Invalid Action Rate (IAR) is the fraction of proposed actions that fail tool-call parsing, role constraints, or parser checks.

\begin{equation}
    \mathrm{IAR} = 1 - \frac{\sum_{r \in \mathcal{R}}\#\mathrm{valid\_actions}(r)}{\sum_{r \in \mathcal{R}}\#\mathrm{proposed\_actions}(r)}.
\end{equation}

We therefore treat lower IAR (Eq.~2) as a direct instruction-following proxy. Specifically, to further understand the sources, we separate invalid actions into two categories:
\begin{itemize}
    \item \textbf{No-Tool-Call (NTC)} means the model does not emit any executable tool call at all, typically because overly long thinking leads to truncation or because the final output does not satisfy the required tool-call formatting. In practice, this often appears either as free-form natural-language output with no tool invocation, or as an unfinished tool-call block that failed to be parsed.
    \item  \textbf{Out-of-Space (OOS)} means the model does return a tool call, but the call falls outside the legal action space: for example, a CUA may request a forbidden key or mouse operation, a generalist agent may emit an unregistered action, or the tool call may omit required arguments or provide malformed parameters. 
\end{itemize}

Table~\ref{tab:invalid_action_examples} shows representative invalid-action categories with placeholder outputs, and Table~\ref{tab:iar} reports the aggregate invalid-action breakdowns. Appendix~\ref{sec:app_action_validation} further details the low-level legality checker. Overall, under long interactive contexts, weaker models are more likely to forget the available action space and emit non-executable or non-permitted tool calls.

\subsubsection{Failure Modes and Analysis}
We identify four task-failure categories that are useful for understanding game agents' performance in the case study: 
\textbf{Instruction-following}, \textbf{Perception}, \textbf{Fine-grained action}, and \textbf{Long-horizon memory}.

\paragraph{Perception failures.}
The agent misreads visual state (objects, UI cues, or spatial layout) of the game, causing incorrect action decisions. These errors are often visible in the model's intermediate reasoning process, for example when it incorrectly identifies the position of obstacles or misjudges the traversable region of a map. Such errors are particularly pronounced in cluttered scenes or under partial observability, where fine-grained visual discrimination is required.

\paragraph{Fine-grained action failures.}
The high-level intent is correct but low-level execution is mistimed or imprecise (\textit{e.g.,} jump timing, key-combo duration). Even when the model correctly understands the current game state, it may still fail to choose or execute the right action for that state, often because it does not fully capture the game mechanics or the effect of its own actions. These failures highlight the gap between strategic reasoning and the precise motor-level control demanded by real-time gameplay.

\paragraph{Instruction-following failures.}
The agent proposes actions that violate declared controls, output schema, or task-level action constraints. In some cases, the agent ignores specific parts of the user instruction. Under longer interaction trajectories, it may even drift away from the final task objective and start executing irrelevant or unproductive behaviors. This typically manifests as invalid key bindings, malformed action outputs, or attempts to invoke unavailable mechanics.

\paragraph{Long-horizon memory failures.}
The agent loses critical historical context, repeats ineffective loops, or fails to preserve multi-step plans. This behavior is especially common in weaker base models, where the agent may repeatedly issue the same ineffective action, receive no useful feedback, and enter a loop without self-correction. This reflects fundamental limitations in the agent's ability to maintain coherent goal representations across extended interaction horizons.

\section{Case Study}

We provide three case studies to illustrate the interaction of game agents with the environment. Each case is presented through 5 key frames together with a short reasoning summary, the proposed and executed action, and the corresponding verifiable state change from \texttt{gameAPI}.

\subsection{Game Agent Interface Comparison: Generalist Agent vs.\ CUA}

\begin{figure}[!ht]
    \centering
    \includegraphics[page=2,width=\linewidth]{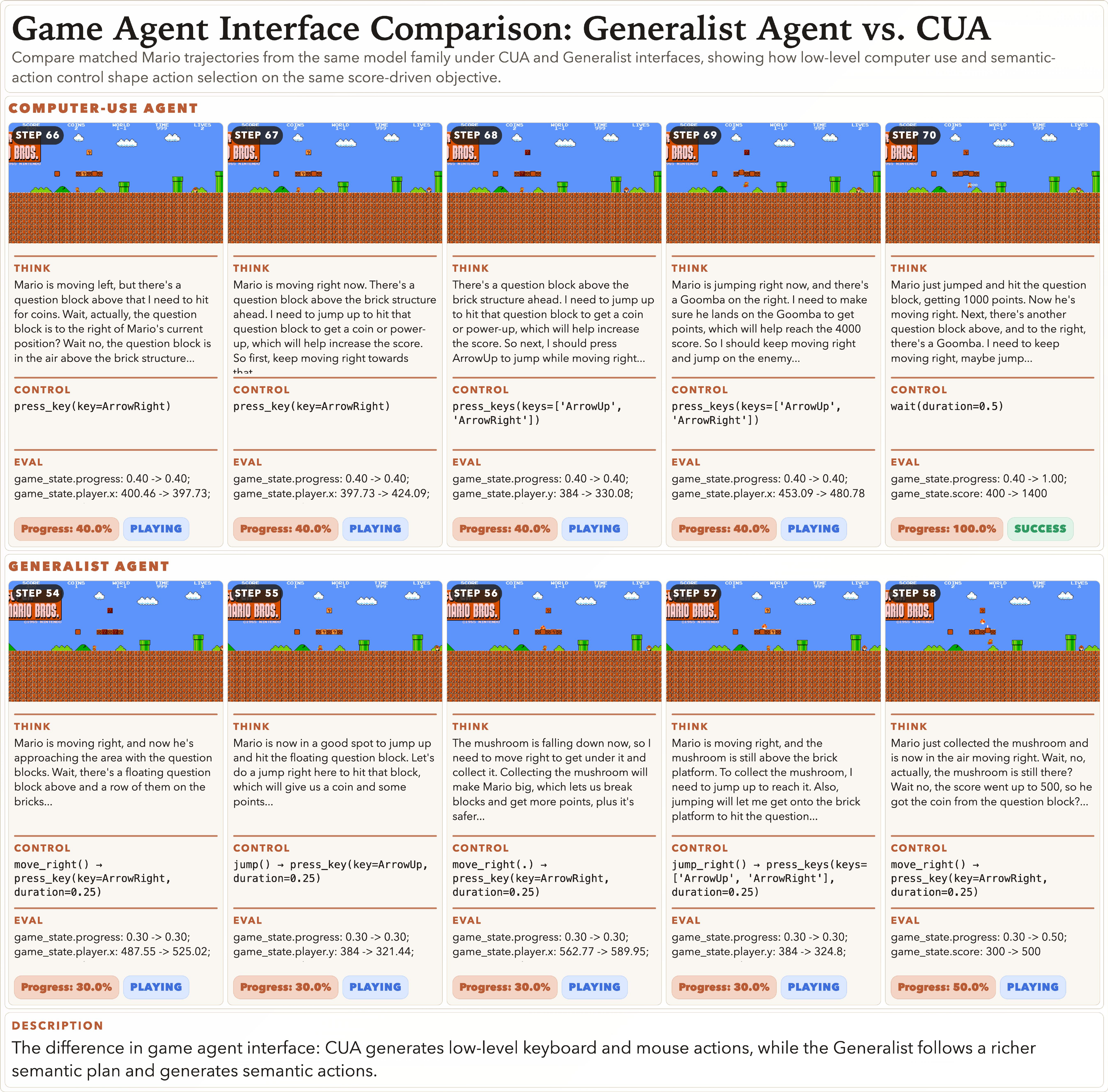}
    \caption{\captfont{Case-study visualization for game-agent interface comparison. This \texttt{Mario} page stacks a matched CUA trajectory above a Generalist trajectory so that their divergence can be attributed to the action interface rather than the backbone.}}
    \label{fig:case_interface}
\end{figure}

To better illustrate the differences of two game agent interfaces, we firstly present an example of a shared task accoss CUA and Generalist agents. Figure~\ref{fig:case_interface} shows matched trajectories of \texttt{Mario-Game} under CUA and Generalist interfaces. With the same model backbone and game environment, the difference only existing in control interface: CUA emits low-level keyboard and mouse actions directly, while the Generalist follows a richer semantic plan and produces semantic actions.

\begin{figure}[!t]
    \centering 
    \includegraphics[page=1,width=\linewidth]{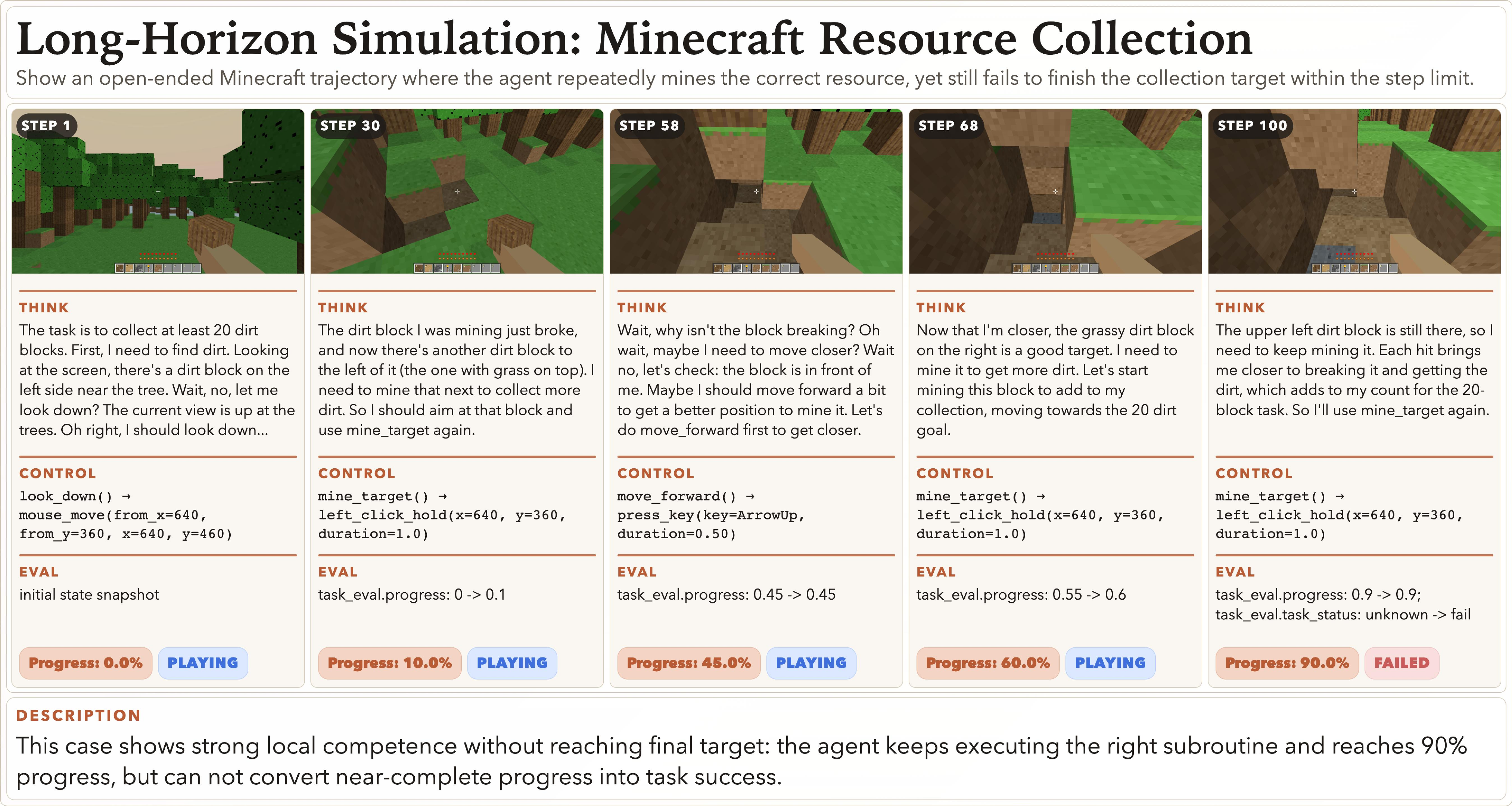}
    \caption{\captfont{Case-study visualization for long-horizon simulation and resource collection. This \texttt{Minecraft Clone} trajectory shows locally plausible interactions that advance the progress, but still fail to fully meet the task target.}}
    \label{fig:case_long_horizon}
\end{figure}


\begin{figure}[!h]
    \centering
    \vspace{1em}
    \includegraphics[page=3,width=\linewidth]{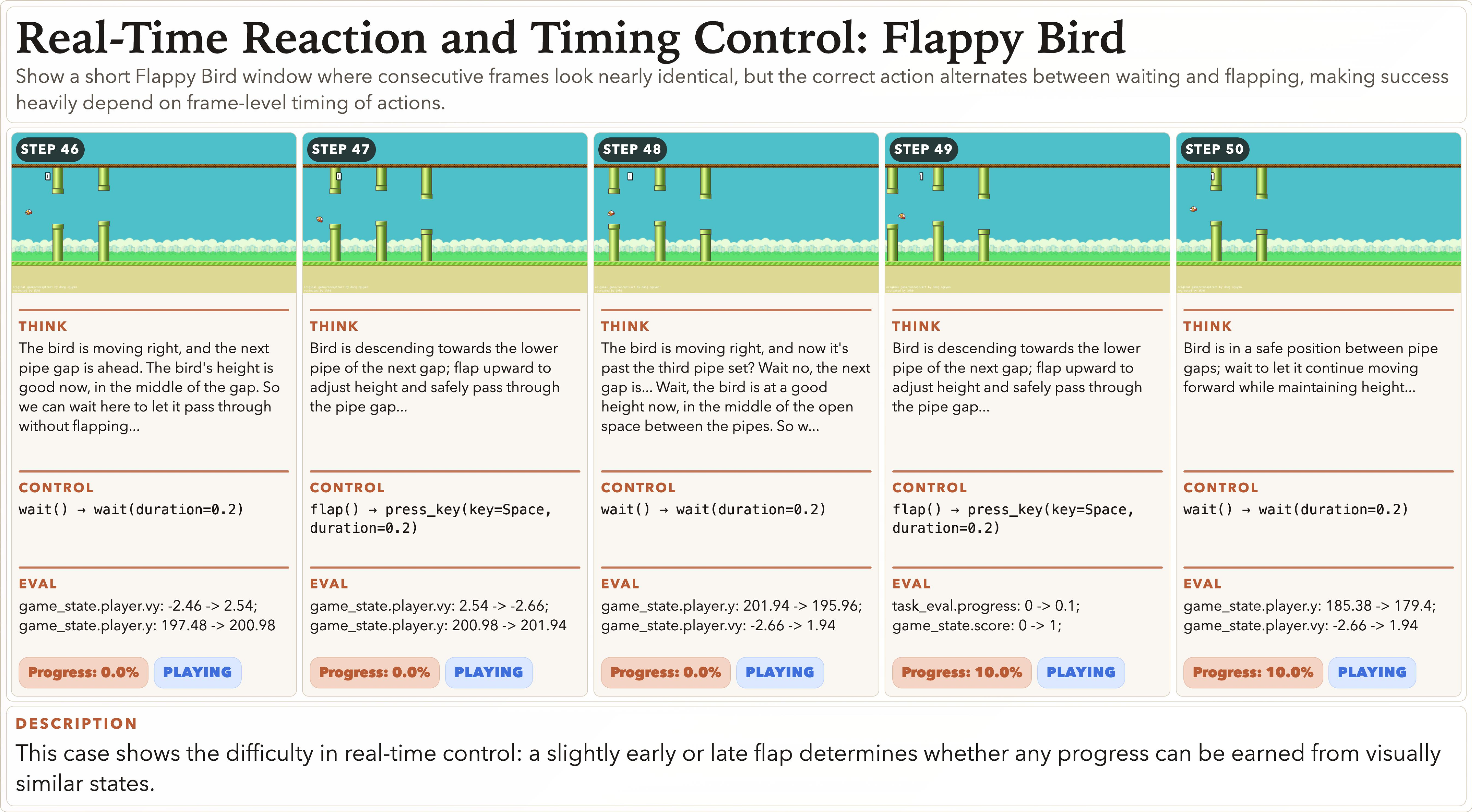}
    \caption{\captfont{Case-study visualization for real-time reaction and timing control. This \texttt{Flappy Bird} window shows how a visually small timing error can still be mechanically decisive under tight control constraints.}}
    \label{fig:case_realtime}
\end{figure}

\subsection{Long-Horizon Simulation: Minecraft Resource Collection}

Figure~\ref{fig:case_long_horizon} shows an open-ended trajectory of \texttt{Minecraft-Clone-GLM}, in which the agent repeatedly mines the resource toward the target number. The failure is not instruction-following but missing closure: the run reaches 90\% progress yet still fails to finish the collection target within the step limit.

\subsection{Real-Time Reaction and Timing Control: Flappy Bird}

As shown in Figure~\ref{fig:case_realtime}, the third case uses a short \texttt{Flappy-Bird} interaction sequence. The consecutive frames look nearly identical, while the correct action alternates between waiting and flapping. This highlights the real-time control difficulty of video games: a slightly early or late flap determines whether any progress should be inferred from visually similar states.

\section{Related Work}
\label{sec:related}

\subsection{Computer-Use Benchmarks with Online Environments}
Computer-use has emerged as a major direction in the recent rise of digital agents and serves as an important testbed for advancing agent capabilities. Its core challenge lies in designing a standardized action space and interactive environment that allow agents to flexibly control complex interfaces and operations. The design of these benchmarks largely determines how effectively agents can assist with human computer tasks and whether they can support increasingly complex and diverse workflows.
WebArena~\cite{zhou2023webarena} and OSWorld~\cite{xie2024osworld} establish strong templates for agent benchmarking in browser and desktop environments, highlighting the importance of outcome-based evaluation. Cradle~\cite{tan2024cradle} and early studies of computer-use agents~\cite{deng2023mindweb,zheng2024seeact,hu2024dawn,gao2023assistgui} further demonstrate that foundation models can operate general-purpose GUIs, even professional softwares or complex games. OSWorld-MCP~\cite{jia2025osworldmcp} extends this line by highlighting fairness issues in hybrid action pathways and tool-use decision quality. These computer-use benchmarks provide important guidance for making agent evaluation more standardized and scalable. \gameworld{} transfers these insights to game-agent evaluation through interactive environments, parallel instances, and outcome-based, state-verifiable evaluation.

\subsection{Video Game Benchmarks for LLM and MLLM Agents}
Game environments have long served as AI testbeds~\cite{cote2019textworld,beattie2016deepmind,meta2022human,xu2025deepphybenchmarkingagenticvlms}. In open-ended vision-centric games, early work focuses on training and agent construction: MineDojo~\cite{fan2022minedojo} supplies internet-scale knowledge for Minecraft, VPT~\cite{vpt} learns behavioral priors from unlabeled gameplay video, Steve-1~\cite{lifshitz2023steve} generates text-conditioned behaviors, JARVIS-1~\cite{wang2023jarvis1} adds multimodal memory for long-horizon Minecraft tasks, See and Think~\cite{zhao2023see} combines vision, language instruction, and code actions in Minecraft, and Voyager~\cite{wang2023voyager} demonstrates lifelong skill acquisition through LLM planning. More recent work also uses gameplay itself as a learning signal rather than only an evaluation target: Game-RL~\cite{tong2025gamerl} synthesizes verifiable game tasks for RL, while Play to Generalize~\cite{xie2025playtogeneralize} shows that post-training on arcade-style gameplay can transfer to broader multimodal reasoning benchmarks. As models grow more capable, the bottleneck shifts from training to reliable evaluation. MCU~\cite{zheng2025mcu} scales open-ended Minecraft evaluation through compositional atomic tasks and human-aligned assessment. LMGame-Bench~\cite{hu2025lmgamebench} exposes prompt sensitivity by modularly toggling perception, memory, and reasoning. BALROG~\cite{paglieri2025balrog}, LVLM-Playground~\cite{wang2025goodgameplayers}, and V-MAGE~\cite{zheng2025v} stress long-horizon, structured, or vision-centric reasoning in interactive games. VideoGameBench~\cite{zhang2025videogamebench} shows that inference latency dominates real-time failure and introduces a paused track to isolate it. FlashAdventure~\cite{ahn2025flashadventure} focuses on full-story-arc completion in 34 Flash adventure games and introduces CUA-as-a-Judge for automated milestone verification. Orak~\cite{park2025orak} provides a fine-tuning pipeline with held-out cross-game transfer studies. Specialized benchmarks further target collaboration or downstream workflows: Collab-Overcooked~\cite{sun2025collabovercooked} evaluates language-mediated multi-agent coordination, while VideoGameQA-Bench~\cite{taesiri2025videogameqa} measures game QA tasks such as visual regression, glitch detection, and bug-report generation. 
Notably, concurrent work GameVerse~\cite{zhang2026gameverse} shares our core idea of combining semantic and GUI control in a dual action space, and further introduces a reflect-and-retry protocol based on failure trajectories and tutorials. However, it still faces challenges in heuristic evaluation with using VLMs to quantify progress.

\subsection{Game Agents and Scalable Infrastructure}
Generalist agents can already accomplish a wide range of tasks in digital worlds. To move toward real-world embodied agents, researchers increasingly test agent capabilities in games and simulated environments.
A parallel line of work builds generalist agents that operate across many games. Game-TARS~\cite{wang2025gametars} anchors all actions to native keyboard-mouse inputs, while Jarvis-VLA~\cite{li2025jarvis} shows that large vision-language models can be post-trained to act directly through the same interface. WebGym~\cite{bai2026webgym} scales training environments to 300K realistic web tasks, showing that environment diversity directly improves out-of-distribution agent performance. NitroGen~\cite{magne2025nitrogen} extracts action labels from internet-scale gameplay video and wraps games with a universal Gym-style API~\cite{openai2016gym}. Lumine~\cite{tan2025lumine} unifies perception, reasoning, and action in one vision-language model that transfers across 3D open worlds with game-specific fine-tuning. The SIMA project evolves from instruction-following across many simulated worlds in SIMA~\cite{simateam2024scaling} to the richer interactive-partner setting of SIMA~2~\cite{simateam2025sima2}, with held-out-environment evaluation and broader user interaction. As these game agents of different interfaces mature, standardized benchmarks become essential for measuring their performance. To this end, \gameworld{} provides a comprehensive benchmark with diverse games and tasks under a unified and verifiable evaluation protocol.

\section{Conclusion}
\label{sec:conclusion}

\paragraph{Discussion.}
Our results show that current multimodal game agents can often make partial progress, yet still struggle to convert that progress into reliable task completion across diverse browser games. Under one shared runtime and verifier, \gameworld{} further exposes interface-conditioned weaknesses in real-time interaction, context-memory sensitivity, and action validity. These findings suggest that stronger game agents will require not only better reasoning, but also more reliable action grounding, more useful trajectory memory, and greater robustness to latency. We hope \gameworld{} serves as a reproducible benchmark for measuring such progress under standardized, outcome-based state-verifiable evaluation.

\paragraph{Limitations and Future Work.} 
The benchmark necessitates designing unique instruction sets for each new environment, which tightly couples the action space to the task and constrains the model's scalability. Automating the producing and alignment process of Semantic Action Parsing through MLLM-powered agent exploration is left for future work. Further discussions, including the guideline on licensing and compliance of our benchmark and the cost summary, are detailed in Appendix \ref{sec:app_licensing}.

\paragraph{Conclusion.} 
\gameworld{} provides a standardized and verifiable benchmark for evaluating multimodal game agents in browser environments. Across 34 games, 170 tasks, and 18 model--interface pairs, our results show that current agents can often make meaningful partial progress yet remain far from reliable task completion and human-level performance. Together with robustness, real-time, context-memory, and action-validity analyses, these findings establish \gameworld{} as a reproducible foundation for studying multimodal agents in complex, open-ended interactive environments.

\addtocontents{toc}{\protect\setcounter{tocdepth}{1}}
\clearpage
\beginappendix
\lstdefinestyle{prompttemplate}{
    basicstyle=\ttfamily\scriptsize\linespread{0.8}\selectfont,
    breaklines=true,
    breakatwhitespace=false,
    breakautoindent=false,
    breakindent=0pt,
    columns=fullflexible,
    keepspaces=true,
    frame=single
}

\lstdefinestyle{gamepromptblock}{
    basicstyle=\ttfamily\scriptsize\linespread{0.8}\selectfont,
    breaklines=true,
    breakatwhitespace=false,
    columns=fullflexible,
    keepspaces=true,
    frame=single,
    framesep=3pt,
    aboveskip=0.25em,
    belowskip=0.25em
}

\definecolor{jsonkey}{HTML}{2B59C3}
\definecolor{jsonstring}{HTML}{0B7A75}
\definecolor{jsonnumber}{HTML}{D94A7A}
\definecolor{jsonkeyword}{HTML}{7C4DFF}
\definecolor{jsondelim}{HTML}{7A3E00}
\definecolor{jsonpunct}{HTML}{7A7A7A}

\lstdefinelanguage{json}{
    basicstyle=\ttfamily\scriptsize\linespread{0.8}\selectfont,
    showstringspaces=false,
    breaklines=true,
    breakatwhitespace=false,
    columns=fullflexible,
    keepspaces=true,
    upquote=true,
    morestring=[b]",
    stringstyle=\color{jsonstring},
    keywordstyle=\color{jsonkeyword}\bfseries,
    keywords={true,false,null},
    moredelim=[s][\color{jsonkey}]{"}{":},
    literate=
     *{0}{{{\color{jsonnumber}0}}}{1}
      {1}{{{\color{jsonnumber}1}}}{1}
      {2}{{{\color{jsonnumber}2}}}{1}
      {3}{{{\color{jsonnumber}3}}}{1}
      {4}{{{\color{jsonnumber}4}}}{1}
      {5}{{{\color{jsonnumber}5}}}{1}
      {6}{{{\color{jsonnumber}6}}}{1}
      {7}{{{\color{jsonnumber}7}}}{1}
      {8}{{{\color{jsonnumber}8}}}{1}
      {9}{{{\color{jsonnumber}9}}}{1}
      {.}{{{\color{jsonnumber}{.}}}}{1}
      {-}{{{\color{jsonnumber}{-}}}}{1}
      {:}{{{\color{jsonpunct}{:}}}}{1}
      {,}{{{\color{jsonpunct}{,}}}}{1}
      {\{}{{{\color{jsondelim}{\{}}}}{1}
      {\}}{{{\color{jsondelim}{\}}}}}{1}
      {[}{{{\color{jsondelim}{[}}}}{1}
      {]}{{{\color{jsondelim}{]}}}}{1},
}

\lstdefinestyle{jsonblock}{
    language=json,
    basicstyle=\ttfamily\scriptsize\linespread{1}\selectfont,
    showstringspaces=false,
    breaklines=true,
    breakatwhitespace=false,
    columns=fullflexible,
    keepspaces=true,
    frame=single,
    framesep=3pt,
    rulecolor=\color{headerblue},
    backgroundcolor=\color{bggray}
}

\section{Benchmark Runtime}
\label{sec:app_runtime}

\subsection{Preset Configuration}
\label{sec:preset_config}
Each standalone benchmark run is launched using a registry preset passed via the command-line flag \texttt{--config}. The preset has the form:

\begin{center}
\small\texttt{<game\_id>+<task\_id>+<model\_spec>}
\end{center}

\noindent where the three components are resolved independently and then composed into a concrete runtime configuration:
\begin{itemize}
    \item \texttt{<game\_id>}: contributes game rules, role definitions, low-level control constraints, and semantic action definitions.
    
    \item \texttt{<task\_id>}: contributes the task instruction, evaluator configuration, target metrics, maximum step budget, and optional URL suffixes.
    
    \item \texttt{<model\_spec>}: contributes the model identifier(s), provider-specific overrides, prompt template, and output-format prompts.
\end{itemize}

This decomposition keeps game definitions, tasks, and model profiles reusable and flexible: changing a task or swapping models does not require duplicating the underlying game configuration.

\subsection{Suite Runner}

The suite runner reads a suite YAML file and expands each benchmark case into explicit runs by enumerating the specified games, tasks, and models. The resulting preset for each child run reuses the same syntax as standalone execution in Section~\ref{sec:preset_config}. Expanded runs are grouped into repeat waves. Within each wave, the runner launches up to \texttt{--max-parallel} child processes in parallel. Each child process invokes a standalone benchmark run with a dedicated port, a unique session ID, and an isolated run directory that stores logs and task-evaluation outputs. After all runs finish, the runner writes per-model summary files that support both interactive monitoring and subsequent aggregate analysis.

\section{Observation-Action-Evaluation Loop}
\label{sec:app_loop}

\subsection{Runtime Coordinator}

The benchmark loop is coordinated by a \texttt{Runtime Coordinator}.
Each MLLM is wrapped in an \texttt{Agent} object with an agent ID, model type, model client, role-specific controls, a semantic-control map, and a step counter.
The runtime also instantiates a \texttt{GameEnv} for the browser environment and the playable game, together with an \texttt{Evaluator} for task-progress tracking and evaluation.

At a high level, each interaction round executes the following sequence:
\begin{enumerate}

    \item Capture a screenshot of the current game environment.
    \item Optionally pause the game during model inference.
    \item Get the raw response from the model using the current screenshot and the assembled prompt template.
    \item Resume the game if paused.
    \item Parse the raw model output into an executable action payload.
    \item Execute the action in the browser environment.
    \item Capture a verifiable game-state snapshot.
    \item Evaluate task progress and check stopping or resetting conditions.
\end{enumerate}

\subsection{Evaluator and Reset-on-Fail}

\textbf{Evaluator.} The evaluator receives the current state snapshot, global step index, maximum step budget, target threshold, and accumulated metrics.
It combines four stop or reset signals:
\begin{itemize}
    \item terminal status from \texttt{state.terminal},
    \item exhaustion of the fixed step budget,
    \item reaching the task target score, and
    \item any task-specific end-field rule.
\end{itemize}

\noindent \textbf{Reset-on-Fail.} If \texttt{continue\_on\_fail} is enabled and the game reports a terminal failure, the runtime does not end the run immediately.
Instead, it calls \texttt{gameAPI.reset()} to reset the task, allocates a new episode ID, waits for the game to become initialized again, and continues under the same global step budget.

\section{Browser Sandbox and Game API}
\label{sec:app_browser}

\subsection{Browser Management}

The \texttt{GameLauncher} starts a local HTTP server and serves the game HTML.
The browser manager then launches Chromium with a fixed viewport and disables common background-throttling behaviors. It also injects a dynamic speed-control script and a deterministic-randomness script by overriding JavaScript headers. Screenshots are captured through the Chrome DevTools Protocol rather than through a standard page-screenshot call to avoid visible flashing in headed mode. The browser-environment initialization process includes:
\begin{itemize}
    \item Start the environment and open the game in Chromium.
    \item Wait until the game becomes actionable under the readiness gate described in Section~\ref{sec:readiness_gate}.
\end{itemize}

\subsection{Readiness Gate}
\label{sec:readiness_gate}

Before the first agent action and after every reset, the runtime waits until the game reports an actionable state. The default actionable statuses are \texttt{ready} and \texttt{playing}. Table~\ref{tab:supp_status} lists the common status values consumed by the readiness gate and the evaluator.

\begin{table}[h]
\centering
\small
\caption{Status values consumed by the readiness gate and evaluator.}
\begin{tabular}{p{0.18\linewidth}p{0.72\linewidth}}
\toprule
\textbf{Status} & \textbf{Runtime description} \\
\midrule
\texttt{loading} & Assets or engine bootstrap are not yet ready for model control. \\
\texttt{menu} & The game is initialized but still in a pre-play state such as title screen, level select, or pause menu. \\
\texttt{ready} & The game is initialized and ready to begin, but may still require one trusted start action. \\
\texttt{playing} & The gameplay loop is active and safe for model control. \\
\texttt{paused} & The game is temporarily paused; this status is used by the sandbox pause mechanism during model inference. \\
\texttt{terminal} & The current in-game episode has ended. \\
\bottomrule
\end{tabular}
\label{tab:supp_status}
\end{table}

\subsection{Verifiable State: Game API Schema}
\label{sec:app_game_api}

To enable verifiable evaluation, every benchmark game is required to expose a serializable \texttt{window.gameAPI} with three callable methods:
\texttt{init(config)}, \texttt{reset(options)}, and \texttt{getState()}.
The returned state contains a game ID, a timestamp, lifecycle status, terminal metadata, a structured \texttt{game\_state} object, task metrics, and raw game-specific details. This serves as the verifiable game state for the outcome-based evaluation.

Here we provide an example Game API schema from \texttt{17\_mario-game}. The exact game-specific fields can vary, but the top-level contract remains the same across all games in the \gameworld{} benchmark.

\begin{lstlisting}[style=jsonblock]
{
    "gameId": "17_mario-game",
    "seed": 42,
    "timestampMs": 1760001234567,
    "gameTimeMs": 18420,
    "status": "playing",
    "terminal": {
        "isTerminal": false,
        "outcome": null,
        "reason": null
    },
    "game_state": {
    "score": 3200,
    "level": "1-1",
    "progress": 0.37,
    "player": {
        "x": 128,
        "y": 80,
        "vx": 0,
        "vy": 0,
        "power": 1,
        "alive": true,
        "name": "Mario"
    },
    "board": null,
    "entities": null
    },
    "metrics": {
        "lives": 3,
        "coins": 8,
        "distance": 3200,
        "attempts": 1,
        "time_left_s": 999,
        "enemies_alive": 5,
        "level_progress_percent": 42
    },
    "raw": {
        "world": 1,
        "stage": 1,
        "levelId": "1-1",
        "worldDisplay": "1-1",
        "coins": 8,
        "coinsCollected": 8,
        "lives": 3,
        "timeLeft": 999,
        "mapTime": 999,
        "paused": false,
        "playerPower": 1,
        "playerName": "Mario",
        "levelProgress": 0.37,
        "levelProgressPercent": 37
    }
}
\end{lstlisting}

This example illustrates the benchmark convention: \texttt{game\_state} stores structured in-game state for evaluation, \texttt{metrics} stores compact comparable counters, and \texttt{raw} preserves optional game-specific details for further analysis.

\section{Agent Details}
\label{sec:app_agent_details}

\subsection{Rolling Memory}

The base client supports a rolling memory store. The memory module records each interaction round in the fixed order:
\texttt{user\_prompt $\rightarrow$ screenshot $\rightarrow$ reasoning $\rightarrow$ action}.
At inference time, the client reinjects a filtered slice of the most recent rounds according to \texttt{memory\_rounds}, \texttt{memory\_format}, and \texttt{memory\_include\_fields}. Text entries are inserted under an \texttt{Action History} header, while screenshot entries are reattached as multimodal image items, so the resulting memory context is an interleaved multimodal history rather than a pure text block.

\subsection{Low-Level Action and Validation}
\label{sec:app_action_validation}

\textbf{Low-Level Action Normalization.} All executable actions are normalized before they reach Playwright. The runtime-facing normalized action schema handled by the executor consists of:
\begin{itemize}
    \item mouse actions: \texttt{click}, \texttt{click\_hold}, \texttt{drag}, \texttt{mouse\_move}, \texttt{scroll};
    \item keyboard actions: \texttt{type}, \texttt{press\_key}, \texttt{press\_keys};
    \item timing action: \texttt{wait}.
\end{itemize}
Table~\ref{tab:agent_interfaces} defines the conceptual unified control space in terms of atomic events. The current implementation introduces a slightly higher-level normalized runtime action layer above that space for parser compatibility, legality checks, and one-action-per-step execution. Variants such as \texttt{left\_click}, \texttt{right\_click}, and \texttt{left\_click\_hold} are first normalized into this shared schema. During execution, the normalized actions are then translated into Playwright mouse and keyboard primitives such as button down/up, key down/up, wheel, move, and type events. This extra layer is also pragmatic because Playwright itself already exposes several higher-level interaction primitives. Thus the atomic event space remains the execution-layer semantics, while the runtime interface used inside the repository is slightly higher-level.

\textbf{Action Legality Validation.} Legality is role-aware and strictly configured by the role definition.
For generalist agents, the runtime coordinator first checks whether the proposed semantic control resolves against the registered semantic-control map, and then the executor checks whether the mapped low-level action satisfies the current role controls. For computer-use agents, the executor validates the proposed low-level action directly. The role controls come from the registry definition, including \texttt{allowed\_keys}, \texttt{allow\_clicks}, \texttt{hold\_duration}, and \texttt{key\_durations}. Invalid tool calls or disallowed low-level actions are logged as invalid and ignored at execution time. Key aliases are normalized (for example, \texttt{left} to \texttt{ArrowLeft}), and the executor can fall back to naming conventions when the requested key is not legal but has a semantically equivalent allowed key.
This makes the action interface more robust without expanding the legal action space beyond the role definition.

\subsection{Semantic Action Parsing}
\label{sec:app_semantic_action_parsing}

Generalist agents do not emit raw keyboard or mouse actions directly.
Instead, they emit a semantic control payload whose control identifier can arrive as \texttt{action}, \texttt{tool\_name}, or \texttt{tool\_id}, together with runtime arguments. The runtime resolves this identifier through a registry-built semantic-control map with case-insensitive and alias-aware lookup, merges the runtime arguments into the YAML-defined binding, and applies optional \texttt{cell\_bindings} when a semantic cell reference should expand into coordinates. The mapped result then enters the same low-level execution chain as computer-use actions. Unknown control identifiers are marked invalid and reduce to a no-op at execution time.

\section{Prompt Templates and Game Prompt Blocks}
\label{sec:app_prompts}

The runtime assembles the final model prompt from four pieces into a shared template: a game-level rules block, a role-and-controls block, a task-specific instruction block, and a model-specific output-format block.

\subsection{Prompt Assembly: Shared Templates}
\label{sec:app_prompt_templates}

Prompt assembly is driven by the prompt templates.
The stock templates for generalist agents and computer-use agents share a fixed section order:
\begin{enumerate}
    \item \texttt{\# Game Rules}
    \item \texttt{\# Role and Controls}
    \item \texttt{\# Task Instruction}
    \item \texttt{\# Output Format}
\end{enumerate}

For computer-use agents, the role block combines the role description with an explicit textual control specification.
For generalist agents, the role block combines the role description with an automatically rendered semantic action list.
This list is built directly from the registry \texttt{semantic\_controls} entries and therefore stays synchronized with the executable action space.

\subsubsection{Generalist Agent Template}
Below is the prompt template for generalist agents.

\begin{lstlisting}[style=prompttemplate]
You are an expert game agent specialized in playing video games. Your goal is to play the game and achieve the task goal.
Observe the current game screen to identify your character and key objects. Your action must follow the game rules and the instructions of the current task.
Execute the actions frame-by-frame.
You do NOT have direct access to keyboard or mouse actions. You must act through the registered semantic control list only.

{% if game_rules_block %}
# Game Rules
{{ game_rules_block }}
{% endif %}

{% if role_control_block_semantic %}
# Role and Controls
{{ role_control_block_semantic }}
{% endif %}

{% if task_instruction_block %}
# Task Instruction
{{ task_instruction_block }}
{% endif %}

{% if model_output_format_block %}
# Output Format
{{ model_output_format_block }}
{% endif %}
\end{lstlisting}

\subsubsection{Computer-Use Agent Template}
Below is the prompt template for computer-use agents.
\begin{lstlisting}[style=prompttemplate]
You are an expert game agent specialized in playing video games. Your goal is to play the game and achieve the task goal.
Observe the current game screen to identify your character and key objects. Your action must follow the game rules and the instructions of the current task.
Execute the actions frame-by-frame.

{% if game_rules_block %}
# Game Rules
{{ game_rules_block }}
{% endif %}

{% if role_control_block_computer_use %}
# Role and Controls
{{ role_control_block_computer_use }}
{% endif %}

{% if task_instruction_block %}
# Task Instruction
{{ task_instruction_block }}
{% endif %}

{% if model_output_format_block %}
# Output Format
{{ model_output_format_block }}
{% endif %}
\end{lstlisting}

\subsection{Per-Game Prompt Library}
\label{sec:app_prompt_library}

For each benchmark game below, we include the exact game-rules block together with the role prompt blocks loaded from the registry.
For each role, we include the shared role prompt text, the computer-use controls prompt text, and the semantic action list rendered for generalist agents from the registered \texttt{semantic\_controls} entries.
We also list the five benchmark task prompts for that game.

{
\renewcommand{\paragraph}[1]{\par\medskip\noindent{\normalfont\bfseries #1}\par\smallskip}
\paragraph{1-2048 (2048).}
Game Rules Prompt.
\begin{lstlisting}[style=gamepromptblock]
You are playing 2048, a sliding tile puzzle game.

Game Objective.
- Combine matching tiles to create higher values.
Game Rules.
- All tiles slide in the direction you press.
- When two tiles with the same number collide, they merge into one tile with doubled value.
- After each valid move (any tile is moved), a new tile of 2 or 4 appears in a random empty cell.
- Game ends when no more moves are possible (board is full and no merges available).
\end{lstlisting}
Role 0 (player) Game Agent Role Prompt.
\begin{lstlisting}[style=gamepromptblock]
You control the 2048 board. Choose exactly one action per step to slide tiles.
\end{lstlisting}
Role 0 (player) Computer-Use Controls Prompt.
\begin{lstlisting}[style=gamepromptblock]
ACTION SPACE (ONLY LEGAL ACTIONS):

- Wait: Do nothing.
- ArrowUp: Slide all tiles up
- ArrowDown: Slide all tiles down
- ArrowLeft: Slide all tiles left
- ArrowRight: Slide all tiles right
\end{lstlisting}
Role 0 (player) Generalist Semantic Action List.
\begin{lstlisting}[style=gamepromptblock]
REGISTERED ACTIONS (Semantic Controls).
Choose exactly one action per step:

- wait: Do nothing.
- move_up: Slide all tiles up.
- move_down: Slide all tiles down.
- move_left: Slide all tiles left.
- move_right: Slide all tiles right.
\end{lstlisting}

\paragraph{2-another-gentlemans-adventure (Another Gentleman's Adventure).}
Game Rules Prompt.
\begin{lstlisting}[style=gamepromptblock]
You are playing Another Gentleman's Adventure, a platformer game.

Game Objective.
- Navigate through levels, avoiding obstacles and enemies to get coins.
Game Rules.
- Move left and right to navigate platforms.
- Jump (left or right) to avoid obstacles and reach higher platforms.
- Jump from below to hit the Golden Question Blocks and reveal items.
- Jumping on enemies kills them.
\end{lstlisting}
Role 0 (player) Game Agent Role Prompt.
\begin{lstlisting}[style=gamepromptblock]
You control the gentleman character.
\end{lstlisting}
Role 0 (player) Computer-Use Controls Prompt.
\begin{lstlisting}[style=gamepromptblock]
ACTION SPACE (ONLY LEGAL ACTIONS):

- Wait: Stay still.
- Arrow Left/Right: Move horizontally
- Arrow Up: Jump
- Arrow Left/Right + Arrow Up (key combination): Jump while moving left/rights
\end{lstlisting}
Role 0 (player) Generalist Semantic Action List.
\begin{lstlisting}[style=gamepromptblock]
REGISTERED ACTIONS (Semantic Controls).
Choose exactly one action per step:

- wait: Stay still briefly.
- move_left: Walk left.
- move_right: Walk right.
- jump: Jump vertically.
- jump_left: Jump while moving left.
- jump_right: Jump while moving right.
\end{lstlisting}

\paragraph{3-astray (Astray).}
Game Rules Prompt.
\begin{lstlisting}[style=gamepromptblock]
You are playing Astray, a maze navigation game.

Game Objective.
- Navigate the ball through the maze to reach the exit.
Game Rules.
- Navigate through the maze carefully to find the path to the exit.
- The exit is typically at the right-top far corner of the maze.
\end{lstlisting}
Role 0 (player) Game Agent Role Prompt.
\begin{lstlisting}[style=gamepromptblock]
You control the ball in the maze. Use arrow keys to navigate to the exit.
\end{lstlisting}
Role 0 (player) Computer-Use Controls Prompt.
\begin{lstlisting}[style=gamepromptblock]
ACTION SPACE (ONLY LEGAL ACTIONS):

- Wait: Wait briefly without moving
- ArrowUp: Move the ball up
- ArrowDown: Move the ball down
- ArrowLeft: Move the ball left
- ArrowRight: Move the ball right
\end{lstlisting}
Role 0 (player) Generalist Semantic Action List.
\begin{lstlisting}[style=gamepromptblock]
REGISTERED ACTIONS (Semantic Controls).
Choose exactly one action per step:

- wait: Wait briefly without moving.
- move_left: Move the ball left.
- move_right: Move the ball right.
- move_up: Move the ball up.
- move_down: Move the ball down.
\end{lstlisting}

\paragraph{4-boxel-rebound (Boxel Rebound).}
Game Rules Prompt.
\begin{lstlisting}[style=gamepromptblock]
You are playing Boxel Rebound, a precision platformer.

Game Objective.
- Time your jumps to navigate through each level to reach the end without dying.
- Your character automatically moves forward continuously.
Game Rules.
- You can only jump - no left/right movement control.
- Avoid spikes and falling into pits.
\end{lstlisting}
Role 0 (player) Game Agent Role Prompt.
\begin{lstlisting}[style=gamepromptblock]
You control the boxel character. Time your jumps to navigate through the level.
\end{lstlisting}
Role 0 (player) Computer-Use Controls Prompt.
\begin{lstlisting}[style=gamepromptblock]
ACTION SPACE (ONLY LEGAL ACTIONS):

- Wait: Do nothing to continue movement.
- Space or ArrowUp: Jump
\end{lstlisting}
Role 0 (player) Generalist Semantic Action List.
\begin{lstlisting}[style=gamepromptblock]
REGISTERED ACTIONS (Semantic Controls).
Choose exactly one action per step:

- wait: Do nothing to continue movement.
- jump: Jump to avoid obstacles or cross gaps.
\end{lstlisting}

\paragraph{5-breakout (Breakout).}
Game Rules Prompt.
\begin{lstlisting}[style=gamepromptblock]
You are playing Breakout, a classic brick-breaking game.

Game Objective.
- Break all the bricks to complete each level. Each brick broken adds to your score.
Game Rules.
- Use the paddle to bounce the ball upward.
- Break bricks by hitting them with the ball.
- Don't let the ball fall past the paddle.
\end{lstlisting}
Role 0 (player) Game Agent Role Prompt.
\begin{lstlisting}[style=gamepromptblock]
You control the paddle. Move left and right to keep the ball in play and break bricks.
\end{lstlisting}
Role 0 (player) Computer-Use Controls Prompt.
\begin{lstlisting}[style=gamepromptblock]
ACTION SPACE (ONLY LEGAL ACTIONS):

- Wait: Stay still.
- Arrow Left/Right: Move paddle
- Space: Launch ball/Start game
\end{lstlisting}
Role 0 (player) Generalist Semantic Action List.
\begin{lstlisting}[style=gamepromptblock]
REGISTERED ACTIONS (Semantic Controls).
Choose exactly one action per step:

- wait: Stay still briefly.
- move_left: Move paddle left.
- move_right: Move paddle right.
\end{lstlisting}

\paragraph{6-captaincallisto (Captain Callisto).}
Game Rules Prompt.
\begin{lstlisting}[style=gamepromptblock]
You are playing Captain Callisto, a platform adventure.

Game Objective.
- Reach the exit of each area.
- Collect coins along the way.
Game Rules.
- Move left/right/up/down and jump across platforms.
- Use the jetpack to fly up after you collected the fuels.
- Kill the enemies by jumping on them.
- When you touch the enemies (not from above) or falling, you will respawn.
- You need a move to start the level after respawning.
\end{lstlisting}
Role 0 (player) Game Agent Role Prompt.
\begin{lstlisting}[style=gamepromptblock]
You control Captain Callisto. Control the player to move, jump, and reach the exit.
\end{lstlisting}
Role 0 (player) Computer-Use Controls Prompt.
\begin{lstlisting}[style=gamepromptblock]
ACTION SPACE (ONLY LEGAL ACTIONS):

- Wait: Do nothing.
- ArrowLeft / ArrowRight or A/D: move left/right
- ArrowUp / ArrowDown or W/S: climb or move vertically
- Space: jump
- Shift: jetpack
- r: restart
\end{lstlisting}
Role 0 (player) Generalist Semantic Action List.
\begin{lstlisting}[style=gamepromptblock]
REGISTERED ACTIONS (Semantic Controls).
Choose exactly one action per step:

- wait: Do nothing.
- move_left: Move left.
- move_right: Move right.
- move_up: Move up.
- move_down: Move down.
- jump: Jump over obstacles or gaps.
- jetpack: Use the jetpack to fly up (only if available)
- restart: Restart the level. (You will need a move to start the level after respawning)
\end{lstlisting}

\paragraph{7-chrome-dino (Chrome Dino).}
Game Rules Prompt.
\begin{lstlisting}[style=gamepromptblock]
You are playing the Chrome Dino runner game. Jump to start the run.

Game Objective.
- Survive as long as possible while the dino runs.
- Avoid obstacles to keep the run going and increase score.
\end{lstlisting}
Role 0 (player) Game Agent Role Prompt.
\begin{lstlisting}[style=gamepromptblock]
You control the dinosaur runner. Jump to avoid obstacles and survive.
\end{lstlisting}
Role 0 (player) Computer-Use Controls Prompt.
\begin{lstlisting}[style=gamepromptblock]
ACTION SPACE (ONLY LEGAL ACTIONS):

- Wait: Do nothing to keep the run going.
- Space/ArrowUp: Jump or start the run
\end{lstlisting}
Role 0 (player) Generalist Semantic Action List.
\begin{lstlisting}[style=gamepromptblock]
REGISTERED ACTIONS (Semantic Controls).
Choose exactly one action per step:

- wait: Do nothing briefly to keep the run going.
- jump: Jump over obstacles.
- start: Start or restart the run.
\end{lstlisting}

\paragraph{8-core-ball (Core Ball).}
Game Rules Prompt.
\begin{lstlisting}[style=gamepromptblock]
You are playing Core Ball, a high-speed 2D timing-and-precision game. Shoot numbered balls into a rotating core and attach them to it without collisions.

Game Objective.
- Attach all balls for the current level to the core.
- Complete the level before the queue is exhausted and the core becomes too crowded.
- Levels get faster and denser as you progress.
Game Rules.
- Press Space to shoot one ball toward the center.
- A ball collision with another ball fails the run immediately.
\end{lstlisting}
Role 0 (player) Game Agent Role Prompt.
\begin{lstlisting}[style=gamepromptblock]
You control precision ball launches toward the spinning core. Time each shot carefully and keep the path clear.
\end{lstlisting}
Role 0 (player) Computer-Use Controls Prompt.
\begin{lstlisting}[style=gamepromptblock]
ACTION SPACE (ONLY LEGAL ACTIONS):

- Wait: Do nothing to wait the core to rotate
- Space: shoot one ball
\end{lstlisting}
Role 0 (player) Generalist Semantic Action List.
\begin{lstlisting}[style=gamepromptblock]
REGISTERED ACTIONS (Semantic Controls).
Choose exactly one action per step:

- wait: Pause briefly and observe target alignment.
- shoot: Shoot one ball toward the core.
\end{lstlisting}

\paragraph{9-cubefield (Cubefield).}
Game Rules Prompt.
\begin{lstlisting}[style=gamepromptblock]
You are playing Cubefield, a 3D endless runner where you navigate a triangular ship through a field of cubes.

Game Objective.
- Survive as long as possible by avoiding cube obstacles.
Game Rules.
- Move left and right to dodge cubes.
- The ship automatically moves forward.
- Hitting any cube ends the game instantly.
\end{lstlisting}
Role 0 (player) Game Agent Role Prompt.
\begin{lstlisting}[style=gamepromptblock]
You control the triangular ship. Move left or right to navigate through the cube field.
\end{lstlisting}
Role 0 (player) Computer-Use Controls Prompt.
\begin{lstlisting}[style=gamepromptblock]
ACTION SPACE (ONLY LEGAL ACTIONS):

- Wait: Continue straight.
- Arrow Left: Move left
- Arrow Right: Move right
\end{lstlisting}
Role 0 (player) Generalist Semantic Action List.
\begin{lstlisting}[style=gamepromptblock]
REGISTERED ACTIONS (Semantic Controls).
Choose exactly one action per step:

- wait: Continue straight.
- move_left: Move left to avoid cubes.
- move_right: Move right to avoid cubes.
\end{lstlisting}

\paragraph{10-doodle-jump (Doodle Jump).}
Game Rules Prompt.
\begin{lstlisting}[style=gamepromptblock]
You are playing Doodle Jump, a vertical platformer.

Game Objective.
- Keep ascending by landing on higher platforms.
Game Rules.
- Green platforms: normal platforms.
- Blue movable platforms: moving left and right.
- Red breakable platforms: if you step on it, it will break and you will fall.
- White vanishable platforms: can step on it once, it will vanish.
- The left and right sides wrap around each other.
\end{lstlisting}
Role 0 (player) Game Agent Role Prompt.
\begin{lstlisting}[style=gamepromptblock]
You control Doodle. Control the player to drift and avoid falling.
\end{lstlisting}
Role 0 (player) Computer-Use Controls Prompt.
\begin{lstlisting}[style=gamepromptblock]
ACTION SPACE (ONLY LEGAL ACTIONS):

- Wait: Do nothing to continue vertical movement.
- ArrowLeft / ArrowRight: move horizontally.
\end{lstlisting}
Role 0 (player) Generalist Semantic Action List.
\begin{lstlisting}[style=gamepromptblock]
REGISTERED ACTIONS (Semantic Controls).
Choose exactly one action per step:

- wait: Do nothing to continue vertical movement.
- move_left: Drift left.
- move_right: Drift right.
\end{lstlisting}

\paragraph{11-edge-surf (Edge Surf).}
Game Rules Prompt.
\begin{lstlisting}[style=gamepromptblock]
You are playing Edge Surf, an endless surfing game. You control a surfer riding waves and collecting items.

Game Objective.
- Survive as long as possible while surfing.
- Achieve the highest score and distance with 3 lives.
Game Rules.
- You can only turn left or right, slow down or boost the surfer.
- Hitting obstacles costs lives.
- Collecting boosts gives temporary speed, and shields gives temporary invincibility.
\end{lstlisting}
Role 0 (player) Game Agent Role Prompt.
\begin{lstlisting}[style=gamepromptblock]
You control the surfer character. Turn left and right to avoid obstacles and collect items.
\end{lstlisting}
Role 0 (player) Computer-Use Controls Prompt.
\begin{lstlisting}[style=gamepromptblock]
ACTION SPACE (ONLY LEGAL ACTIONS):

- Wait: Continue surfing in the same direction without action.
- Arrow Left: Turn left
- Arrow Right: Turn right
- Arrow Up: Slow down the surfer
- Arrow Down: Boost the surfer (requires a boost item)
\end{lstlisting}
Role 0 (player) Generalist Semantic Action List.
\begin{lstlisting}[style=gamepromptblock]
REGISTERED ACTIONS (Semantic Controls).
Choose exactly one action per step:

- wait: Continue surfing in the same direction.
- turn_left: Turn left.
- turn_right: Turn right.
- slow_down: Slow down the surfer.
- boost: Boost the surfer (requires a boost item).
\end{lstlisting}

\paragraph{12-fireboy-and-watergirl (Fireboy and Watergirl).}
Game Rules Prompt.
\begin{lstlisting}[style=gamepromptblock]
You are playing 'Fireboy and Watergirl in The Forest Temple'. Some puzzles require both characters to cooperate to solve.

Game Objective.
- Both characters must reach their respective exit doors to complete the level.
- Collect all diamonds for a higher score.
Game Rules.
- Green toxic liquid: Kills BOTH characters
- Red lava pools: Kills Watergirl only
- Blue water pools: Kills Fireboy only
- Buttons and levers control platforms and doors
- Boxes can be pushed to reach higher platforms
\end{lstlisting}
Role 0 (watergirl) Game Agent Role Prompt.
\begin{lstlisting}[style=gamepromptblock]
You are Agent 0 controlling the Watergirl character (blue girl).

YOUR OBJECTIVES:

- Collect blue diamonds (Watergirl's gems)
- Reach the blue exit door
- NEVER touch red lava (instant death)
\end{lstlisting}
Role 0 (watergirl) Computer-Use Controls Prompt.
\begin{lstlisting}[style=gamepromptblock]
ACTION SPACE (ONLY LEGAL ACTIONS):

- W: Jump
- A: Move left
- D: Move right
- W + A: Jump left
- W + D: Jump right
\end{lstlisting}
Role 0 (watergirl) Generalist Semantic Action List.
\begin{lstlisting}[style=gamepromptblock]
REGISTERED ACTIONS (Semantic Controls).
Choose exactly one action per step:

- wait: Stay still briefly.
- move_left: Walk left (hold).
- move_right: Walk right (hold).
- jump_left: Jump diagonally left (jump + left).
- jump_right: Jump diagonally right (jump + right).
\end{lstlisting}
Role 1 (fireboy) Game Agent Role Prompt.
\begin{lstlisting}[style=gamepromptblock]
You are Agent 1 controlling the Fireboy character (red boy).

YOUR OBJECTIVES:

- Collect red diamonds (Fireboy's gems)
- Reach the red exit door
- NEVER touch blue water (instant death)
\end{lstlisting}
Role 1 (fireboy) Computer-Use Controls Prompt.
\begin{lstlisting}[style=gamepromptblock]
ACTION SPACE (ONLY LEGAL ACTIONS):

- ArrowUp: Jump
- ArrowLeft: Move left
- ArrowRight: Move right
- ArrowUp + ArrowLeft: Jump left
- ArrowUp + ArrowRight: Jump right
\end{lstlisting}
Role 1 (fireboy) Generalist Semantic Action List.
\begin{lstlisting}[style=gamepromptblock]
REGISTERED ACTIONS (Semantic Controls).
Choose exactly one action per step:

- wait: Stay still briefly.
- move_left: Walk left (hold).
- move_right: Walk right (hold).
- jump_left: Jump diagonally left (jump + left).
- jump_right: Jump diagonally right (jump + right).
\end{lstlisting}

\paragraph{13-flappy-bird (Flappy Bird).}
Game Rules Prompt.
\begin{lstlisting}[style=gamepromptblock]
You are playing Flappy Bird, a one-button flying game.

Game Objective.
- Keep the bird flying between pipes for as long as possible.
- Pass pipes to increase your score.
Game Rules.
- Press Space to flap upward; gravity pulls you down.
- Hitting pipes or the ground ends the run.
\end{lstlisting}
Role 0 (player) Game Agent Role Prompt.
\begin{lstlisting}[style=gamepromptblock]
You control the bird. Tap to flap and avoid the pipes.
\end{lstlisting}
Role 0 (player) Computer-Use Controls Prompt.
\begin{lstlisting}[style=gamepromptblock]
ACTION SPACE (ONLY LEGAL ACTIONS):

- Space: Flap upward
- Wait: Do nothing briefly to let gravity pull bird downward
\end{lstlisting}
Role 0 (player) Generalist Semantic Action List.
\begin{lstlisting}[style=gamepromptblock]
REGISTERED ACTIONS (Semantic Controls).
Choose exactly one action per step:

- wait: Do nothing briefly to let gravity pull bird downward
- flap: Flap upward.
\end{lstlisting}

\paragraph{14-geodash (GeoDash).}
Game Rules Prompt.
\begin{lstlisting}[style=gamepromptblock]
You are playing GeoDash, a Geometry Dash-style auto-running platformer.

Game Objective.
- Survive for as long as possible.
- Jump over spikes and hazards.
Game Rules.
- The character auto-runs continuously.
- Jump timing is the core mechanic.
- Crashing into obstacles ends the run.
\end{lstlisting}
Role 0 (player) Game Agent Role Prompt.
\begin{lstlisting}[style=gamepromptblock]
You control the runner. Time jumps to avoid spikes and run as long as possible.
\end{lstlisting}
Role 0 (player) Computer-Use Controls Prompt.
\begin{lstlisting}[style=gamepromptblock]
ACTION SPACE (ONLY LEGAL ACTIONS):

- Space
\end{lstlisting}
Role 0 (player) Generalist Semantic Action List.
\begin{lstlisting}[style=gamepromptblock]
REGISTERED ACTIONS (Semantic Controls).
Choose exactly one action per step:

- wait: Do nothing briefly to keep the direction.
- jump: Jump.
\end{lstlisting}

\paragraph{15-google-snake (Google Snake).}
Game Rules Prompt.
\begin{lstlisting}[style=gamepromptblock]
You are playing Google Snake, the classic snake game.

Game Objective.
- Eat apples to grow longer and increase score.
- Avoid crashing into walls or your own body.
Game Rules.
- The snake moves continuously in the chosen direction.
- You can turn up, down, left, or right.
- The game ends when you hit a wall or your body.
\end{lstlisting}
Role 0 (player) Game Agent Role Prompt.
\begin{lstlisting}[style=gamepromptblock]
You control the snake. Use direction keys to guide it toward apples safely.
\end{lstlisting}
Role 0 (player) Computer-Use Controls Prompt.
\begin{lstlisting}[style=gamepromptblock]
ACTION SPACE (ONLY LEGAL ACTIONS):

- ArrowUP: Turn upward
- ArrowDown: Turn downward
- ArrowLeft: Turn left
- ArrowRight: Turn right
- Wait: Do nothing briefly to keep the direction
\end{lstlisting}
Role 0 (player) Generalist Semantic Action List.
\begin{lstlisting}[style=gamepromptblock]
REGISTERED ACTIONS (Semantic Controls).
Choose exactly one action per step:

- wait: Do nothing briefly to keep the direction.
- move_up: Turn upward.
- move_down: Turn downward.
- move_left: Turn left.
- move_right: Turn right.
\end{lstlisting}

\paragraph{16-hextris (Hextris).}
Game Rules Prompt.
\begin{lstlisting}[style=gamepromptblock]
You are playing Hextris, a fast-paced hexagon-based puzzle game.

Game Objective.
- Matching Mechanism: Group 3 or more blocks of the same color on any of the six sides to clear them and get scores.
- Prevent blocks from stacking outside the outer boundary of the central hexagon.
Game Rules.
- The game runs continuously and blocks fall from the edges into the central hexagon.
- The speed of the falling blocks increases over time.
- Left/Right arrows rotate the hexagon in opposite directions.
- Press Down to speed up falling block motion.
\end{lstlisting}
Role 0 (player) Game Agent Role Prompt.
\begin{lstlisting}[style=gamepromptblock]
You control the rotating hexagon. Keep the board manageable by rotating quickly and balancing incoming blocks.
\end{lstlisting}
Role 0 (player) Computer-Use Controls Prompt.
\begin{lstlisting}[style=gamepromptblock]
ACTION SPACE (ONLY LEGAL ACTIONS):

- ArrowLeft: rotate hexagon left
- ArrowRight: rotate hexagon right
- ArrowDown: accelerate falling blocks
\end{lstlisting}
Role 0 (player) Generalist Semantic Action List.
\begin{lstlisting}[style=gamepromptblock]
REGISTERED ACTIONS (Semantic Controls).
Choose exactly one action per step:

- wait: Briefly wait while the board evolves.
- rotate_left: Rotate the hexagon left once.
- rotate_right: Rotate the hexagon right once.
- accelerate: Temporarily speed up falling blocks.
\end{lstlisting}

\paragraph{17-mario-game (Mario Game).}
Game Rules Prompt.
\begin{lstlisting}[style=gamepromptblock]
You are playing Super Mario Bros. Control Mario through platforming levels, collecting coins and defeating enemies.

Game Objective.
- Reach the flagpole at the end of each level.
- Collect coins from question blocks or kill enemies for points.
Game Rules.
- You have 3 lives and limited time per level.
- Touching enemies from the side or below kills Mario. Jump on enemies to kill enemies.
- Jumping underneath question blocks will reveal coins or power-ups.
- Power-ups: Mushroom make Mario grow big.
\end{lstlisting}
Role 0 (player) Game Agent Role Prompt.
\begin{lstlisting}[style=gamepromptblock]
You control Mario. Navigate through platforming levels by running, jumping, and avoiding hazards.
\end{lstlisting}
Role 0 (player) Computer-Use Controls Prompt.
\begin{lstlisting}[style=gamepromptblock]
ACTION SPACE (ONLY LEGAL ACTIONS):

- ArrowLeft: Move left
- ArrowRight: Move right
- ArrowUp: Jump
- ArrowDown: Duck/crouch (when big)
- ArrowUp + ArrowLeft/ArrowRight: Jump while moving left/right
\end{lstlisting}
Role 0 (player) Generalist Semantic Action List.
\begin{lstlisting}[style=gamepromptblock]
REGISTERED ACTIONS (Semantic Controls).
Choose exactly one action per step:

- wait: Stand still and wait.
- move_right: Walk right.
- move_left: Walk left.
- jump: Jump straight up.
- jump_right: Jump while moving right.
- jump_left: Jump while moving left.
- duck: Duck/crouch (when big).
\end{lstlisting}

\paragraph{18-minecraft-clone-glm (Minecraft Clone).}
Game Rules Prompt.
\begin{lstlisting}[style=gamepromptblock]
You are playing a Minecraft-style first-person sandbox survival game.

Game Objective.
- Gather resources by breaking blocks.
- Move around and mine nearby blocks efficiently.
Game Rules.
- A short left click places the currently selected block.
- Hold the left mouse button on the targeted block to mine it. Hold-mining is accelerated in benchmark mode.
- You can also place blocks with the right mouse button.
- Move around and jump to navigate the terrain.
- Look around to aim at different blocks.
- Select different hotbar slots to switch items.
\end{lstlisting}
Role 0 (player) Game Agent Role Prompt.
\begin{lstlisting}[style=gamepromptblock]
You control the player character. Prioritize efficient nearby resource collection by moving to blocks and mining them.
\end{lstlisting}
Role 0 (player) Computer-Use Controls Prompt.
\begin{lstlisting}[style=gamepromptblock]
ACTION SPACE (ONLY LEGAL ACTIONS):

- ArrowUp: move forward
- ArrowDown: move backward
- ArrowLeft: strafe left
- ArrowRight: strafe right
- Space: jump
- Mouse drag: look around / turn camera (drag from center toward a direction)
- Left click: place the selected block
- Left click and hold: mine the targeted block
- Right mouse: place selected block
- 1-9: select hotbar slot
\end{lstlisting}
Role 0 (player) Generalist Semantic Action List.
\begin{lstlisting}[style=gamepromptblock]
REGISTERED ACTIONS (Semantic Controls).
Choose exactly one action per step:

- wait: Wait briefly.
- move_forward: Move forward briefly.
- move_backward: Move backward briefly.
- strafe_left: Move left briefly.
- strafe_right: Move right briefly.
- jump: Jump once.
- look_left: Turn camera left.
- look_right: Turn camera right.
- look_up: Tilt camera up.
- look_down: Tilt camera down.
- mine_target: Hold left click at center to mine the targeted block.
- place_block: Place the selected block at center.
- select_slot_1: Select hotbar slot 1.
- select_slot_2: Select hotbar slot 2.
- select_slot_3: Select hotbar slot 3.
\end{lstlisting}

\paragraph{19-minesweeper (Minesweeper).}
Game Rules Prompt.
\begin{lstlisting}[style=gamepromptblock]
You are playing Minesweeper, a logic puzzle.

Game Objective.
- Flag all mines.
Game Rules.
- Left click a cell reveals it; right click flags a mine.
- A number shows how many mines are adjacent to that cell.
- Clicking a mine ends the game.
\end{lstlisting}
Role 0 (player) Game Agent Role Prompt.
\begin{lstlisting}[style=gamepromptblock]
You control the board to reveal safe cells and avoid mines.
\end{lstlisting}
Role 0 (player) Computer-Use Controls Prompt.
\begin{lstlisting}[style=gamepromptblock]
ACTION SPACE (ONLY LEGAL ACTIONS):

- Mouse left click: Reveal a cell
- Mouse right click: Flag a mine
\end{lstlisting}
Role 0 (player) Generalist Semantic Action List.
\begin{lstlisting}[style=gamepromptblock]
REGISTERED ACTIONS (Semantic Controls).
Choose exactly one action per step:

- wait: Pause briefly.
- reveal_cell: Reveal a cell by id (cell="a1".."i9"). (required: cell)
- flag_cell: Flag a cell by id (cell="a1".."i9"). (required: cell)
\end{lstlisting}

\paragraph{20-monkey-mart (Monkey Mart).}
Game Rules Prompt.
\begin{lstlisting}[style=gamepromptblock]
You are playing Monkey Mart, a store management game.

Game Objective.
- Stock shelves, serve customers, and earn money.
- Expand the store by unlocking new stations and upgrades.
Game Rules.
- Move within the store to automatically harvest, carry, and restock items.
- Stay near the banana tree to harvest bananas, the corn field to harvest corn, the corresponding shelves to stock items, and most importantly, the counter to serve customers and collect green money.
- The needs of customers will pop up at the top of them, and you need to have the item they want stocked on the shelves for them to buy it.
- Customers take items and pay at the counter.
- Stay at the left of the counter to serve customers and then collect green money.
- Banana tree is at the bottom of the store, banana shelf is in the middle, corn field is at the bottom left, corn shelf is at the top of the corn field, and the counter is on the left of the banana shelf.
- You have one assistant with orange hat to collect and restock corns, but they will occasionally be idle, you need to wake them.
- Assistant will only help collect and restock corns. Therefore, you need to collect and restock bananas, and collect green money by yourself.
- You should interleeave between harvesting, stocking, and serving to keep the store running efficiently. For example, if you keep harvesting and restocking, but never serve customers at the counter, you won't collect any money. On the other hand, if you only stay at the counter to serve customers without restocking, the shelves will run out of stock and customers will wait for restocking.
- Use earnings to unlock new areas and assistants.
\end{lstlisting}
Role 0 (player) Game Agent Role Prompt.
\begin{lstlisting}[style=gamepromptblock]
You control the monkey character. Move around the store to stock and manage items.
\end{lstlisting}
Role 0 (player) Computer-Use Controls Prompt.
\begin{lstlisting}[style=gamepromptblock]
ACTION SPACE (ONLY LEGAL ACTIONS):

- ArrowUp: Move up
- ArrowDown: Move down
- ArrowLeft: Move left
- ArrowRight: Move right
- wait: Stay still to wait for assistants and customers
\end{lstlisting}
Role 0 (player) Generalist Semantic Action List.
\begin{lstlisting}[style=gamepromptblock]
REGISTERED ACTIONS (Semantic Controls).
Choose exactly one action per step:

- wait: Stay still to wait for assistants and customers.
- move_up: Move up.
- move_down: Move down.
- move_left: Move left.
- move_right: Move right.
\end{lstlisting}

\paragraph{21-ns-shaft (NS-Shaft).}
Game Rules Prompt.
\begin{lstlisting}[style=gamepromptblock]
You are playing NS-Shaft, a falling platform game.

Game Objective.
- Descend as far as possible by landing on platforms.
- Avoid hazards while keeping your character alive.
Game Rules.
- The character falls downward automatically.
- Move left/right to land on platforms.
- Life decreases when you touch the pillards
- You will die if you fall all the way to the bottom or run out of the life.
\end{lstlisting}
Role 0 (player) Game Agent Role Prompt.
\begin{lstlisting}[style=gamepromptblock]
You control the falling character. Move left and right to land on platforms safely.
\end{lstlisting}
Role 0 (player) Computer-Use Controls Prompt.
\begin{lstlisting}[style=gamepromptblock]
ACTION SPACE (ONLY LEGAL ACTIONS):

- Arrow Left/Right: Move left/right
\end{lstlisting}
Role 0 (player) Generalist Semantic Action List.
\begin{lstlisting}[style=gamepromptblock]
REGISTERED ACTIONS (Semantic Controls).
Choose exactly one action per step:

- wait: Do nothing briefly.
- move_left: Move left.
- move_right: Move right.
\end{lstlisting}

\paragraph{22-ovo (OVO).}
Game Rules Prompt.
\begin{lstlisting}[style=gamepromptblock]
You are playing OVO, a fast platformer with traps.

Game Objective.
- Reach the exit of each level.
- Collect coins while avoiding hazards.
Game Rules.
- Avoid traps and pits.
- When you go next to the wall, you can jump higher.
- Jump on place and press down to smash the ground.
\end{lstlisting}
Role 0 (player) Game Agent Role Prompt.
\begin{lstlisting}[style=gamepromptblock]
You control the character in OVO. Control the player to run and jump through levels.
\end{lstlisting}
Role 0 (player) Computer-Use Controls Prompt.
\begin{lstlisting}[style=gamepromptblock]
ACTION SPACE (ONLY LEGAL ACTIONS):

- ArrowLeft / ArrowRight: move left/right.
- ArrowUp: jump.
- ArrowUp + ArrowRight / ArrowLeft: jump while moving right / left.
- ArrowDown: smash the ground.
- ArrowDown + ArrowRight / ArrowLeft: slide right / left.
\end{lstlisting}
Role 0 (player) Generalist Semantic Action List.
\begin{lstlisting}[style=gamepromptblock]
REGISTERED ACTIONS (Semantic Controls).
Choose exactly one action per step:

- wait: Pause briefly.
- move_left: Run left.
- move_right: Run right.
- jump: Jump.
- jump_right: Jump while moving right.
- jump_left: Jump while moving left.
- smash_ground: Smash the ground.
- slide_right: Slide right.
- slide_left: Slide left.
\end{lstlisting}

\paragraph{23-pacman (Pac-Man).}
Game Rules Prompt.
\begin{lstlisting}[style=gamepromptblock]
You are playing Pac-Man, a maze chase game.

Game Objective.
- Eat all pellets in the maze.
- Avoid ghosts, or eat them after grabbing a power pellet.
TIPS:

- Use corridors to bait ghosts into corners.
- Save power pellets for tricky sections.
\end{lstlisting}
Role 0 (player) Game Agent Role Prompt.
\begin{lstlisting}[style=gamepromptblock]
You control Pac-Man. Control the player to move through the maze.
\end{lstlisting}
Role 0 (player) Computer-Use Controls Prompt.
\begin{lstlisting}[style=gamepromptblock]
ACTION SPACE (ONLY LEGAL ACTIONS):

- ArrowUp: Move up through the maze
- ArrowDown: Move down through the maze
- ArrowLeft: Move left through the maze
- ArrowRight: Move right through the maze
\end{lstlisting}
Role 0 (player) Generalist Semantic Action List.
\begin{lstlisting}[style=gamepromptblock]
REGISTERED ACTIONS (Semantic Controls).
Choose exactly one action per step:

- wait: Pause briefly.
- move_up: Move up.
- move_down: Move down.
- move_left: Move left.
- move_right: Move right.
\end{lstlisting}

\paragraph{24-restless-wing-syndrome (Restless Wing Syndrome).}
Game Rules Prompt.
\begin{lstlisting}[style=gamepromptblock]
You are playing Restless Wing Syndrome, a platformer with automatic flaps.

Game Objective.
- Reach the exit (looks like a bread) of each level.
- Avoid spikes and hazards.
Game Rules.
- The bird flaps automatically on a timer (flap meter at top left). The flap meter will automatically decrease, one cell at a time. When it is empty, the bird will automatically jump up then the meter will be recharged to full. You can observe the flap meter to time your movements.
- You can move left or right to steer mid-air.
- Hold Up to glide (slow descent) in the air.
- Use Up+Left or Up+Right to glide diagonally.
- Most of the time the movements are achieved by gliding with appropriate direction.
\end{lstlisting}
Role 0 (player) Game Agent Role Prompt.
\begin{lstlisting}[style=gamepromptblock]
You control the bird. Steer left/right to navigate platforms and hazards. Hold Up to glide (slow descent). Use Up+Left/Right to glide diagonally. You can use glide actions consecutively for sustained slow descent.
\end{lstlisting}
Role 0 (player) Computer-Use Controls Prompt.
\begin{lstlisting}[style=gamepromptblock]
ACTION SPACE (ONLY LEGAL ACTIONS):

- ArrowLeft / ArrowRight: Move left/right
- ArrowUp: Hold to glide (slow descent, uses flap meter)
- ArrowUp + ArrowLeft / ArrowRight: Glide while moving left/right
\end{lstlisting}
Role 0 (player) Generalist Semantic Action List.
\begin{lstlisting}[style=gamepromptblock]
REGISTERED ACTIONS (Semantic Controls).
Choose exactly one action per step:

- wait: Stay still briefly.
- move_left: Move left.
- move_right: Move right.
- glide: Hold to slow descent using flap meter.
- glide_left: Glide while moving left (slow descent + left).
- glide_right: Glide while moving right (slow descent + right).
\end{lstlisting}

\paragraph{25-rocket-league-2d (Rocket League 2D).}
Game Rules Prompt.
\begin{lstlisting}[style=gamepromptblock]
You are playing Rocket League 2D, a side-view car soccer game.

Game Objective.
- Score goals by hitting the ball into the opponent's net.
- Prevent the opponent from scoring.
Game Rules.
- Drive left/right to position your car.
- Jump to hit the ball in the air.
- Boost can increase speed if available.
\end{lstlisting}
Role 0 (player) Game Agent Role Prompt.
\begin{lstlisting}[style=gamepromptblock]
You control the blue car. Drive, jump, and boost to hit the ball and score goals.
\end{lstlisting}
Role 0 (player) Computer-Use Controls Prompt.
\begin{lstlisting}[style=gamepromptblock]
ACTION SPACE (ONLY LEGAL ACTIONS):

- A/D: Drive
- W: Jump
- Space: Boost or speed up (if available)
\end{lstlisting}
Role 0 (player) Generalist Semantic Action List.
\begin{lstlisting}[style=gamepromptblock]
REGISTERED ACTIONS (Semantic Controls).
Choose exactly one action per step:

- wait: Pause briefly.
- move_left: Drive left.
- move_right: Drive right.
- jump: Jump to hit the ball.
- boost: Use boost or speed up if supported.
\end{lstlisting}

\paragraph{26-run-3 (Run 3).}
Game Rules Prompt.
\begin{lstlisting}[style=gamepromptblock]
You are playing Run 3, an endless tunnel runner.

Game Objective.
- Survive as long as possible while navigating gaps.
Game Rules.
- Strafe left/right to navigate gaps.
- Jump to cross gaps.
\end{lstlisting}
Role 0 (player) Game Agent Role Prompt.
\begin{lstlisting}[style=gamepromptblock]
You are the runner. Control the player to strafe and jump over gaps. You run through tunnels in Run 3.
\end{lstlisting}
Role 0 (player) Computer-Use Controls Prompt.
\begin{lstlisting}[style=gamepromptblock]
ACTION SPACE (ONLY LEGAL ACTIONS):

- ArrowLeft / ArrowRight: strafe left/right.
- ArrowUp or Space: jump.
\end{lstlisting}
Role 0 (player) Generalist Semantic Action List.
\begin{lstlisting}[style=gamepromptblock]
REGISTERED ACTIONS (Semantic Controls).
Choose exactly one action per step:

- wait: Pause briefly.
- move_left: Strafe left along the tunnel.
- move_right: Strafe right along the tunnel.
- jump: Jump across gaps.
\end{lstlisting}

\paragraph{27-stack (Stack).}
Game Rules Prompt.
\begin{lstlisting}[style=gamepromptblock]
You are playing Stack, a timing-based block stacking game.

Game Objective.
- Drop each moving block to align with the stack.
- Build the highest stack possible.
Game Rules.
- Each block moves back and forth over the tower.
- Press Space to drop the block.
- Any overhanging part is cut off.
- The game ends when there is no overlap.
\end{lstlisting}
Role 0 (player) Game Agent Role Prompt.
\begin{lstlisting}[style=gamepromptblock]
You control the stacking blocks. You are stacking blocks by timing drops.
\end{lstlisting}
Role 0 (player) Computer-Use Controls Prompt.
\begin{lstlisting}[style=gamepromptblock]
ACTION SPACE (ONLY LEGAL ACTIONS):

- Wait: Do nothing to wait the stack to move
- Space: Drop the block once
\end{lstlisting}
Role 0 (player) Generalist Semantic Action List.
\begin{lstlisting}[style=gamepromptblock]
REGISTERED ACTIONS (Semantic Controls).
Choose exactly one action per step:

- wait: Wait briefly.
- drop_block: Drop the block to place it.
\end{lstlisting}

\paragraph{28-temple-run-2 (Temple Run 2).}
Game Rules Prompt.
\begin{lstlisting}[style=gamepromptblock]
You are playing Temple Run 2, an endless runner.

Game Objective.
- Survive as long as possible without missing turns and avoid obstacles.
Game Rules.
- Run along the paths and avoid obstacles.
- Fail to turn, jump, or slide and you will die.
\end{lstlisting}
Role 0 (player) Game Agent Role Prompt.
\begin{lstlisting}[style=gamepromptblock]
You are the runner. Control the player to turn, jump, and slide to avoid obstacles. You run in Temple Run 2.
\end{lstlisting}
Role 0 (player) Computer-Use Controls Prompt.
\begin{lstlisting}[style=gamepromptblock]
ACTION SPACE (ONLY LEGAL ACTIONS):

- A/D: switch lanes and take turns.
- W: jump.
- S: slide.
- Space: start/continue.
\end{lstlisting}
Role 0 (player) Generalist Semantic Action List.
\begin{lstlisting}[style=gamepromptblock]
REGISTERED ACTIONS (Semantic Controls).
Choose exactly one action per step:

- wait: Pause briefly.
- start: Start or continue the run.
- turn_left: Switch left lanes or take a left turn.
- turn_right: Switch right lanes or take a right turn.
- jump: Jump over obstacles.
- slide: Slide under obstacles.
\end{lstlisting}

\paragraph{29-tetris (Tetris).}
Game Rules Prompt.
\begin{lstlisting}[style=gamepromptblock]
You are playing Tetris, a falling block puzzle game.

Game Objective.
- Clear lines by filling them with blocks.
- Prevent the stack from reaching the top and get the highest score possible.
Game Rules.
- Pieces fall from the top and can be moved or rotated.
- Completed horizontal lines clear and score points.
- The game ends when new pieces can no longer spawn or the stack reaches the top.
\end{lstlisting}
Role 0 (player) Game Agent Role Prompt.
\begin{lstlisting}[style=gamepromptblock]
You control the falling pieces. Control the player to move, rotate, drop, and swap pieces. You are playing Tetris. Keep the stack low and clear lines efficiently.
\end{lstlisting}
Role 0 (player) Computer-Use Controls Prompt.
\begin{lstlisting}[style=gamepromptblock]
ACTION SPACE (ONLY LEGAL ACTIONS):

- ArrowLeft / ArrowRight: move
- ArrowUp or X: rotate clockwise
- Z: rotate counter-clockwise
- ArrowDown: soft drop
- Space: hard drop
- Shift or C: Swap piece
- Enter or click: start/retry
\end{lstlisting}
Role 0 (player) Generalist Semantic Action List.
\begin{lstlisting}[style=gamepromptblock]
REGISTERED ACTIONS (Semantic Controls).
Choose exactly one action per step:

- wait: Wait briefly.
- move_left: Shift the piece left.
- move_right: Shift the piece right.
- soft_drop: Soft drop the piece faster.
- rotate_cw: Rotate the piece clockwise.
- rotate_ccw: Rotate the piece counter-clockwise.
- hard_drop: Hard drop the piece.
- hold_piece: Hold or swap the current piece.
\end{lstlisting}

\paragraph{30-vex-3 (Vex 3).}
Game Rules Prompt.
\begin{lstlisting}[style=gamepromptblock]
You are playing Vex 3, a precision platformer with traps.

Game Objective.
- Reach the checkpoints and then the exit of each level.
- Avoid spikes, saws, and falling hazards.
Game Rules.
- Move left/right and jump between platforms.
- Some sections require crouch sliding or dropping through gaps.
- Death resets you to checkpoints.
\end{lstlisting}
Role 0 (player) Game Agent Role Prompt.
\begin{lstlisting}[style=gamepromptblock]
You control the Vex character. Control the player to move, jump, and slide. You are a platform runner. Reach the exit while avoiding traps.
\end{lstlisting}
Role 0 (player) Computer-Use Controls Prompt.
\begin{lstlisting}[style=gamepromptblock]
ACTION SPACE (ONLY LEGAL ACTIONS):

- ArrowLeft / ArrowRight or A/D: move
- ArrowUp + ArrowLeft / ArrowRight: jump left/right
- ArrowDown + ArrowLeft / ArrowRight: crouch slide left/right
\end{lstlisting}
Role 0 (player) Generalist Semantic Action List.
\begin{lstlisting}[style=gamepromptblock]
REGISTERED ACTIONS (Semantic Controls).
Choose exactly one action per step:

- wait: Pause briefly.
- move_left: Move left.
- move_right: Move right.
- jump: Jump.
- jump_left: Jump left.
- jump_right: Jump right.
- crouch_slide_left: Crouch slide left.
- crouch_slide_right: Crouch slide right.
\end{lstlisting}

\paragraph{31-wolf3d (Wolfenstein 3D).}
Game Rules Prompt.
\begin{lstlisting}[style=gamepromptblock]
You are playing Wolfenstein 3D, a first-person shooter game.

Game Objective.
- Search for and defeat enemy guards by shooting them.
- Survive while maximizing kills.
Game Rules.
- You control a soldier in a first-person 3D view.
- Guards will shoot at you when triggered (shot by you or you are nearby), reducing your health.
- You start with a pistol and 8 bullets.
- Killed guards may drop ammo clips.
- If your health reaches 0, you die and respawn (losing one life).
- The game ends when you run out of lives.
- Some enemies are in an adjacent room behind a door. Use the SPACE key near a door to open it.
\end{lstlisting}
Role 0 (player) Game Agent Role Prompt.
\begin{lstlisting}[style=gamepromptblock]
You control a soldier in first-person view. Shoot guards to defeat them. Manage your ammo and health. Use arrow keys to move and turn, X to shoot, SPACE to open doors.
\end{lstlisting}
Role 0 (player) Computer-Use Controls Prompt.
\begin{lstlisting}[style=gamepromptblock]
ACTION SPACE (ONLY LEGAL ACTIONS):

- ArrowUp: Move forward
- ArrowDown: Move backward
- ArrowLeft: Turn left
- ArrowRight: Turn right
- x: Shoot / Attack
- Space: Use / Open door
\end{lstlisting}
Role 0 (player) Generalist Semantic Action List.
\begin{lstlisting}[style=gamepromptblock]
REGISTERED ACTIONS (Semantic Controls).
Choose exactly one action per step:

- wait: Pause briefly to observe.
- move_forward: Move forward.
- move_backward: Move backward.
- turn_left: Turn left.
- turn_right: Turn right.
- shoot: Shoot / Attack.
- open_door: Use / Open door.
\end{lstlisting}

\paragraph{32-wordle (Wordle).}
\normalfont{Game Rules Prompt.}
\begin{lstlisting}[style=gamepromptblock]
You are playing Wordle, a word guessing game.

Game Objective.
- Guess the hidden five-letter word in six tries.
- Use color feedback to refine guesses.
Game Rules.
- Each guess must be a valid five-letter word.
- Green means correct letter in the correct place.
- Yellow means correct letter in the wrong place.
- Gray means the letter is not in the word.
- The guess is auto-submitted when you type the 5th letter.
\end{lstlisting}
Role 0 (player) Game Agent Role Prompt.
\begin{lstlisting}[style=gamepromptblock]
You control the keyboard input. Type a valid five-letter word to make a guess. The guess is auto-submitted when 5 letters are entered. Use color feedback (green/yellow/gray) to narrow down the answer.
\end{lstlisting}
Role 0 (player) Computer-Use Controls Prompt.
\begin{lstlisting}[style=gamepromptblock]
ACTION SPACE (ONLY LEGAL ACTIONS):

- Type a 5-letter word (e.g., type "hello"). 
- The guess is auto-submitted when 5 letters are entered.
- Use Backspace to delete a letter
\end{lstlisting}
Role 0 (player) Generalist Semantic Action List.
\begin{lstlisting}[style=gamepromptblock]
REGISTERED ACTIONS (Semantic Controls).
Choose exactly one action per step:

- wait: Wait briefly.
- type_text: Type letters by setting a text value.
- submit_guess: Confirm the guess by pressing Enter.
\end{lstlisting}

\paragraph{33-worlds-hardest-game (World's Hardest Game).}
Game Rules Prompt.
\begin{lstlisting}[style=gamepromptblock]
You are playing World's Hardest Game, a precision maze dodge game.

Game Objective.
- Collect all coins in the level.
- Reach the green exit zone to complete the level.
Game Rules.
- You control a red square in a maze.
- Blue enemies move on fixed paths and kill you on contact.
- Checkpoints save progress after collecting items. Use checkpoints to split the level into safe segments.
- Observe enemy cycles carefully to avoid them.
\end{lstlisting}
Role 0 (player) Game Agent Role Prompt.
\begin{lstlisting}[style=gamepromptblock]
You control the red square. Move to collect coins and reach the exit while avoiding enemies. Avoid blue enemies, collect all coins, and reach the green exit.
\end{lstlisting}
Role 0 (player) Computer-Use Controls Prompt.
\begin{lstlisting}[style=gamepromptblock]
ACTION SPACE (ONLY LEGAL ACTIONS):

- ArrowUp: Move up
- ArrowDown: Move down
- ArrowLeft: Move left
- ArrowRight: Move right
\end{lstlisting}
Role 0 (player) Generalist Semantic Action List.
\begin{lstlisting}[style=gamepromptblock]
REGISTERED ACTIONS (Semantic Controls).
Choose exactly one action per step:

- wait: Pause briefly to observe.
- move_up: Move up.
- move_down: Move down.
- move_left: Move left.
- move_right: Move right.
\end{lstlisting}

\paragraph{34-worlds-hardest-game-2 (World's Hardest Game 2).}
Game Rules Prompt.
\begin{lstlisting}[style=gamepromptblock]
You are playing World's Hardest Game, a precision maze dodge game.

Game Objective.
- Collect all coins in the level.
- Reach the green exit zone to complete the level.
Game Rules.
- You control a red square in a maze.
- Blue enemies move on fixed paths and kill you on contact.
- Observe enemy cycles carefully to avoid them
\end{lstlisting}
Role 0 (player) Game Agent Role Prompt.
\begin{lstlisting}[style=gamepromptblock]
You control the red square. Move to collect items and reach the exit while avoiding enemies. Avoid blue enemies, collect all items, and reach the green exit.
\end{lstlisting}
Role 0 (player) Computer-Use Controls Prompt.
\begin{lstlisting}[style=gamepromptblock]
ACTION SPACE (ONLY LEGAL ACTIONS):

- ArrowUp: Move up
- ArrowDown: Move down
- ArrowLeft: Move left
- ArrowRight: Move right
\end{lstlisting}
Role 0 (player) Generalist Semantic Action List.
\begin{lstlisting}[style=gamepromptblock]
REGISTERED ACTIONS (Semantic Controls).
Choose exactly one action per step:

- wait: Pause briefly to observe.
- move_up: Move up.
- move_down: Move down.
- move_left: Move left.
- move_right: Move right.
\end{lstlisting}

\normalsize

}

\subsection{Model Output-Format Blocks}
\label{sec:app_output_formats}

Below we list the exact \texttt{output\_format} block from each registered model specification.

{
\renewcommand{\paragraph}[1]{\par\medskip\noindent{\normalfont\bfseries #1}\par\smallskip}
\paragraph{Claude-Sonnet-4.6 (Computer-Use).}
\begin{lstlisting}[style=gamepromptblock]
- Use the Claude computer-use tool to return exactly one action or key combination per step.
- You are not allowed to take screenshots on your own.
- Do not output free-form text outside tool calls.
\end{lstlisting}

\paragraph{Claude-Sonnet-4.6 (Generalist).}
\begin{lstlisting}[style=gamepromptblock]
- You must call exactly ONE tool per step.
- The tool name must be a registered action id.
- Include `reasoning` as a short rationale.
- Do not output free-form text.
\end{lstlisting}

\paragraph{Gemini-2.5-Computer-Use.}
\begin{lstlisting}[style=gamepromptblock]
- Use the built-in computer-use tool to return one action or key combination per step.
- Do not output free-form text outside tool calls.
\end{lstlisting}

\paragraph{Gemini-3-Flash-Preview.}
\begin{lstlisting}[style=gamepromptblock]
- You must call exactly ONE tool per step.
- The tool name must be a registered action id.
- Include `reasoning` as a short rationale.
- Do not output free-form text.
\end{lstlisting}

\paragraph{GLM-4.6V.}
\begin{lstlisting}[style=gamepromptblock]
- You must call exactly ONE tool per step with tool_calls.
- The tool name must be a registered action id.
- Include `reasoning` as a short rationale.
- Do not output free-form text.
\end{lstlisting}

\paragraph{Grok-4.1-Fast-Reasoning.}
\begin{lstlisting}[style=gamepromptblock]
- You must call exactly ONE tool per step.
- The tool name must be a registered action id.
- Include `reasoning` as a short rationale.
- Do not output free-form text.
\end{lstlisting}

\paragraph{Kimi-K2.5.}
\begin{lstlisting}[style=gamepromptblock]
- You must call exactly ONE tool per step.
- The tool name must be a registered action id.
- Include `reasoning` as a short rationale.
- Do not output free-form text.
\end{lstlisting}

\paragraph{Qwen3-VL-235B-A22B (Computer-Use).}
\begin{lstlisting}[style=gamepromptblock]
Response format for every step:
A <think> ... </think> block of a very short sentence describing what to do.
A single <tool_call>...</tool_call> block containing only the JSON: {"name": "<function-name>", "arguments": <args-json-object>}

Use the `computer_use` tool call and return exactly one action per step.
Use only the <think> and the <tool_call> block; do not add any other text.
\end{lstlisting}

\paragraph{Qwen3-VL-235B-A22B (Generalist).}
\begin{lstlisting}[style=gamepromptblock]
Response format for every step:
A <think> ... </think> block of a very short sentence describing what to do.
A single <tool_call>...</tool_call> block containing only the JSON: {"name": "<function-name>", "arguments": <args-json-object>}

Use only the <think> and the <tool_call> block; do not add any other text.
\end{lstlisting}

\paragraph{Qwen3-VL-30B-A3B (Computer-Use).}
\begin{lstlisting}[style=gamepromptblock]
Response format for every step:
A <think> ... </think> block of a very short sentence describing what to do.
A single <tool_call>...</tool_call> block containing only the JSON: {"name": "<function-name>", "arguments": <args-json-object>}

Use the `computer_use` tool call and return exactly one action per step.
Use only the <think> and the <tool_call> block; do not add any other text.
\end{lstlisting}

\paragraph{Qwen3-VL-30B-A3B (Generalist).}
\begin{lstlisting}[style=gamepromptblock]
Response format for every step:
A <think> ... </think> block of a very short sentence describing what to do.
A single <tool_call>...</tool_call> block containing only the JSON: {"name": "<function-name>", "arguments": <args-json-object>}

Use only the <think> and the <tool_call> block; do not add any other text.
\end{lstlisting}

\paragraph{OpenAI-Computer-Use.}
\begin{lstlisting}[style=gamepromptblock]
- Use the computer-use tool to return exactly one action or key combination per step.
- Do not output free-form text outside tool calls.
- Do not take screenshots on your own.
\end{lstlisting}

\paragraph{GPT-5.2.}
\begin{lstlisting}[style=gamepromptblock]
- You must call exactly ONE tool per step.
- The tool name must be a registered action id.
- Include `reasoning` as a short rationale.
- Do not output free-form text.
\end{lstlisting}

\paragraph{Qwen3-VL-Plus (Computer-Use).}
\begin{lstlisting}[style=gamepromptblock]
- Use the `computer_use` tool call and return exactly one action or key combination per step.
- Do not output free-form text outside <tool_call> blocks.
\end{lstlisting}

\paragraph{Qwen3-VL-Plus (Generalist).}
\begin{lstlisting}[style=gamepromptblock]
- You must call exactly ONE tool per step.
- The tool name must be a registered action id.
- Include `reasoning` as a short rationale.
- Do not output free-form text.
\end{lstlisting}

\paragraph{Seed-1.8 (Computer-Use).}
\begin{lstlisting}[style=gamepromptblock]
ACTION FORMAT (use exactly this syntax):
- Press single key: hotkey(key='<key>')
- Press multiple keys: hotkey(key='<key1> <key2>')
- Click at position: click(point='<point>x y</point>')
- Right click at position: right_single(point='<point>x y</point>')
- Wait/observe: wait()

Examples:
- hotkey(key='w')           # Press W
- hotkey(key='w d')         # Press W and D together (jump right)
- hotkey(key='arrowup')     # Press Up arrow
- click(point='<point>640 360</point>')  # Click at center
- right_single(point='<point>640 360</point>')  # Right click at center
\end{lstlisting}

\paragraph{Seed-1.8 (Generalist).}
\begin{lstlisting}[style=gamepromptblock]
- You must call exactly ONE tool per step.
- The tool name must be a registered action id.
- Include `reasoning` as a short rationale.
- Do not output free-form text.
\end{lstlisting}

\paragraph{UI-TARS-1.5-7B.}
\begin{lstlisting}[style=gamepromptblock]
ACTION FORMAT (use exactly this syntax):
- Press single key: hotkey(key='<key>')
- Press multiple keys: hotkey(key='<key1> <key2>')
- Click at position: click(point='<point>x y</point>')
- Right click at position: right_single(point='<point>x y</point>')
- Wait/observe: wait()

Examples:
- hotkey(key='w')           # Press W
- hotkey(key='w d')         # Press W and D together (jump right)
- hotkey(key='arrowup')     # Press Up arrow
- click(point='<point>640 360</point>')  # Click at center
- right_single(point='<point>640 360</point>')  # Right click at center
\end{lstlisting}

\normalsize

}

\section{Costs and Licensing Considerations}
\label{sec:app_licensing}

\paragraph{Licensing Considerations.}

\gameworld{} spans both proprietary and open-source browser games. We are sincerely grateful to the original game creators and rights holders whose work makes this benchmark possible. Our release policy is designed to respect upstream authorship, licensing terms, and distribution requirements. In particular, users are responsible for purchasing or obtaining lawful access to any benchmarked game. Any game access or distribution must comply with the applicable licenses and permissions. 

Here we explicitly state the following licensing and compliance guidelines: "\textit{\emph{This project is intended strictly for research and benchmarking purposes. It does not grant any rights to access, reproduce, distribute, modify, or commercially use third-party games or related assets beyond those permitted by applicable licenses, terms of service, and law. Users are solely responsible for purchasing or obtaining lawful access and any necessary permissions for evaluation, dataset creation, model development, or downstream use.}}"

\paragraph{Cost Summary.}
\begin{table}[!h]
    \centering
    \small
    \setlength{\tabcolsep}{5pt}
    \caption{Estimated benchmark cost of evaluating all 170 tasks for each model. Input / output token counts are averaged per step.}
    \label{tab:benchmark_cost}
    \begin{tabular}{@{}lccc@{}}
    \toprule
    \textbf{Model} & \shortstack{\textbf{Input} \\ \textbf{Tokens / Step}} & \shortstack{\textbf{Output} \\ \textbf{Tokens / Step}} & \shortstack{\textbf{Total Cost} \\ \textbf{(USD)}} \\
    \midrule
    Claude-Sonnet-4.6 (Computer-Use) & 3344.9 & 131.7 & 172.46 \\
    Claude-Sonnet-4.6 (Generalist) & 4865.8 & 86.3 & 244.03 \\
    Gemini-2.5-Computer-Use & 2086.6 & 15.4 & 41.06 \\
    Gemini-3-Flash-Preview & 4005.6 & 40.5 & 29.13 \\
    GLM-4.6V & 4820.2 & 253.2 & 24.79 \\
    GPT-5.2 & 3924.6 & 38.9 & 110.68 \\
    Grok-4.1-Fast-Reasoning & 2551.1 & 907.7 & 9.86 \\
    Kimi-K2.5 & 5030.9 & 250.0 & 45.77 \\
    OpenAI-Computer-Use & 1684.4 & 87.2 & 94.65 \\
    Qwen3-VL-Plus (Computer-Use) & 2112.3 & 27.7 & 4.99 \\
    Qwen3-VL-Plus (Generalist) & 3675.1 & 61.3 & 9.01 \\
    Seed-1.8 (Computer-Use) & 2184.0 & 210.6 & 13.91 \\
    Seed-1.8 (Generalist) & 2642.6 & 179.9 & 14.85 \\
    \midrule
    \multicolumn{3}{r}{\textbf{Total Cost (all listed models)}} & \textbf{815.19} \\
    \bottomrule
    \end{tabular}
\end{table}

Table~\ref{tab:benchmark_cost} reports average input and output tokens \emph{per step}, together with the estimated total dollar cost for evaluating all 170 benchmark tasks.
For the input-token column, we include cached input tokens when present, i.e., \texttt{per\_step\_input + per\_step\_cache} if the model API supports caching.
The total cost column is the measured average cost per task from all the trace logs. 
The underlying model pricing estimates are taken from the pricing snapshot recorded on March 7, 2026.
The cost for open-weight models, including Qwen3-VL-235B-A22B, Qwen3-VL-30B-A3B, and UI-TARS-1.5-7B, is not included in the calculation.
The final total cost for evaluating all 170 tasks across all listed models is \textbf{815.19 USD}.

\bibliographystyle{plainnat}
\bibliography{main}

\end{document}